%% file: neurips_2026.tex
\documentclass{article}

\usepackage[main, final, nonatbib]{neurips_2026}

\usepackage[numbers]{natbib}
\usepackage[utf8]{inputenc} %
\usepackage[T1]{fontenc}    %
\usepackage{hyperref}       %
\usepackage{url}            %
\usepackage{paralist}       %

\usepackage{booktabs}       %
\usepackage{amsfonts}       %
\usepackage{nicefrac}       %
\usepackage{microtype}      %
\usepackage[table]{xcolor}         %
\usepackage{custom_sty}
\usepackage{pgfmath}
\usepackage{amssymb}

\input{math_commands.tex}

\usepackage{graphicx}
\usepackage{subcaption}
\usepackage{multirow}
\usepackage{arydshln}
\usepackage{fontawesome5}
\definecolor{cAll}{HTML}{7B44DB}       %
\definecolor{cIID}{HTML}{007A20}       %
\definecolor{cOOD}{HTML}{FB6400}       %
\definecolor{cEnc}{HTML}{4A90D9}       %
\definecolor{cEmb}{HTML}{E8A838}       %
\definecolor{cAgr}{HTML}{5C6BC0}       %
\usepackage{amsmath}
\usepackage{cleveref}
\crefname{figure}{Fig.}{Figs.}
\crefname{table}{Tab.}{Tabs.}
\crefname{equation}{Eq.}{Eqs.}
\crefname{appendix}{App.}{Apps.}
\usepackage{microtype}
\usepackage{pgf}
\usepackage{tikz}
\usepackage{pgfplots}
\usepackage{algpseudocode}
\usepackage{makecell}
\usepackage{contour}
\usepackage{placeins}
\contourlength{0.6pt}
\usetikzlibrary{calc, shapes, patterns, shadings, positioning, arrows.meta,intersections,plotmarks}
\usepgfplotslibrary{fillbetween}

\usepackage[skip=5pt]{caption}
\usepackage{titlesec}
\titleformat{\paragraph}[runin]
  {\normalfont\normalsize\bfseries}
  {\theparagraph}
  {1em}
  {}
\titlespacing*{\paragraph}
  {0pt}
  {5pt plus 2pt minus 1pt}
  {0.5em}

\newcommand{\showchange}[1]{%
    \pgfmathsetmacro{\result}{#1}%
    \pgfmathparse{\result > 0 ? "green" : "red"}%
    \textcolor{\pgfmathresult}{%
        \ifdim\result pt>0pt\tiny$\uparrow$\else\tiny$\downarrow$\fi%
        \tiny\pgfmathprintnumber[fixed, precision=1, showpos]{\result}%
    }%
}
\newcommand{\showchangedark}[1]{%
    \pgfmathsetmacro{\result}{#1}%
    \pgfmathparse{\result > 0 ? "cIID" : "red"}%
    \textcolor{\pgfmathresult}{%
        \ifdim\result pt>0pt\tiny$\uparrow$\else\tiny$\downarrow$\fi%
        \tiny\pgfmathprintnumber[fixed, precision=1, showpos]{\result}%
    }%
}

\newcommand{\omitme}[1]{}
\newcommand{\td}[1]{}%

\newcommand{\orangefire}{{\color{orange}\faFire}}
\newcommand{\bluesnow}{{\color{cyan}\faSnowflake}}

\title{Who Needs Labels? Adapting Vision Foundation Models With the Metadata You Already Have}

\author{%
  Elouan Gard\`es\textsuperscript{1,2} \And
  Seung Eun Yi\textsuperscript{1} \And
  Kartik Ahuja\textsuperscript{1} \AND
  Th\'eo Moutakanni\textsuperscript{1} \And
  Huy V.~Vo\textsuperscript{1} \And
  Piotr Bojanowski\textsuperscript{1} \And
  Wolfgang M.~Pernice\textsuperscript{3}$^*$ \And
  Lo\"ic Landrieu\textsuperscript{2}$^*$ \And
  Camille Couprie\textsuperscript{1}$^*$
  \end{tabular}\hfil\linebreak[4]\hfil\begin{tabular}[t]{@{}c@{}}\rule{0pt}{14pt}%
  \normalfont\footnotesize
  \textsuperscript{1}Meta FAIR\enspace
  \textsuperscript{2}LIGM, CNRS, Univ Gustave Eiffel, ENPC, IPP\enspace
  \textsuperscript{3}Columbia University, New York\enspace
  $^*$Equal supervision
}

\pgfmathsetseed{1679554}
\begin{document}
\setlength{\leftmargini}{3em}

\maketitle

\begin{abstract}
We propose a label-free approach to adapt powerful but generic vision foundation models to specialized scientific domains. Standard supervised fine-tuning is often ill-suited to these settings: labels are scarce, and task-specific training can reduce the generality needed to cope with complex domain shifts. We instead leverage metadata to adapt representations to new domains in a weakly-supervised manner.
Our method, \textsc{FINO}, combines a standard self-supervised objective with flexible metadata guidance that handles both highly granular discrete metadata and continuous metadata. It encourages the representation to preserve informative factors while suppressing spurious ones.
Across subcellular fluorescence microscopy, Earth observation, wildlife monitoring, and medical imaging, \textsc{FINO} outperforms matched self-supervised and fully supervised adaptation baselines. Complete-adaptation repeats confirm consistent performance across domains. Without task labels for backbone adaptation, \textsc{FINO} also obtains higher reported scores than selected specialised systems on three domains and matches the fourth under differing cross-paper protocols.\\
Code is available at \href{https://github.com/facebookresearch/dinov3/tree/FINO#metadata-guided-training}{\texttt{github.com/facebookresearch/dinov3/tree/FINO}}.
\end{abstract}

\setlength{\parskip}{5.5pt plus 6pt}
\section{Introduction}
\label{sec:intro}
\input{sections/1_intro}

\pagebreak
\setlength{\parskip}{5.5pt}
\section{Related work}
\label{sec:related}
\input{sections/2_related}

\section{Method}
\label{sec:method}
\input{sections/3_method}

\section{Results}
\label{sec:results}
\input{sections/4_experiments}

\FloatBarrier
\section{Conclusion}
We introduced \emph{representation adaptation} and \textsc{FINO}, which uses freely available metadata to specialise foundation models without task labels. Across four scientific domains, \textsc{FINO} improves over matched supervised, domain-adaptation, and self-supervised baselines, with complete-adaptation repeats demonstrating stable performance. On FMoW and HPA, metadata provides stronger OOD guidance than conventional SupCon, task-label prototypes, or their combination with metadata. \textsc{FINO} also remains effective with limited labels and transfers across datasets and prediction tasks, showing that its representations are reusable rather than tied to one label space. Together, these results establish metadata as practical supervision for robust adaptation, while highlighting the need to distinguish informative, spurious, redundant, and entangled factors.

\paragraph{Broader impact.}
Because \textsc{FINO} uses metadata as weak supervision, it may reproduce or amplify biases encoded by sensitive or imbalanced attributes, such as gender or geographic origin. Treating such factors as informative can lead to uneven performance across groups. Metadata should therefore be selected carefully and audited beyond aggregate performance, particularly in medical and geographic applications. All datasets used in this work are established research benchmarks, and we do not release new personal data.

\paragraph{Reproducibility.}
Our public repository, \url{https://github.com/facebookresearch/dinov3/tree/FINO}, provides the end-to-end code for data and metadata download, preprocessing, dataloading, every training stage, probing, and evaluation, together with exact configurations, environments, splits, seeds, commands, metrics, and table-generation scripts. The appendix documents the corresponding protocols. Laboratory policy prevents redistribution of trained checkpoints.

\begin{ack}
This work was partially supported by NIH/NHGRI award 5R00HG011488-05 to WMAP, Hi!PARIS (ANR-23-IACL-0005), and the PEPR SHARP (PEPR IA, ANR-23-PEIA-0008).
\end{ack}

\clearpage

\bibliographystyle{unsrtnat}
\bibliography{bibl}

\clearpage
\appendix

\maketitlesupplementary

\input{sections/supmat}

\clearpage
\input{checklist.tex}

\end{document}

%% file: math_commands.tex
\usepackage{amsmath,amsfonts,bm}

\def\eqref#1{equation~\ref{#1}}

\def\1{\bm{1}}

\def\vm{{\bm{m}}}

\def\mW{{\bm{W}}}

\DeclareMathAlphabet{\mathsfit}{\encodingdefault}{\sfdefault}{m}{sl}
\SetMathAlphabet{\mathsfit}{bold}{\encodingdefault}{\sfdefault}{bx}{n}

\def\sR{{\mathbb{R}}}

%% file: sections/1_intro.tex
Visual foundation models provide rich, general representations that perform well across many natural-image tasks~\cite{oquab2024dinov2learningrobustvisual,radford2021clip,tschannen2025siglip}. However, they often degrade on specialised domains such as biomedical microscopy~\cite{zhang2025biomedclip, pernice2023out} and Earth observation~\cite{ayush2021geography}, whose distributions differ substantially from natural images~\cite{torralba2011unbiased}. These datasets also exhibit subtle but pervasive domain shifts~\cite{wang2022generalizing} that arise from both meaningful variation, such as geography~\cite{kuriyal2025codexcombiningdomainexpertise} or staining protocols~\cite{noori2026histopath}, and spurious factors such as sensor characteristics or
reagent aliquot, often called \emph{batch effects} in biology~\cite{stacke2021measuring}. Fortunately, such datasets are also typically accompanied by rich metadata correlated to these variations, and which are available at virtually no extra cost.

A standard way to adapt foundation models to a new domain is supervised fine-tuning~\cite{veasey2024parameter,steinertrain,touvron2019fixing,huang2025knowledge}.
In scientific settings, however, labels are often scarce or expensive \cite{sullivan2018deep}, and fine-tuning can overfit task-specific signals, hurting out-of-distribution (OOD) generalisation~\cite{koh2021wildsbenchmarkinthewilddistribution}. In the process, it can erase much of the generality of the pretrained model, occasionally yielding worse performance than zero-shot inference. Unsupervised domain adaptation instead leverages unlabelled data from the target distribution~\cite{ganin2015unsuperviseddomainadaptationbackpropagation,long2015learning}, but remains task-driven and can likewise induce task-specific collapse~\cite{arazo2020pseudo,rosenfeldrisks}.

In this work, we study \emph{representation adaptation}: specialising a pretrained foundation model to a new application domain without task labels, then evaluating the frozen adapted representation with lightweight probes (Fig.~\ref{fig:teaser}). Purely self-supervised adaptation can struggle under distribution shift, as its representations remain sensitive to low-level or context-specific cues~\cite{pernice2023out,haslum2024metadata}. Recent work has shown that metadata can help steer self-supervised adaptation toward more robust representations~\cite{pernice2023out,haslum2024metadata,bourcier2024learning,holland2024metadata,lan2024hierarchical}. However, existing approaches remain partial: they are often tailored to a single modality or assume only a small number of domains, whereas scientific metadata is highly granular, spanning thousands of discrete categories or continuous values.

We propose \textsc{FINO} (FIne tuning with NO labels), a unified framework for metadata-driven representation adaptation (Fig.~\ref{fig:teaser}). It augments a self-supervised objective with metadata-guided losses and targeted regularisation to preserve informative structure while suppressing spurious variation. By combining prototype-based contrastive guidance, self-supervision, and appropriate regularisation, our method scales to heterogeneous and highly granular metadata, where classical domain generalisation methods struggle.

\begin{figure}[t]
    \input{figures/teaser_smaller}
\caption{{\bf Learning with metadata.} Scientific datasets (a) often include metadata describing their content and acquisition. \textsc{FINO} uses these signals as weak supervision (WSL), together with a self-supervised objective (SSL), to adapt a generic foundation model without task labels (b). In matched evaluations, its frozen representations outperform supervised fine-tuning; the right panel compares \textsc{FINO} means to reported specialised-system results under differing protocols (c).}
\label{fig:teaser}
\end{figure}
\footnotetext{The X-ray illustration is not from MIMIC-CXR, but a personal Xray from one of the authors.}

Empirically, \textsc{FINO} improves OOD performance over matched self-supervised, supervised, and domain-adaptation baselines without using unlabelled test data. Complete-adaptation repeats confirm consistent performance across domains, and metadata guidance outperforms task-label guidance in the tested FMoW and HPA settings. \textsc{FINO} also obtains higher reported scores than selected specialised systems on HPA~\citep{ouyang2019hpa}, FMoW~\citep{christie2018fmow}, and iWildCam~\citep{beery2020iwildcam}, and matches the MIMIC-CXR~\citep{johnson2019mimic} reference under differing data, supervision, tuning, and inference protocols. The objective and loss weights are shared across domains, with dataset-specific metadata and evaluation and a documented iWildCam schedule exception.
Our contributions are:
\begin{compactitem}
    \item \textsc{FINO}, a unified metadata-guided adaptation framework for discrete and continuous metadata that requires neither task labels nor metadata at inference time; and
    \item controlled evidence across four scientific domains that metadata guidance improves task-agnostic representations over matched adaptation baselines, supported by complete-adaptation repeats, metadata-assignment evaluations, and task-label-supervised comparisons.
\end{compactitem}

%% file: figures/teaser_smaller.tex
\def\wimg{1}%
\definecolor{MODELCOLOR}{RGB}{153,102,255}%
\definecolor{LOSSCOLOR}{RGB}{230,102,102}%
\definecolor{IMGCOLOR}{RGB}{102,178,255}%
\definecolor{LABELCOLOR}{RGB}{102,204,153}%
\definecolor{METACOLOR}{RGB}{255,178,102}%
\definecolor{FROZENCOLOR}{RGB}{50,225,245}%
\definecolor{SUPCOLOR}{RGB}{27,158,119}%
\definecolor{METACOLOR}{RGB}{231,41,138}%
\definecolor{SOTACOLOR}{RGB}{102,166,30}%
\centering
\begin{minipage}[t]{.48\linewidth}
\vspace{0pt}\centering
\begin{subfigure}[t]{\linewidth}
\centering\footnotesize
\begin{tabular}{@{}>{\centering\arraybackslash}m{0.47\linewidth}@{\hspace{0.06\linewidth}}>{\centering\arraybackslash}m{0.47\linewidth}@{}}
\includegraphics[width=\linewidth, height=\linewidth]{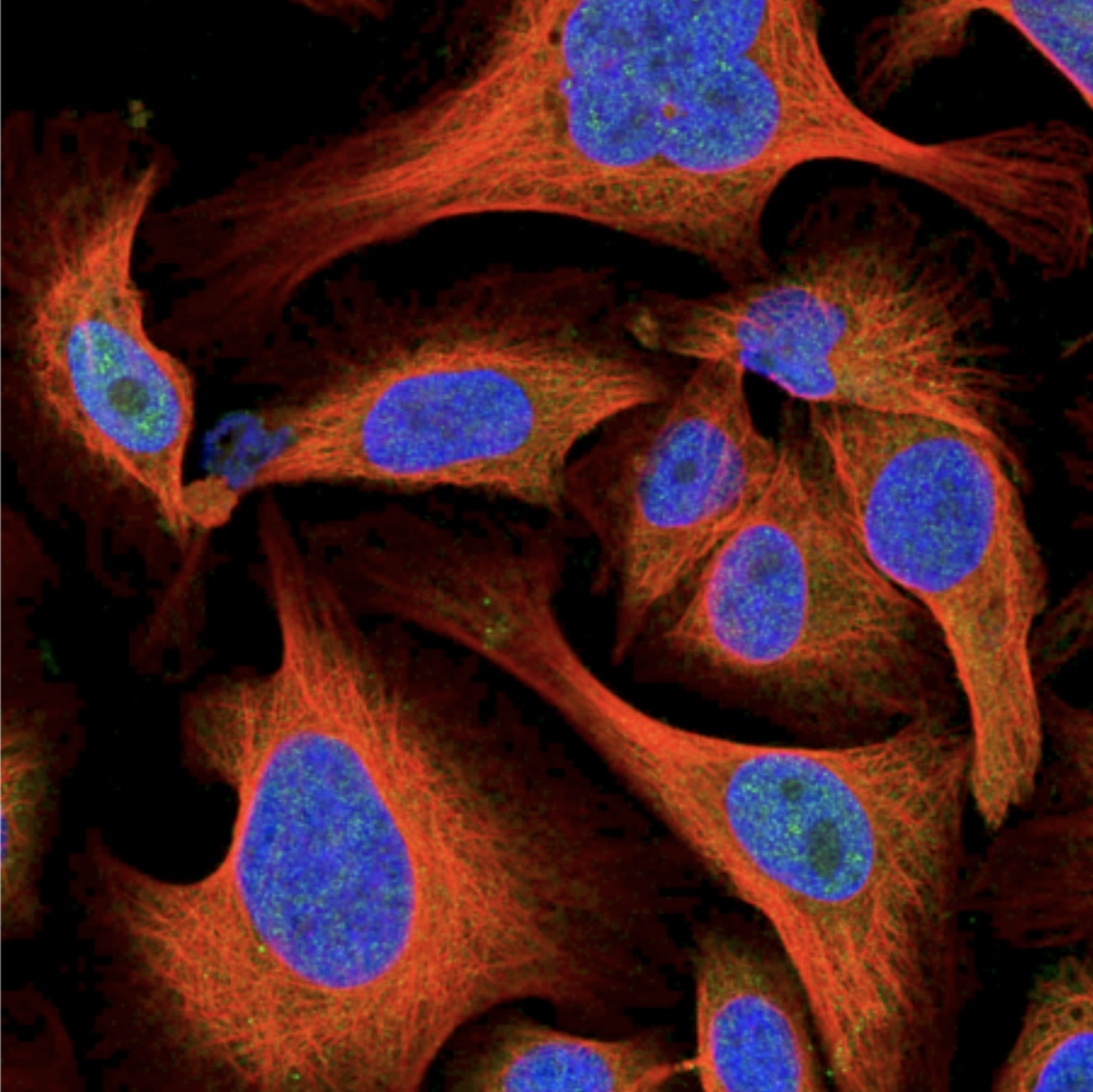} &
\includegraphics[width=\linewidth, height=\linewidth, trim=172 0 173 0, clip]{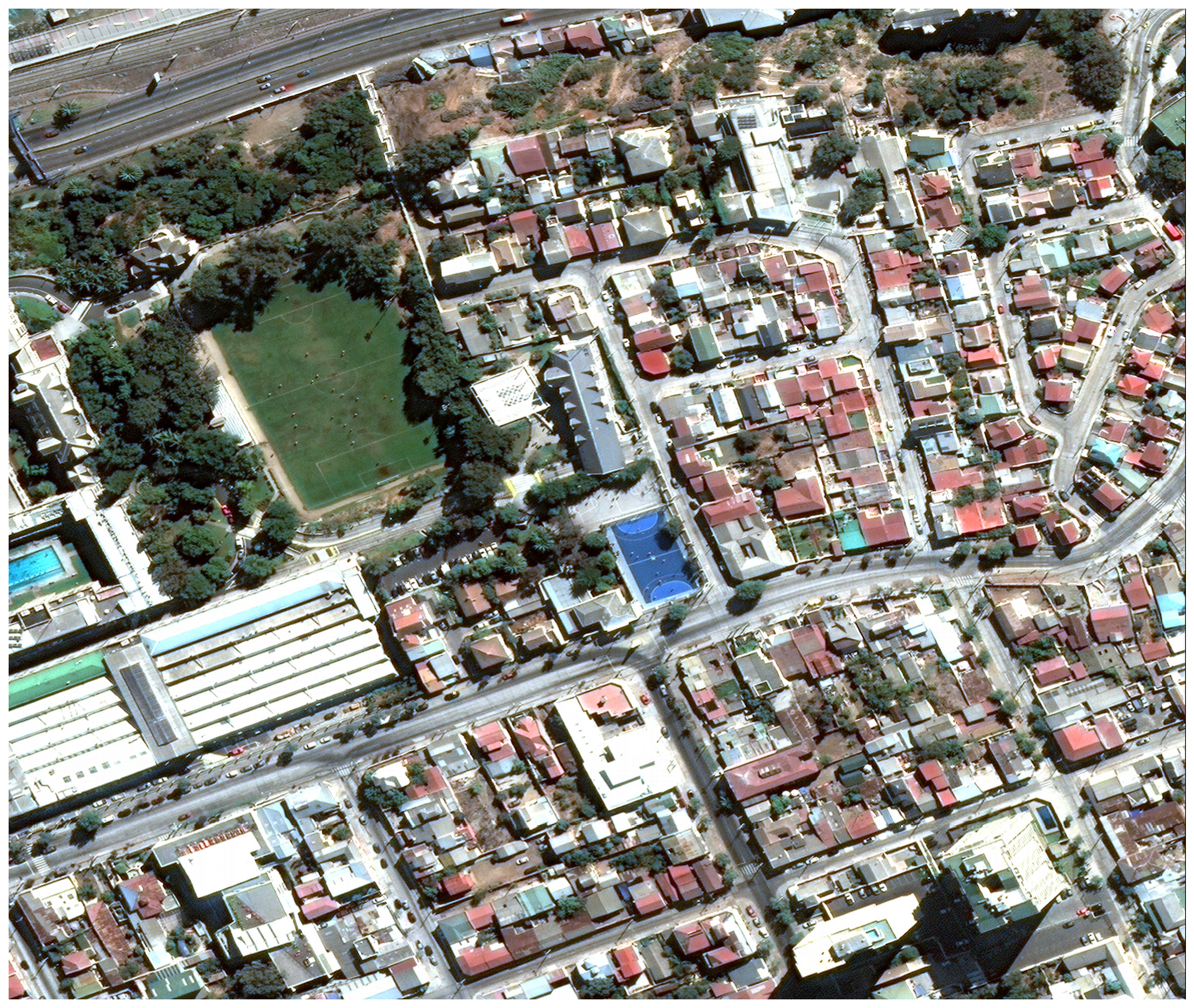} \\
\begin{tabular}[c]{@{}c@{}}
  \multicolumn{1}{c}{HPA} \\[-2pt] \cmidrule(lr){1-1}
  \textcolor{METACOLOR!70!black}{\discrete plates}\\
  \textcolor{METACOLOR!70!black}{\discrete antibody}\\
  \textcolor{METACOLOR!70!black}{\discrete cell line}
\end{tabular}
&
\begin{tabular}[c]{@{}c@{}}
   \multicolumn{1}{c}{FMoW}  \\[-2pt] \cmidrule(lr){1-1}
  \textcolor{METACOLOR!70!black}{\discrete country}\\
  \textcolor{METACOLOR!70!black}{\discrete weather}\\
  \textcolor{METACOLOR!70!black}{\continuous viewing angle}
\end{tabular}
\\
\includegraphics[width=\linewidth, height=\linewidth]{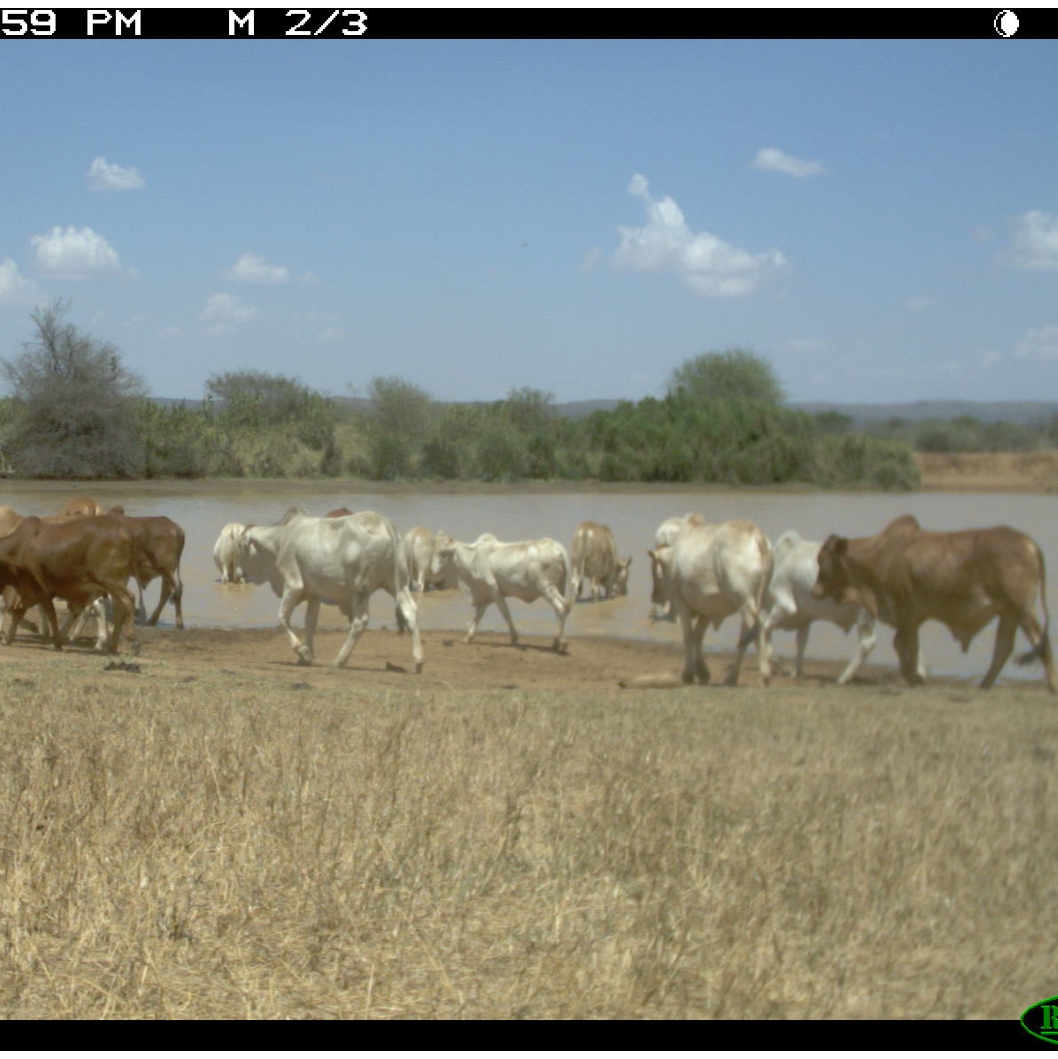}
&
\includegraphics[width=\linewidth, height=\linewidth, trim=0 60 0 60, clip]{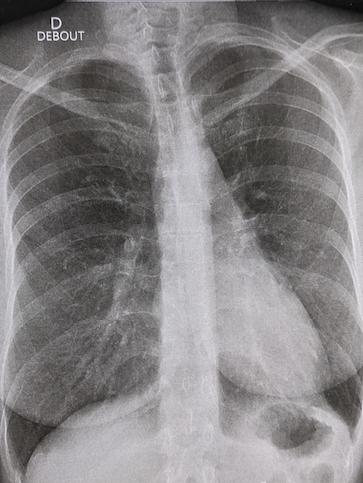} %
\\
\begin{tabular}[c]{@{}c@{}}
  \multicolumn{1}{c}{iWildCam} \\[-2pt] \cmidrule(lr){1-1}
  \textcolor{METACOLOR!70!black}{\discrete trap}\\
  \textcolor{METACOLOR!70!black}{\continuous timestamp}
\end{tabular}
&
\begin{tabular}[c]{@{}c@{}}
  \multicolumn{1}{c}{MIMIC-CXR \protect\footnotemark} \\[-2pt] \cmidrule(lr){1-1}
  \textcolor{METACOLOR!70!black}{\continuous age}\\
  \textcolor{METACOLOR!70!black}{\discrete view position}
\end{tabular}
\end{tabular}
\caption[Data with metadata]{Data with metadata\\[-1pt]\makebox[\linewidth][c]{\scriptsize(\raisebox{0.15em}{\discrete}\,discrete, \continuous\,continuous)}}
\end{subfigure}
\end{minipage}%
\hspace{0.4em}
\begin{minipage}[t]{0.49\linewidth}
\vspace{0pt}
\begin{subfigure}[t]{\linewidth}
\centering
\resizebox*{!}{4.1cm}{%
\begin{tikzpicture}[
    baseline=(current bounding box.north),
    x=1cm,y=1cm,
    node distance=6mm and 10mm,
    >=Latex,
    net/.style={
        very thick, draw=MODELCOLOR, fill=MODELCOLOR!20,
        rounded corners=2pt, inner sep=3pt, minimum width=11mm
    },
    loss/.style={
        very thick, ellipse, draw=LOSSCOLOR, fill=LOSSCOLOR!20,
        inner sep=2pt, minimum width=10mm
    },
    img/.style={
        very thick, draw=IMGCOLOR, fill=IMGCOLOR!20,
        inner sep=2pt, minimum width=\wimg cm
    },
    label/.style={
        very thick, draw=LABELCOLOR, fill=LABELCOLOR!20,
        inner sep=2pt, minimum width=\wimg cm
    },
    meta/.style={
        very thick, draw=METACOLOR, fill=METACOLOR!20,
        inner sep=2pt, minimum width=\wimg cm
    }
]
\node[img, anchor=west]  (ra-img)  at (-0.5,0)    {\small images};
\node[meta, anchor=west] (ra-meta) at (-0.5,-1.0) {\small meta-data};

\node[net]  (ra-net)  at (2.5,-0.0) {\small \shortstack{backbone}};
\node[draw=none, below=-1mm of ra-net.south] {\fire};

\node[loss] (ra-ssl)  at (5.5,0)    {\small SSL};
\node[loss] (ra-wsl)  at (5.5,-1.0) {\small WSL};

\draw[very thick,->] (ra-img.east) -- (ra-net.west);
\draw[very thick,->] (ra-net.east) -- (ra-ssl.west);
\draw[very thick,->] (ra-meta.east) -- (ra-wsl.west);
\draw[very thick,->, rounded corners] (ra-net.east) -- ++(0.25,0) |- (ra-wsl.west);

\node[draw=none] at (2.2,-1.5) {Step 1: Representation adaptation};
\node[img, anchor=west]   (lp-img)    at (-0.5,-2.5) {\small images};
\node[label, anchor=west] (lp-label)  at (-0.5,-3.5) {\small labels};

\node[net]   (lp-net)    at (1.6,-2.5) {\small \shortstack{backbone}};
\node[net]   (lp-probe)    at (3.5,-2.5) {\small \shortstack{probe}};
\node[draw=none, below=-0.2mm of lp-net.south] {\snow};
\node[draw=none, below=-1mm of lp-probe.south] {\fire};
\node[loss]  (lp-sup)    at (5.5,-3.0) {\small Sup.};

\node [draw=none] at (2.5,-4.0) {Step 2: Lightweight probing};

\draw[very thick,->,rounded corners=4pt] (lp-img.east) -- (lp-net) -- (lp-probe);
\draw[very thick,->,rounded corners=4pt] (lp-probe.east) -- ++(0.25,0) coordinate (tmp) |- (lp-sup.west);
\draw[very thick,->,rounded corners=4pt] (lp-label.east) -- (tmp |- lp-label.east) |- (lp-sup.west);
\end{tikzpicture}%
}
\vspace{-1.5mm}
\caption{\textsc{FINO}}
\end{subfigure}
\vspace{3mm}%
\begin{subfigure}[t]{\linewidth}
\hspace*{6mm}\begin{tikzpicture}[baseline=(current bounding box.north)]
\begin{axis}[
    height=4.3cm,
    width=.85\linewidth,
    ymin=70, ymax=105,
    ylabel={\scriptsize relative perf (\%)},
    ylabel style={at={(axis description cs:+0.10,0.5)}},
    ymajorgrids,
    grid style={gray!15},
    xtick={1,3,5,7},
    xticklabels={HPA, iWC, FMoW, MIMIC},
    xtick style={draw=none},
    ytick={70,80,90,100},
    enlargelimits=.05,
    tick label style={font=\scriptsize},
    axis x line*=bottom,
    axis y line*=left,
    legend style={
        at={(0.4,1.05)},
        anchor=south,
        legend columns=-1,
        draw=none
    },
    legend image code/.code={
        \draw[#1, fill=#1] (0cm,-0.1cm) rectangle (0.2cm,0.2cm);
    },
]

\addplot[ybar, bar width=4pt, fill=FROZENCOLOR, draw=FROZENCOLOR] coordinates {
    (0.4,71.0) (2.4,98.8) (4.4,79.9) (6.4, 92.9)
};

\addplot[ybar, bar width=4pt, fill=SUPCOLOR, draw=SUPCOLOR] coordinates {
    (0.7,88.6) (2.7,85.4) (4.7,87.1) (6.7,98.8)
};

\addplot[ybar, bar width=4pt, fill=METACOLOR, draw=METACOLOR] coordinates {
    (1.0,103.0) (3.0,101.7) (5.0,102.9) (7.0,100.0)
};

\addplot[thick, SOTACOLOR, forget plot] coordinates {(0,100) (7.2,100)};

\node[draw=none] at (axis cs: 5,105) {\scriptsize \textcolor{SOTACOLOR}{published reference}};

\addplot[ybar, bar width=4pt, fill=METACOLOR, draw=METACOLOR, forget plot] coordinates {
    (1.0,103.0) (3.0,101.7) (5.0,102.9) (7.0,100.0)
};

\legend{\scriptsize Frozen DINOv3, \scriptsize Supervised FT, \scriptsize \textsc{FINO}}

\end{axis}
\end{tikzpicture}
\vspace{-1.5mm}
\caption{Performance relative to published references}
\end{subfigure}
\end{minipage}

%% file: sections/2_related.tex
\paragraph{Self-supervised Representation Learning.}
Self-supervised learning (SSL) has enabled the emergence of powerful visual representations through contrastive~\citep{chen2020simpleframeworkcontrastivelearning, he2020moco}, self-distillation~\citep{caron2021emergingpropertiesselfsupervisedvision,oquab2024dinov2learningrobustvisual}, bootstrapping~\citep{grill2020bootstraplatentnewapproach}, masking~\citep{he2022maskedautoencodersscalablevision}, and joint-embedding predictive approaches~\citep{assran2023selfsupervisedlearningimagesjointembedding}. Several methods incorporate prototype-based objectives to structure representations, either in a fully unsupervised manner~\citep{caron2020unsupervisedlearningvisualfeatures, li2021pcl} or with supervision~\citep{khosla2020supcon}.
A common remedy to mitigate degradation when transferring to specialised domains such as microscopy~\citep{Krishnan2022, doron2023unbiased} or remote sensing~\citep{wang2022selfsupervisedlearningremotesensing} is supervised fine-tuning~\citep{veasey2024parameter,steinertrain,touvron2019fixing,huang2025knowledge}, but it relies on task-specific annotations that are often scarce in scientific settings and may lead to overfitting to spurious correlations~\citep{choi2024autoft,qu2024connect}. This motivates adapting representations using weak and freely available signals such as metadata.

\paragraph{Domain Generalisation and Invariant Representations.}
Batch effects (systematic variations induced by experimental conditions) are a well-documented source of domain shift in scientific data~\citep{leek2010tackling}, and are known to significantly impact learned representations, including in self-supervised settings~\citep{doron2023unbiased, pernice2023out}. A large body of work, therefore, seeks to learn representations invariant to predefined domain labels by explicitly removing domain-specific information. In supervised settings, this is typically achieved through feature alignment across domains~\citep{sun2016correlationalignmentunsuperviseddomain}, gradient reversal mechanisms~\citep{ganin2015unsuperviseddomainadaptationbackpropagation}, contrastive fairness objectives~\citep{shen2021contrastivelearningfairrepresentations}, or compositional attribute decompositions~\citep{mahajan2025compositionalriskminimization}. In the self-supervised setting, \citet{scalbert2024domaininvariantselfsupervisedlearningbatch} propose to standardise Fourier statistics across batches. Complementary to these approaches, post-hoc methods such as Harmony~\citep{korsunsky2019harmony} and Symphony~\citep{kang2021symphony} correct batch effects in latent space.

These approaches treat each selected factor as a nuisance to be removed, and do not provide a mechanism to preserve or encourage informative metadata within the same framework. They also typically assume a small number of discrete domains, whereas scientific metadata often spans thousands of classes or continuous values. In contrast, we propose a unified metadata-guided framework that encourages informative factors while suppressing spurious ones, using prototype-based guidance~\citep{khosla2020supcon} for high-cardinality discrete metadata and learned regressors for continuous factors.

\paragraph{Metadata Guidance in Scientific Imaging.}
Metadata can be viewed as a form of weak supervision for representation learning in domain-specific settings~\citep{zhou2018brief}. In microscopy and related medical imaging, such signals have been exploited through protein identity~\citep{kobayashi2022cytoself}, antibody labels~\citep{gupta2024subcell}, microscopy-specific consistency objectives~\citep{haslum2024metadata}, and patient metadata in retinal imaging~\citep{lee2025preti,holland2024metadata}. These methods show the promise of metadata, but remain tied to specific modalities, task formulations, or supervision regimes.
In remote sensing, SatMAE~\citep{cong2022satmae} and Scale-MAE~\citep{reed2023scalemae} incorporate temporal, spectral, or resolution metadata through modified positional encodings. However, these methods require metadata at test time. Closer to our approach, SatMIPS~\citep{bourcier2024learning} proposes a metadata-guided CLIP-like framework~\citep{radford2021clip} for satellite imagery, but is specifically tailored to continuous geospatial metadata.
Side information extends beyond image-attached acquisition fields. Prior work uses patient identity~\citep{azizi2021micle}, paired radiology reports~\citep{zhang2022convirt}, biological taxonomy~\citep{stevens2024bioclip}, image geolocation~\citep{cepeda2023geoclip}, naturally co-occurring modalities~\citep{girdhar2023imagebind}, and wildlife observations as ecological context for satellite imagery~\citep{daroya2025wildsat}. These signals differ in availability and deployment requirements: \textsc{FINO} focuses on structured metadata available during adaptation and not required at inference. Framing each field as an auxiliary objective is also related to multi-task learning~\citep{caruana1997multitask}; unlike contextual-token methods~\citep{bao2023contextualvisiontransformers}, our framework does not modify the backbone, and unlike language-guided sampling~\citep{el2023learning}, it directly accommodates both discrete and continuous factors at scale.

%% file: sections/3_method.tex
\begin{figure}
    \centering
    \input{figures/pipeline}
\caption{{\bf \textsc{FINO}.} Left: overall representation adaptation pipeline, which combines DINO, iBOT, metadata guidance (informative factors $\bM_+$ encouraged, spurious factors $\bM_-$ suppressed via gradient reversal), and SIGReg to adapt a pretrained encoder without task labels. Right: discrete metadata guidance for one metadata type $t$. The student embedding is contrasted against an EMA prototype bank indexed by the metadata value, and the matched prototype is updated from the teacher embedding. Continuous metadata uses a predictor head instead of a prototype bank.}
    \label{fig:method}
\end{figure}

Our goal is to adapt the representation of a pretrained foundation model to a target application domain without task labels and independently of any downstream task. To this end, we leverage freely available metadata to encourage the model to organise its features along informative factors while reducing sensitivity to spurious ones.

\paragraph{Setting.}
Our approach is based on the standard DINO self-supervised setup\cite{oquab2024dinov2learningrobustvisual}, with a student--teacher architecture operating on augmented views and a teacher updated by exponential moving average. As shown in \cref{fig:method}, we extend this framework to accommodate metadata supervision.
Let $x$ be an image and $\phi$ a backbone encoder. We denote by $\phi(x) \in \mathbb{R}^D$ the student embedding (\eg, the class token of a ViT), and by $\phi_{\text{teacher}}(x)$ the corresponding teacher embedding. Each sample is associated with one or several metadata types $t \in \mathcal{T}$, which may be discrete (\eg, antibody, country) or continuous (\eg, timestamp, viewing angle).

\paragraph{Informative and spurious metadata.}
We partition metadata types into informative factors $\bM_+$, which the representation should preserve, and spurious factors $\bM_-$, which it should suppress. Informative metadata capture meaningful structure in the data distribution and correlates with variables of interest. For example, in fluorescence microscopy, the antibody or staining protocol determines which cellular structures are highlighted and is therefore predictive of protein localisation. In Earth observation, geographic location influences both label distributions (land use, crop types) and their appearance. Organising representations along such factors provides a meaningful supervisory signal.
In contrast, spurious metadata correspond to variations introduced by the acquisition process that are not related to image content. For instance, plate identifiers in high-throughput microscopy often encode batch effects (\eg, illumination or preparation artifacts), while sensor characteristics or viewing conditions in satellite imagery reflect acquisition settings rather than scene content. Representations that encode such factors tend to generalise poorly beyond conditions observed in training \cite{pernice2023out}.

\emph{Choosing metadata factors.} Assigning metadata to $\bM_+$ or $\bM_-$ is application-dependent but does not require target-domain labels. We use prior knowledge, desired deployment invariances, and held-out-validation sweeps when needed; no test metric is used for metadata, checkpoint, or hyperparameter selection. The final configurations encourage antibody on HPA, encourage subregion and suppress year on FMoW, suppress time of day on iWildCam, and encourage view position on MIMIC-CXR. Figure~\ref{fig:compass} evaluates all 15 candidate factors separately under both signs. Most have a clear beneficial sign; HPA cell line is harmful under either sign because it is entangled with antibody identity (App.~\ref{app:cell_line}). Before suppressing correlated fields, we audit their source-only dependence, group redundant fields, and retain one representative at the desired granularity (App.~\ref{app:correlated_factors}).

\begin{figure}[t]
    \centering
    \begin{minipage}{0.52 \linewidth}
    \caption{{\bf Impact of Metadata Guidance.} We measure the impact of adding individual metadata factor when encouraged ($x$-axis, assigned to $\bM_+$) or suppressed ($y$-axis, assigned to $\bM_-$), reported as the $\Delta$ on each dataset's target metric relative to a no-guidance baseline. The quadrants characterise each factor's role: {\textcolor{lbeneficial}{\textit{always beneficial}}} factors help in both regimes; {\textcolor{linformative}{\textit{informative only}}} and {\textcolor{lspurious}{\textit{spurious only}}} factors help only under correct assignment; {\textcolor{lharmful}{\textit{always harmful}}} factors degrade performance regardless of assignment, occupied only by cell line on HPA, an entanglement artefact analysed in \ref{app:cell_line}.  The dashed diagonal separates factors that benefit more from encouragement (below) vs suppression (above). See \ref{app:metadata_analysis} for a detailed per-factor analysis.}
    \label{fig:compass}
    \end{minipage}
    \hspace{1em}
    \begin{minipage}{0.43\linewidth}
     \input{figures/compass}
    \end{minipage}
\end{figure}

\paragraph{Metadata guidance.}
We define a guidance loss $\cL_{\text{meta}}^{(t)}$ for each metadata type $t$. Its form depends on whether the metadata is discrete or continuous, and its sign determines whether it encourages alignment with informative factors or suppresses spurious ones via reversed gradients.

\emph{Discrete metadata.}
For a discrete metadata type $t$ taking values in $\{1, \dots, M\}$, we maintain a prototype bank $\{p_m^{t}\}_{m=1}^{M}$, where each prototype corresponds to one possible metadata value. For an image $x$ with value $m$ for metadata $t$, we define the following prototype-based contrastive objective, inspired by supervised contrastive learning~\citep{khosla2020supcon} and prototypical contrastive learning~\citep{li2021pcl}:
\begin{align}
    \cL_{\text{meta}}^{(t)}(x)
    &=
    -
    \log
    \frac{
        \exp\left( \langle \phi(x), p^{t}_{m} \rangle / \tau \right)
    }{
        \sum_{n=1}^{M}
        \exp\left( \langle \phi(x), p^{t}_{n} \rangle / \tau \right)
    }~,
\end{align}
where $\tau>0$ is a temperature parameter, $\phi(x)$ and $p^{t}_{m}$ are $\ell_2$-normalised so that $\langle \cdot, \cdot \rangle$ is a cosine similarity (see App.~\ref{app:training}). Unlike standard supervised contrastive learning, no positive or negative samples are needed within the batch, allowing it to scale to metadata with large numbers of classes $M$.
Prototypes are updated via an exponential moving average of teacher embeddings:
\begin{align}
    p^{t}_{m}
    &\leftarrow
    \alpha p^{t}_{m}
    +
    \left( 1 - \alpha \right)\, {\phi}_{\text{teacher}}(x)~,
\end{align}
where $\alpha$ is the EMA update rate; $p^{t}_{m}$ is detached and re-projected onto the unit sphere after each update. The approach is illustrated in Fig.~\ref{fig:method}.

\emph{Continuous metadata.}
For a metadata $t$ with continuous values $m \in \mathbb{R}^{d}$, we learn a predictor $g^{(t)}$ (\eg, an MLP) and define a regression loss:
\begin{align}
    \cL_{\text{meta}}^{(t)}(x)
    &=
    \left\| g^{(t)}\left( \phi(x) \right) - m \right\|_2^2~.
\end{align}

\emph{Encouraging and suppressing metadata.}
We use the same metadata loss for all metadata types, but control its effect on the encoder via gradient reversal. For informative metadata $t \in \bM_+$, the encoder minimises $\cL_{\text{meta}}^{(t)}$, encouraging alignment with the corresponding factor. For spurious metadata $t \in \bM_-$, we reverse the gradient before it reaches the encoder, effectively discouraging the representation from encoding this information.
When the guidance branch includes trainable parameters (\eg, the predictor $g^{(t)}$), they are always optimised to minimise $\cL_{\text{meta}}^{(t)}$, following the standard adversarial formulation of~\citet{ganin2015unsuperviseddomainadaptationbackpropagation}.

\emph{Balancing metadata losses.}
Discrete and continuous metadata losses have different scales and gradient profiles, which can hinder joint training. When multiple metadata branches are used jointly, we therefore balance their gradients with a simple EMA-smoothed multiplicative factor, in the spirit of prior gradient-balancing methods~\citep{chen2018gradnorm,liu2021imtl,javaloy2022rotograd}. Full details are provided in App.~\ref{app:grad_equal}.

\paragraph{Overall objective.}
We build on a standard self-supervised setup combining the DINO and iBOT losses respectively on the CLS and patch tokens. Following LeJEPA~\citep{balestriero2025lejepa}, we additionally regularise the pre-normalisation DINO bottleneck with SIGReg (see App.~\ref{app:sigreg}). The training objective is
\begin{align}
\cL(x)
&=
\cL_{\text{DINO}}(x)
+
\cL_{\text{iBOT}}(x) %
+
\lambda_{\text{SIGReg}}\cL_{\text{SIGReg}}(x)
+
\sum_{t \in \bM_+ \cup \bM_-} \lambda_{\text{meta}}^{(t)} \cL_{\text{meta}}^{(t)}(x)~,
\end{align}
where $\{\lambda_{\text{meta}}^{(t)}\}_{t}$, and $\lambda_{\text{SIGReg}}$ are nonnegative weights. In practice these weights are not tuned: $\lambda_{\text{SIGReg}}$ is taken as-is from the LeJEPA paper~\citep{balestriero2025lejepa}, and a single value $\lambda_{\text{meta}}^{(t)}{=}0.03$ is reused for every metadata branch across all our application domains (see App.~\ref{app:training}), meaning the method can be applied to a new dataset without an extra hyperparameter sweep. For $t \in \bM_-$, gradients from $\cL_{\text{meta}}^{(t)}$ are reversed before reaching the encoder. See \cref{fig:algo} for a pseudo-code implementation.

\paragraph{Implementation details.}
We initialise the backbone from an open-source DINOv3~\citep{simeoni2025dinov3} ViT-L checkpoint obtained by distilling a larger 7B-parameter teacher. As this checkpoint provides only backbone weights, we initialise all heads (DINO, iBOT, and metadata) from random weights.%
Adaptation is one unlabeled job with two regimes. First, the randomly initialised DINO, iBOT, and metadata heads and the patch embedding are trained while the transformer blocks remain frozen. The backbone is then unfrozen with a linear learning-rate warm-up and the full model is trained end-to-end. The frozen downstream probe is evaluation rather than a representation-training phase. This schedule prevents random heads from destabilising the pretrained backbone; App.~\ref{app:training} gives the full protocol and its iWildCam exception.

%% file: figures/pipeline.tex
\def\wtoken{0.4}%
\definecolor{MODELCOLOR}{RGB}{145,95,235}  %
\definecolor{TOKENCOLOR}{RGB}{245,170,55}  %
\definecolor{POSCOLOR}{RGB}{60,130,230}    %
\definecolor{NEGCOLOR}{RGB}{220,70,70}     %
\definecolor{LOSSCOLOR}{RGB}{230,80,180}   %
\begin{tabular}{@{}c@{\;\;}c@{}}
\begin{subfigure}{0.53\linewidth}
\centering
\resizebox{\linewidth}{!}{
\input{figures/high-level-pipeline}
}
\caption{Metadata-Driven Representation Adaptation}
\end{subfigure}
&
\begin{subfigure}{0.45\linewidth}
\centering
\resizebox{\linewidth}{!}{
\input{figures/meta-guidance}
}
\caption{Discrete metadata guidance for metadata type $t$}
\end{subfigure}
\end{tabular}

%% file: figures/high-level-pipeline.tex
\def\xinput{0}%
\def\xloss{3.0}%
\def\xlossmeta{4.0}%
\def\ymain{0}%
\def\yloss{1.0}%
\def\ypos{1.8}%
\def\yneg{-1.4}%
\def\wimg{1.9}%
\def\step{0.5}%
\def\wcircle{0.35}%
\hspace{-3mm}
\begin{tikzpicture}
[
module/.style={draw=MODELCOLOR, fill=MODELCOLOR!20, very thick, trapezium, text width=0.7cm, rotate=-90,trapezium angle=70,text centered, inner sep=0pt, outer sep=0pt},
token/.style={draw=TOKENCOLOR, fill=TOKENCOLOR!20, very thick, minimum width=\wtoken cm, minimum height=\wtoken cm,  inner sep=1pt},
metatoken/.style={regular polygon, regular polygon sides=3, very thick, scale=0.7},
meta/.style={rounded corners, text width=3.5cm, align=center, very thick},
loss/.style={very thick, ellipse , inner sep=1pt},
round/.style={draw=black, fill=white, circle, minimum width=\wcircle cm, minimum height=\wcircle cm, very thick},
]
    \node[draw=none] (input) at (\xinput, \ymain) {\includegraphics[width=\wimg cm, height=\wimg cm]{images/HPAWhole.jpg}};

    \node[metatoken, draw=POSCOLOR, fill=POSCOLOR!20] (metaposone) at (\xinput-0.5, \ypos) {};
    \node[metatoken, draw=POSCOLOR, fill=POSCOLOR!20] (metapostwo) at (\xinput, \ypos) {};
    \node[metatoken, draw=POSCOLOR, fill=POSCOLOR!20] (metapos) at (\xinput+0.5, \ypos) {};

    \node[meta, draw=none, below=-1mm of metapostwo] (metatxt) {\footnotesize\begin{tabular}{c}
         \bf Informative meta-data \\[-0.5mm]
         location, antibody, etc
    \end{tabular}};

    \node[metatoken, draw=none, fill=none] (metanegtwo) at (\xinput, \yneg) {};
    \node[metatoken, draw=NEGCOLOR, fill=NEGCOLOR!20] (metanegone) at (\xinput-0.25, \yneg) {};
    \node[metatoken, draw=NEGCOLOR, fill=NEGCOLOR!20] (metaneg) at (\xinput+0.25, \yneg) {};

     \node[meta, draw=none, below=-1.0mm of metanegtwo]  {\footnotesize \begin{tabular}{c}
         \bf Spurious meta-data \\[-0.5mm]
         plate, resolution, etc
    \end{tabular}};

    \node[module, right=\step cm of input.east, anchor=south] (encoder) {\rotatebox{90}{\scriptsize \shortstack{Encoder}}};

    \node[token] (token) at (\xlossmeta, \ymain) {};
    \node[draw=none, right=0mm of token] {\footnotesize \texttt{CLS}};

    \coordinate (crossroad) at (\xlossmeta, \ymain);

    \node[loss, draw=LOSSCOLOR, fill=LOSSCOLOR!20] at (\xloss, \yloss) (dinoloss)  {\shortstack{SSL \\ loss}};

    \node[loss, draw=POSCOLOR, fill=POSCOLOR!20] at (\xlossmeta, \ypos) (losspos)  {\footnotesize Meta-guidance};

    \node[loss, draw=NEGCOLOR, fill=NEGCOLOR!20] at (\xlossmeta, \yneg) (lossneg)  {\footnotesize Meta-guidance};

    \node[round, above=0.35cm of lossneg]  (grl)  {};
    \draw [very thick, black] (grl.west) ++ (+0.1,0) -- ($(grl.east) + (-0.1,0)$);

    \node[draw=none, right=0cm of grl, text=black] {\footnotesize \shortstack{gradient\\reversal}};
    \if 1 0

    \node[loss] at (\xloss, \ypos) (losspos)  {InfoNCE};

    \node[draw=none] at (\xbank, \yposbank) (bankpos)  {
    \begin{tikzpicture}
    \foreach \i in {1,...,\numbank} {
   \pgfmathsetmacro{\ip}{(\i-1) * \wtoken * 1.2}
   \defrandomcolor{tmpcolor}
    \node[token, fill=tmpcolor!20, draw=tmpcolor] at (\ip,0) {};
}
    \end{tikzpicture}
    };

    \node [rounded corners, draw=black] (emapos) at (crossroad |- bankpos) {\scriptsize \shortstack{domain-wise \\ EMA}};

    \node[loss] at (\xbank, \yneg) (lossneg)  {InfoNCE};

    \node[round, left=0.5cm of lossneg]  (grl)  {};
    \draw [very thick, black] (grl.west) ++ (+0.1,0) -- ($(grl.east) + (-0.1,0)$);

    \node[draw=none] at (\xbank, \ynegbank) (bankneg)  {
    \begin{tikzpicture}
    \foreach \i in {1,...,\numbank} {
   \pgfmathsetmacro{\ip}{(\i-1) * \wtoken * 1.2}
    \defrandomcolor{tmpcolor}
    \node[token, fill=tmpcolor!20, draw=tmpcolor] at (\ip,0) {};
   }
    \end{tikzpicture}
    };

    \node [rounded corners, draw=black] (emaneg) at (crossroad |- bankneg) {\scriptsize \shortstack{domain-wise \\ EMA}};

    \fi

    \draw [very thick, ->, MODELCOLOR] (input) -- (encoder) -- (token);
     \draw [very thick, ->, MODELCOLOR] (encoder.north) ++ (0,0.25) -| (dinoloss);

    \draw [very thick, ->, POSCOLOR] (token) -- (losspos);

    \draw [very thick, ->, NEGCOLOR] (token) -- (grl) -- (lossneg);

    \draw [very thick, ->, POSCOLOR] (metapos) -- (losspos);
    \draw [very thick, ->, NEGCOLOR] (metaneg) -- (lossneg);

\end{tikzpicture}

%% file: figures/meta-guidance.tex
\def\xinput{0}%
\def\xtoken{2.35}%
\def\xloss{3.0}%
\def\xlegend{2.75}%
\def\ytoken{1}%
\def\ymeta{-0.0}%
\def\bankup{0.65}%
\def\emaup{2.0}%
\def\ylegend{4.25}%
\def\numbank{5}%
\def\chosenmeta{3}%
\def\cellscale{1.5}%
\def\wcircle{0.35}%
\definecolor{LOSSCOLOR}{RGB}{255,150,40}%
\definecolor{TEACHERCOLOR}{RGB}{25,200,90}%
\begin{tikzpicture}
[
loss/.style={very thick, ellipse , inner sep=1pt, draw=LOSSCOLOR, fill=LOSSCOLOR!20},
token/.style={very thick, minimum width=\wtoken cm, minimum height=\wtoken cm,  inner sep=1pt},
grid/.style={ minimum width=\wtoken*\cellscale cm, minimum height=\wtoken*\cellscale cm, inner sep=0pt},
round/.style={draw=black, fill=white, circle, minimum width=\wcircle cm, minimum height=\wcircle cm, very thick},
metatoken/.style={regular polygon, regular polygon sides=3, very thick, scale=0.7},
]

\node[metatoken, fill=black!10, draw=black] (meta) at (\xinput, \ymeta) {};

\node[token, draw=TOKENCOLOR, fill=TOKENCOLOR!20] (studenttoken) at (\xinput, \ytoken) {};

\begin{scope}[shift={(\xloss,\ytoken)}]
    \foreach \i in {1,...,\numbank} {
    \pgfmathsetmacro{\ip}{(\i-1) * \wtoken * \cellscale}
    \expandafter\defrandomcolor{tmpcolor}
    \ifnum\i=\chosenmeta
        \node[grid, fill=black, draw=black, text=white] (chosencell) at (\ip,0) {\Large +};
        \node[token, fill=tmpcolor!20, draw=tmpcolor] (chosentoken) at (\ip,\bankup) {\scriptsize $p^t_{m}$};
    \else
        \ifnum\i=1
        \node[token, fill=tmpcolor!20, draw=tmpcolor] at (\ip,\bankup) {\scriptsize $p^t_{1}$};
            \node[grid, fill=white, draw=black, text height=1.5ex, text depth=.25ex, inner sep=0pt,  align=center] (firstcell) at (\ip,0) {\Large -};
        \else
            \ifnum\i=\numbank
            \node[token, fill=tmpcolor!20, draw=tmpcolor] at (\ip,\bankup) {\scriptsize $p^t_{K}$};
            \node[grid, fill=white, draw=black, text height=1.5ex, text depth=.25ex, inner sep=0pt,  align=center] at (\ip,0) {\Large -};
            \else
                \node[token, fill=tmpcolor!20, draw=tmpcolor] at (\ip,\bankup) {\tiny $\cdots$};
                \node[grid, fill=white, draw=black, text height=1.5ex, text depth=.25ex, inner sep=0pt,  align=center] at (\ip,0) {\Large -};
            \fi
        \fi
    \fi
   }
\end{scope}

\node[draw=none, below right=0mm and 0mm of chosencell] {\footnotesize InfoNCE};
\node[draw=none, above right=0mm and 0mm of chosentoken] {\footnotesize \shortstack{Prototype\\bank}};

 \node [draw=black!10] at (\xlegend, \ylegend)
 {
 \begin{tabular}{rl}
    \tikz[baseline=-.6ex, scale=1]
    {\node[grid, fill=black, text=white, draw=black, minimum width=\wtoken*0.9 cm, minimum height=\wtoken*0.9 cm] {+};}
    /
    \tikz[baseline=-.6ex, scale=1]
    {\node[grid, fill=white, draw=black, minimum width=\wtoken*0.9 cm, minimum height=\wtoken*0.9 cm, text height=1.5ex, text depth=.25ex, inner sep=0pt,  align=center] {\Large -};} &
    \footnotesize positive/negative match in InfoNCE
\end{tabular}
};

\node [rounded corners, draw=black, very thick, above=\emaup cm of chosencell] (ema) {\scriptsize \shortstack{EMA update}};
\node[token, fill=TEACHERCOLOR!10, draw=TEACHERCOLOR]  (teachertoken) at (studenttoken |- ema) {};
\draw[very thick, ->] (studenttoken) -- node[pos=-0, above, anchor=south west] {\footnotesize \shortstack{student \texttt{CLS} token}} (firstcell);

\draw[very thick, ->] (meta) -| node[pos=-0, above, anchor=south west] {\footnotesize meta-data value $m$} (chosencell);

\draw [dashed, very thick, ->] (teachertoken) -- node[pos=0.0, below, anchor=north west] {\footnotesize \shortstack{teacher \texttt{CLS} token}} (ema) -- (chosentoken) node[pos=0.5, left, anchor=east] {\footnotesize $m$-th prototype};

\end{tikzpicture}

%% file: figures/compass.tex
\definecolor{HPACOLOR}{HTML}{E11D48}   %
\definecolor{WILDCOLOR}{HTML}{1D4ED8}  %
\definecolor{MIMICCOLOR}{HTML}{0D9488} %
\definecolor{FMOWCOLOR}{HTML}{047857}  %
\begin{tikzpicture}
\begin{axis}[
    width=1\linewidth,
    height=0.9\linewidth,
    xmin=-2, xmax=2,
    ymin=-2, ymax=2,
    xtick={-2,0,+2},
    xticklabels={-2,0,+2},
    ytick={-2,0,+2},
    yticklabels={-2,0,+2},
    enlargelimits=false,
    clip=false,
    xlabel={$\Delta$ metric when encouraged},
    ylabel={$\Delta$ metric when suppressed},
    xlabel style={at={(axis description cs:0.5,0.05)}, anchor=north},
    ylabel style={at={(axis description cs:0.2,0.5)}, anchor=south},
    legend style={
        at={(0.5,-0.2)},
        anchor=north,
        font=\scriptsize,
        draw=black!15,
        fill=white,
        fill opacity=0.85,
        text opacity=1,
        rounded corners=1pt,
    },
    legend columns=-1,
]

\fill[qbeneficial!40]   (axis cs:0,0)   rectangle (axis cs:2,2);
\fill[qspurious!40]     (axis cs:-2,0)  rectangle (axis cs:0,2);
\fill[qinformative!40]  (axis cs:0,-2)  rectangle (axis cs:2,0);
\fill[qharmful!40]      (axis cs:-2,-2) rectangle (axis cs:0,0);

\draw[->, black!20] (axis cs:0,-2) -- (axis cs:0,2);
\draw[->, black!20] (axis cs:-2,0) -- (axis cs:2,0);

\draw[dashed, black!45] (axis cs:-2,-2) -- (axis cs:2,2);

\fill[fill=gray!5, draw=black!20] (axis cs:0,0) circle[radius=10];

\node[
    font=\small,
    fill=qspurious!40!white,
    rounded corners=2pt,
    inner sep=1.5pt, anchor=north west
] at (axis cs:-1.96,1.96) {spurious only};

\node[
    font=\small,
    fill=qbeneficial!20!white,
    rounded corners=2pt,
    inner sep=1.5pt, anchor=north east
] at (axis cs:1.96,1.96) {\shortstack{always\\beneficial}};

\node[
    font=\small,
    fill=qinformative!20!white,
    rounded corners=2pt,
    inner sep=1.5pt, anchor=south east
] at (axis cs:1.96,-1.96) {informative only};

\node[
    font=\small,
    fill=qharmful!20!white,
    rounded corners=2pt,
    inner sep=1.5pt, anchor=south west
] at (axis cs:-1.96,-1.96) {always harmful};

\addplot[
    only marks,
    mark=pentagon,
    mark size=3pt,
    FMOWCOLOR,
    thick
] coordinates {
    (2.0,0.7)     %
    (-0.3,0.4)    %
    (0.6,-0.1)    %
};
\addplot[
    only marks,
    mark=pentagon*,
    mark size=3pt,
    FMOWCOLOR,
    thick
] coordinates {
    (-0.5,0.3)    %
    (-0.4,0.4)    %
    (0.9,-0.6)    %
    (0.7,-1.4)    %
};

\addplot[
    only marks,
    mark=diamond,
    mark size=3pt,
    WILDCOLOR,
    thick
] coordinates {
    (1.27,0.00)    %
    (-0.80,0.46)   %
};
\addplot[
    only marks,
    mark=diamond*,
    mark size=3pt,
    WILDCOLOR,
    thick
] coordinates {
    (-1.62,1.39)   %
};

\addplot[
    only marks,
    mark=square,
    mark size=3pt,
    HPACOLOR,
    thick
] coordinates {
    (1.80,-0.96)    %
    (-1.30,-0.52)   %
    (0.31,-0.51)    %
};

\addplot[
    only marks,
    mark=triangle,
    mark size=3.5pt,
    MIMICCOLOR,
    thick
] coordinates {
    (0.6,-0.2)     %
};
\addplot[
    only marks,
    mark=triangle*,
    mark size=3.5pt,
    MIMICCOLOR,
    thick
] coordinates {
    (0.7,0.1)      %
};

\node[font=\scriptsize, anchor=east]       at (axis cs:2.0,0.7)   {subregion};
\node[font=\scriptsize, anchor=east] at (axis cs:0.6,-0.1)  {month};
\node[font=\scriptsize, anchor=west] at (axis cs:0.72,-1.4)  {time of day};
\node[font=\scriptsize, anchor=north] at (axis cs:0.9,-0.6)  {sun elev.};
\node[font=\scriptsize, anchor=west]       at (axis cs:1.27,0.00) {trap ID};
\node[font=\scriptsize, anchor=east] at (axis cs:-0.80,0.46) {month};
\node[font=\scriptsize, anchor=west] at (axis cs:-1.62,1.39) {time of day};
\node[font=\scriptsize, anchor=east, yshift=-0.5mm] at (axis cs:1.80,-0.96)  {antibody};
\node[font=\scriptsize, anchor=south, yshift=0.5mm] at (axis cs:-1.30,-0.50) {cell line};
\node[font=\scriptsize, anchor=east, xshift=-0.5mm] at (axis cs:0.31,-0.51)  {plate};
\node[font=\scriptsize, anchor=south] at (axis cs:0.7,0.1)    {age};
\node[font=\scriptsize, anchor=north, yshift=-0.5mm] at (axis cs:0.6,-0.2)    {view};

\node[font=\scriptsize, anchor=west] at (axis cs:-0.25,0.45) {year};
\node[font=\scriptsize, anchor=south, xshift=-2mm, yshift=0.5mm] at (axis cs:-0.4,0.4) {\shortstack{reso-\\[-1mm]lution}};
\node[font=\scriptsize, anchor=north] at (axis cs:-0.55,0.3) {coords};

\end{axis}
\end{tikzpicture}
\vspace{-2mm}
\begin{center}
\scriptsize \begin{tabular}{@{}r@{\,:\,}l@{\;\;}r@{\,:\,}l@{\;\;}r@{\,:\,}l@{\;\;}r@{\,:\,}l@{\;\;} r@{\,:\,}l@{\;\;} r@{\,:\,}l@{}}
    \tikz[baseline=-0.5ex]{\draw [thick, black] (0,0) circle (0.8mm);}
     & \tiny discrete
     &
     \tikz[baseline=-0.5ex]{\fill [thick, black] (0,0) circle (0.8mm);}
     & \tiny continuous
     &
     \tikz[baseline=-0.5ex]{
  \draw[thick, HPACOLOR, fill=HPACOLOR]
    plot[mark=square*, mark size=1.5pt] coordinates {(0,0)};
}
     & \tiny HPA
     &
    \tikz[baseline=-0.5ex]{
  \draw[thick, WILDCOLOR, fill=WILDCOLOR]
    plot[mark=diamond*, mark size=1.5pt] coordinates {(0,0)};
}
     & \tiny iWCam
     &
    \tikz[baseline=-0.5ex]{
  \draw[thick, FMOWCOLOR, fill=FMOWCOLOR]
    plot[mark=pentagon*, mark size=1.6pt] coordinates {(0,0)};
}
     & \tiny FMoW
      &
     \tikz[baseline=-0.5ex]{
  \draw[thick, MIMICCOLOR, fill=MIMICCOLOR]
    plot[mark=triangle*, mark size=1.7pt] coordinates {(0,0)};
}
     & \tiny MIMIC
\end{tabular}
\end{center}

%% file: sections/4_experiments.tex
\subsection{Datasets and Evaluation}
\label{sec:datasets}

\paragraph{Datasets.}
We evaluate our approach on four datasets from distinct scientific domains: fluorescence microscopy with Human Protein Atlas (HPA)~\citep{thul2017subcellular,thul2018hpa,ouyang2019hpa}, Earth observation with Functional Map of the World (FMoW)~\citep{christie2018fmow,koh2021wildsbenchmarkinthewilddistribution}, wildlife monitoring with iWildCam~\citep{beery2020iwildcam}, and chest X-ray analysis with MIMIC-CXR~\citep{johnson2019mimic}.
Additionally, we evaluate cross-dataset transfer within the same application domain with three additional datasets: OpenCell ~\citep{cho2022opencell}, CheXpert~\citep{irvin2019chexpert}, and FLAIR-Hub~\citep {garioud2026flair}. Detailed descriptions of all datasets, splits, metadata, and evaluation protocols are provided in \cref{app:datasets}.

\paragraph{Evaluation Protocol.}
We use the same ViT-L backbone (306M parameters) for all methods. Unless specified otherwise, models are initialised from a public DINOv3 checkpoint~\cite{simeoni2025dinov3}; see Appendix \cref{app:siglip} for a study of generalisation to other backbones. We compare our approach to two classes of methods: task-centric domain adaptation and representation adaptation.
For task-centric methods, the model is trained using labelled data from the training set, optionally combined with robustness-inducing regularisation losses. For unsupervised domain adaptation (UDA), we additionally leverage unlabelled data from the test set. In both cases, we follow the standard protocol of fully fine-tuning all model parameters; note that this involves learning a task-specific prediction head.
In contrast, in \emph{representation adaptation} we aim to improve the intrinsic quality of learned representations independent of a specific target task or dataset. We first adapt model weights using unlabelled source data and, optionally, metadata. The model is then frozen, and a lightweight attentive~\citep{chen2024context} or linear probe is trained on top. The linear probe concatenates the CLS token with the average-pooled patch embeddings and applies a single linear layer. The attentive probe uses a single cross-attention layer that attends to patch embeddings from the last four transformer blocks and aggregates them into a single representation. Both probes remain lightweight (2.7M parameters for the attentive probe; under 1M for the linear probe) compared to a full fine-tuning of the model.

\subsection{Results and Analysis}
\label{sec:results_analysis}
\begin{table}[tbh]
\centering
\caption{{\bf Quantitative results.} Matched adaptations and published references: \citet{ouyang2019hpa} for HPA, \citet{choi2024autoft} for iWildCam/FMoW, and \citet{moutakanni2024advancinghumancentricairobust} for MIMIC-CXR. Published references report mean$\pm$standard deviation where available. \textsc{FINO} reports mean$\pm$standard deviation over 8, 5, and 5 complete runs on iWildCam, FMoW, and MIMIC. $^{*}$HPA is the full-high-resolution run; App.~\ref{app:additional_robustness} reports its low-resolution repeat. ``Param.'' counts parameters trained with labels; \bluesnow/\orangefire{} denote frozen/trainable backbones. WGA is worst-group accuracy.}
\label{tab:main_extended}
\input{tables/main_results2}
\end{table}

We compare our approach to existing adaptation methods across several application domains, analyse the role of different metadata factors and their effect on the learned representation space, and further evaluate the method in cross-dataset evaluation and low-supervision settings.

\paragraph{Adaptation performance.}
We report the performance of all adaptation methods in \cref{tab:main_extended}. First, training a ViT from scratch is generally not viable on the considered datasets. Second, despite the domain gap between its pretraining data and our target domains, fine-tuning a pretrained DINOv3 backbone yields inconsistent gains over probing the frozen model, and even degrades performance on iWildCam. Unsupervised domain adaptation methods  yield inconsistent gains over supervised fine-tuning: substantially improving on FMoW but underperforming on iWildCam, with broadly comparable performance on HPA and MIMIC, with the exception of CORAL on HPA. Applying the standard DINO self-supervised adaptation recipe generally outperforms supervised fine-tuning, but its gains remain inconsistent, even underperforming the frozen DINO baseline on iWildCam.

In contrast, \textsc{FINO} is the strongest evaluated method under the matched backbone, data, and evaluation protocol. The complete-run means are $52.9{\pm}0.8$ on iWildCam, $53.3{\pm}0.8$ on FMoW, and $81.7{\pm}0.1$ on MIMIC; the full-resolution HPA result is $61.2$. All configurations, metadata assignments (\cref{tab:selection_ledger}), checkpoints, and probe hyperparameters were selected on held-out validation data; neither HPA Kaggle public nor private scores informed selection. Against the listed published references, \textsc{FINO} reports higher scores on HPA, iWildCam, and FMoW and matches MIMIC under differing upstream data, supervision, resolution, search, and inference protocols. The objective and loss weights are shared across domains, with dataset-specific metadata/evaluation and the iWildCam schedule exception in App.~\ref{app:training}.

\begin{table}[t]
\centering
\caption{{\bf Final metadata assignments and selection signals.} A dash denotes no factor in that branch. Signs, checkpoints, and probe hyperparameters were chosen with the listed held-out metric; no test-set labels or test metrics were used for selection.}
\label{tab:selection_ledger}
\input{tables/selection_ledger}
\end{table}

\paragraph{Which metadata helps?}
We report in \cref{fig:compass} the impact of each metadata factor when treated as either informative or spurious.
Some metadata are consistently informative: using them as supervisory signals improves performance, indicating that they capture meaningful structure aligned with downstream tasks.
In contrast, some metadata are purely spurious: performance improves only when their influence is suppressed, suggesting that they encode nuisance variations that hinder generalisation.
Interestingly, certain metadata are beneficial in both regimes: both predicting and actively suppressing these factors provide useful training signals, acting as auxiliary objectives that steer the self-supervised loss toward more semantic representations.
A few metadata factors can be harmful in both regimes because they entangle task-relevant and acquisition structure; App.~\ref{app:metadata_analysis} gives the full factor-level analysis.

\begin{table}[t]
\centering
\caption{{\bf All-positive metadata guidance without sign selection.} Every available metadata field is encouraged, with no informative/spurious selection. HPA uses the low-resolution configuration.}
\label{tab:all_positive}
\input{tables/all_positive}
\end{table}

\paragraph{Is metadata selection necessary?}
As a no-selection stress test, \cref{tab:all_positive} encourages every available metadata field. All-positive guidance matches or exceeds unguided SSL on all three tested datasets. It matches the curated configuration on low-resolution HPA and is within $0.2$ points on iWildCam; on FMoW, curation improves WGA by $4.9$ points when the 13 available factors include overlapping geographic variables. All-positive guidance is therefore a robust no-selection baseline, while curation pays off for highly redundant metadata.

\begin{table}[t]
\centering
\caption{{\bf Metadata versus task-label guidance.} Same adaptation and evaluation; only the guidance signal changes. Differences are relative to SSL adaptation without guidance. Metadata alone is best on both datasets, and adding task labels to metadata hurts. HPA uses the low-resolution configuration; full IID/OOD results and the multi-label HPA setup are in App.~\ref{app:additional_robustness}.}
\label{tab:task_guidance}
\input{tables/task_guidance}
\end{table}

\paragraph{Task-label guidance.}
If downstream task labels are available, should they guide adaptation instead of metadata? Since they define the evaluated task, one would expect them to guide better. We test this alongside our DINO and iBOT objectives, keeping all other evaluation protocols fixed. We compare metadata-only \textsc{FINO} with conventional SupCon~\citep{khosla2020supcon}, task labels used through \textsc{FINO}'s prototype objective, and SupCon combined with the selected metadata. For multi-label HPA, SupCon positives share at least one active label, while prototype guidance updates every active class prototype. We find that metadata guidance not only competes with task-label guidance, but actually outperforms it on both FMoW and HPA. Adding SupCon to metadata also does not improve over metadata guidance alone. This suggests that the metadata framework provides a real robustness benefit over using downstream labels, even when those labels are available during adaptation.

\begin{figure}[t]
    \centering
    \input{figures/umap}

    \caption{{\bf HPA representation geometry.} UMAPs of 3,000 validation images with a single protein-location label, coloured by cell line (top) or protein localisation (bottom). The displayed k-NN scores use $k{=}100$ for cell line and $k{=}10$ for localisation. \textsc{FINO}, guided only by antibody identity, strengthens target structure while attenuating cell-line structure without directly suppressing cell line.}
    \label{fig:umap}
\end{figure}

\paragraph{Representation geometry.}
\cref{fig:umap} visualises the HPA feature space. The off-the-shelf DINOv3 representation is poorly organised by protein localisation (4\% k-NN). Fully supervised fine-tuning produces label-aligned clusters but generalises poorly to the OOD test set (48\%). Self-supervised adaptation remains strongly structured by cell line (87\%). Antibody-guided \textsc{FINO} raises protein-localisation k-NN accuracy to 83\% and reduces cell-line accuracy to 75\%, despite never using protein-location labels during adaptation or explicitly suppressing cell line.

\begin{table}[t]
    \centering
    \caption{{\bf Transfer learning.} A DINOv3 model is adapted on HPA, FMoW, or MIMIC-CXR and then frozen, probed, and evaluated on OpenCell, FLAIR-Hub, or CheXpert, respectively.}
    \label{tab:transfert}
    \input{tables/transfert}
\end{table}

\paragraph{Low-supervision regime.}
As shown in \cref{fig:lowsup}, \textsc{FINO} consistently outperforms all baselines across all label fractions, with particularly large gains in the low-data regime. When only 1--10\% of labels are available, performance remains high (50.9--57.0\%), while supervised fine-tuning and Group DRO degrade sharply (\eg, 53.3\% $\rightarrow$ 29.3\% and 52.6\% $\rightarrow$ 25.7\% at 1\%).
Notably, \textsc{FINO} also surpasses fully supervised fine-tuning even at 100\% of labels, indicating that label-free representation adaptation can produce features that are both more robust and better aligned with the task. In contrast, Group DRO provides limited gains and remains below standard supervised training, suggesting that robustness to predefined groups is insufficient in this setting.

\paragraph{Transfer learning.}
A key advantage of label-free adaptation is that the representation does not collapse around a fixed label set. This improves robustness to domain shift and enables transfer across datasets within the same application domain but with different class definitions. In \cref{tab:transfert}, we adapt a generic DINOv3 model using data and metadata from one dataset, then probe and evaluate on a different dataset from the same domain (\cref{fig:datasets} summarises the structural differences between each transfer pair). When input channels differ between datasets, we drop unsupported input channels or their corresponding model weights, \eg, the nIR band in FLAIR-Hub or the microtubule and ER channels absent from OpenCell. For OpenCell, we report the Adjusted Rand Index (ARI) between protein-level embeddings and ground-truth localisation clusters, following the protocol of \citet{kobayashi2022cytoself}. We observe that supervised adaptation is either stable or improves over a frozen DINOv3 backbone, while \textsc{FINO} consistently improves over both. On \textsc{OpenCell}, the off-the-shelf DINOv3 backbone alone already exceeds prior published methods; \textsc{FINO} adds a further $+0.122$ ARI on top, despite never being trained on OpenCell data (\cref{app:transfer}).

\begin{figure}[t]
\centering
\begin{minipage}{0.5\linewidth}
\caption{{\bf Low-supervision regime.} F1 score on HPA as a function of the fraction of labelled data (out of 94,270 annotated images) used to train the downstream predictor. We compare \textsc{FINO}, supervised fine-tuning, and Group DRO (using cell lines as groups). Results use a single fold without cross-validation, so absolute scores differ from \cref{tab:main_extended}. Each point is averaged over 5 seeds (stratified label samplings), except at 100\% where the full label set is deterministic. Shaded areas indicate the standard deviation across these seeds.}
\label{fig:lowsup}
\end{minipage}
\hspace{1em}
\begin{minipage}{0.45\linewidth}
\resizebox{\linewidth}{!}{\input{figures/lowsup}}
\end{minipage}
\end{figure}

\subsection{Ablation Study}
\label{sec:ablation}
We evaluate in \cref{tab:ablation_hpa} the impact of the main components of \textsc{FINO} on the HPA benchmark, and in \cref{tab:siglip_ablation} its dependence on the backbone.
\begin{compactitem}%
    \item \textit{Loss terms.} Removing metadata guidance reduces F1 by $1.7$, while replacing SIGReg with KoLeo reduces it by $0.9$. Removing both produces a larger $2.8$-point decrease, showing that metadata guidance and feature regularisation make complementary contributions.

    \item \textit{Pretraining.} Initialising from DINOv3 rather than training the backbone from scratch improves F1 by $3.3$, showing that general-purpose visual features remain valuable even after substantial adaptation to fluorescence microscopy.

    \item \textit{Prototype guidance.} Replacing the EMA prototype bank with an MLP classifier reduces F1 by $3.3$. The momentum-updated prototypes therefore provide a stronger mechanism for learning from high-cardinality discrete metadata than direct metadata classification.

    \item \textit{Backbone.} Replacing DINOv3 with an image--text-pretrained SigLIP2 ViT-L/16, with all hyperparameters unchanged, preserves the same ordering on all four datasets (\cref{tab:siglip_ablation}): SSL adaptation improves over the frozen backbone, and metadata guidance adds a further $+1.6$ to $+4.5$ points. The benefit of \textsc{FINO} is therefore not tied to the DINOv3 recipe (App.~\ref{app:siglip}).

    \item \textit{Compute.} The metadata heads add a negligible overhead: on a single H100 GPU with batch size 64, \textsc{FINO} runs at $0.76$\,s/iter, against $0.74$\,s/iter for pure SSL adaptation (App.~\ref{app:general}).
\end{compactitem}
Together, these ablations show that \textsc{FINO}'s performance does not come from a single component: metadata guidance, regularisation, pretrained visual features, and prototype-based supervision each contribute to the final representation.

\begin{table}[t]
\centering
\raisebox{-0.5\height}{%
\begin{minipage}{0.42\linewidth}
\caption{{\bf Ablation on HPA Kaggle private set.} The full \textsc{FINO} row is the mean$\pm$standard deviation over 10 complete low-resolution runs; ablation rows are single runs and deltas are relative to that mean. All rows use one fold at $224^2$, rather than the $768^2$ full-resolution HPA setting in \cref{tab:main_extended}.}
\label{tab:ablation_hpa}
\end{minipage}}%
\hfill
\raisebox{-0.5\height}{%
\begin{minipage}{0.56\linewidth}
\centering
\resizebox{\linewidth}{!}{
\input{tables/ablation}
}
\end{minipage}}

\vspace{8pt}
\raisebox{-0.5\height}{%
\begin{minipage}{0.36\linewidth}
\caption{{\bf SigLIP2 backbone.} Same pipeline and hyperparameters as with DINOv3, primary metric per dataset as in \cref{tab:main_extended}.}
\label{tab:siglip_ablation}
\end{minipage}}%
\hfill
\raisebox{-0.5\height}{%
\begin{minipage}{0.6\linewidth}
\centering
\input{tables/siglip}
\end{minipage}}
\end{table}

%% file: tables/main_results2.tex
\begingroup
\setlength{\tabcolsep}{2.8pt}
\footnotesize
\begin{tabular}{@{}c l c c c c c c@{}}
\toprule
\multirow{2}{*}{Type} & \multirow{2}{*}{Method} & \multicolumn{2}{c}{Label supervision} & HPA & iWildCam & FMoW & MIMIC \\
\cmidrule(lr){3-4}\cmidrule(lr){5-5}\cmidrule(lr){6-6}\cmidrule(lr){7-7}\cmidrule(lr){8-8}
& & Param. & Backbone & \textit{F1} & \textit{OOD F1} & \textit{WGA} & \textit{AUROC} \\
\midrule
\multicolumn{2}{@{}l}{Published reference} & \makecell{80--\\300M} & \raisebox{0.5ex}{\orangefire\;}\hspace{-2pt}/\hspace{-2pt}\raisebox{-0.5ex}{\;\bluesnow} & 59.4 & $52.0{\pm}0.4$ & $51.8{\pm}0.4$ & $81.7{\pm}0.1$ \\
\greyrule
\multirow{2}{*}{\shortstack{Supervised\\training}}
& From scratch & 306M & \orangefire & 44.7 & 10.1 & 24.3 & 76.8 \\
& Fine-tuning & 306M & \orangefire & 52.6 & 43.4 & 40.5 & 80.7 \\
\greyrule
\multirow{3}{*}{\shortstack{Domain adapt.\\and general.}}
& DANN & 306M & \orangefire & 50.2 & 38.5 & 46.2 & --- \\
& CORAL & 306M & \orangefire & 41.1 & 38.5 & 45.9 & --- \\
& Group-DRO & 306M & \orangefire & 52.6 & 26.1 & 46.4 & 80.0 \\
\greyrule
\multirow{3}{*}{\shortstack{Representation\\adaptation}}
& Frozen DINOv3 & 2.7M & \bluesnow & 42.2 & 51.4 & 45.0 & 75.9 \\
& DINO (SSL) & 2.7M & \bluesnow & 60.0 & 50.0 & 48.4 & 81.1 \\
& \textsc{FINO} (ours) & 2.7M & \bluesnow
& $\mathbf{61.2}^{*}$
& $\mathbf{52.9{\pm}0.8}$
& $\mathbf{53.3{\pm}0.8}$
& $\mathbf{81.7{\pm}0.1}$ \\
\bottomrule
\end{tabular}
\endgroup

%% file: tables/selection_ledger.tex
\begingroup
\setlength{\tabcolsep}{5pt}
\small
\begin{tabular}{@{}llll@{}}
\toprule
Dataset & Informative $\bM_+$ & Spurious $\bM_-$ & Held-out selection metric \\
\midrule
HPA & antibody & --- & 5-fold cell-line-split macro-F1 \\
iWildCam & --- & time of day & OOD-validation macro-F1 \\
FMoW & subregion & year & OOD-validation WGA \\
MIMIC-CXR & view position & --- & IID-validation AUROC \\
\bottomrule
\end{tabular}
\endgroup

%% file: tables/all_positive.tex
\begingroup
\setlength{\tabcolsep}{5pt}
\small
\begin{tabular}{@{}llcccc@{}}
\toprule
Dataset & Metric & Factors & No guide & Curated & All positive \\
\midrule
FMoW & OOD WGA & 13 & 48.4 & $\mathbf{53.3}$ & 48.4 \\
HPA ($224^2$) & private F1 & 3 & 53.0 & $\mathbf{54.7}$ & $\mathbf{54.7}$ \\
iWildCam & OOD macro-F1 & 4 & 50.0 & $\mathbf{52.9}$ & 52.7 \\
\bottomrule
\end{tabular}
\endgroup

%% file: tables/task_guidance.tex
\begingroup
\setlength{\tabcolsep}{5pt}
\small
\begin{tabular}{@{}l c@{\hspace{0.3em}}l c@{\hspace{0.3em}}l@{}}
\toprule
Guidance during adaptation & \multicolumn{2}{c}{FMoW OOD WGA} & \multicolumn{2}{c}{HPA OOD F1} \\
\midrule
None (SSL adaptation) & $48.4$ & & $53.0$ & \\
\midrule
SupCon (task labels) & $48.3$ & \showchangedark{-0.1} & $54.5$ & \showchangedark{1.5} \\
Task-label prototypes & $48.5$ & \showchangedark{0.1} & $53.1$ & \showchangedark{0.1} \\
SupCon + selected metadata & $51.0$ & \showchangedark{2.6} & $54.0$ & \showchangedark{1.0} \\
\textsc{FINO} (metadata only) & $\mathbf{53.3}$ & \showchangedark{4.9} & $\mathbf{54.7}$ & \showchangedark{1.7} \\
\bottomrule
\end{tabular}
\endgroup

%% file: figures/umap.tex
\begin{tabular}{@{}l@{\;}c@{\;}c@{\;}c@{\;}c@{\;}l}
&\small Frozen DINOv3 & \small Supervised baseline & \small SSL Adaptation & \small \textsc{FINO} & \\
\rotatebox{90}{\footnotesize \qquad\; spurious factor}
&
\includegraphics[width=0.21\linewidth]{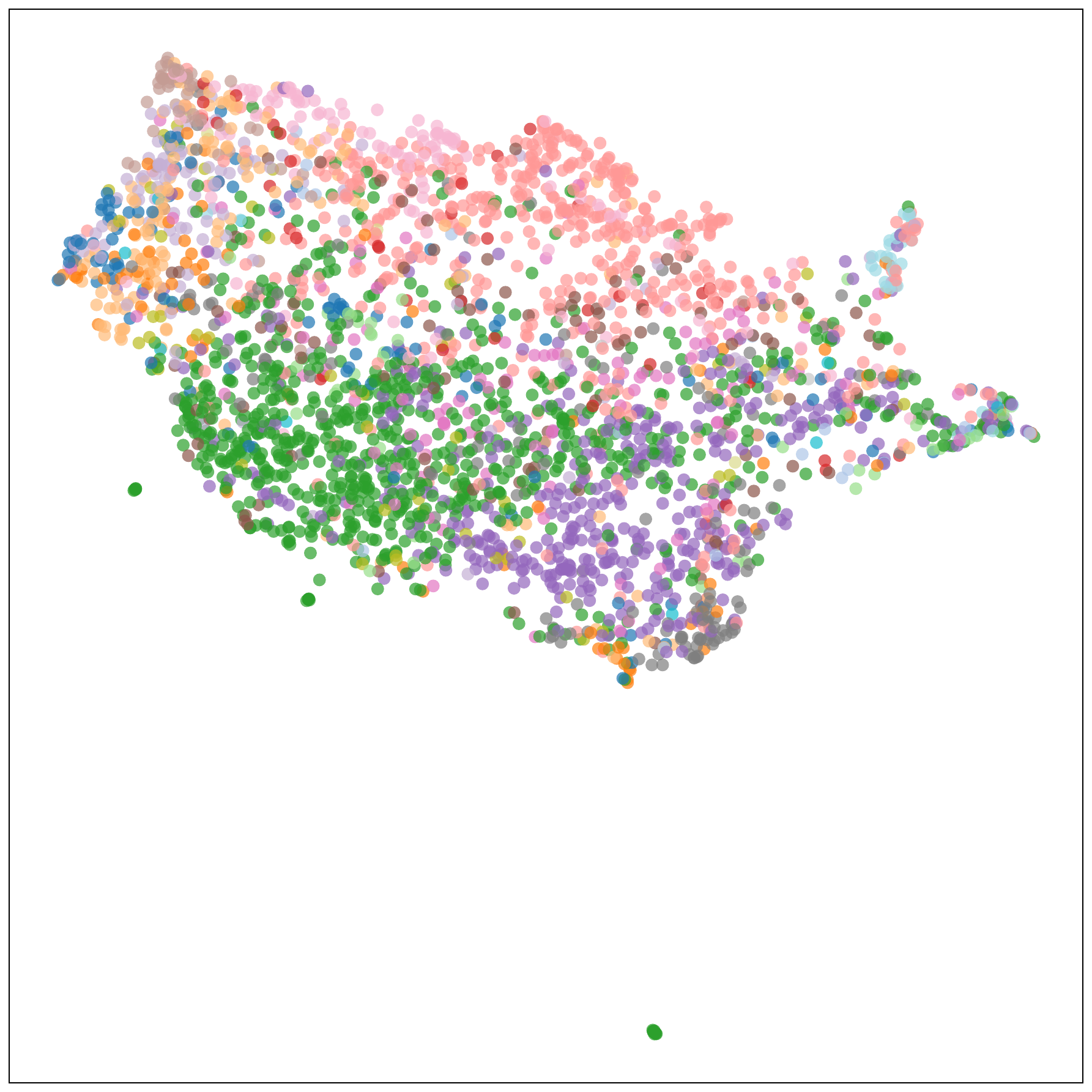}&
\includegraphics[width=0.21\linewidth]{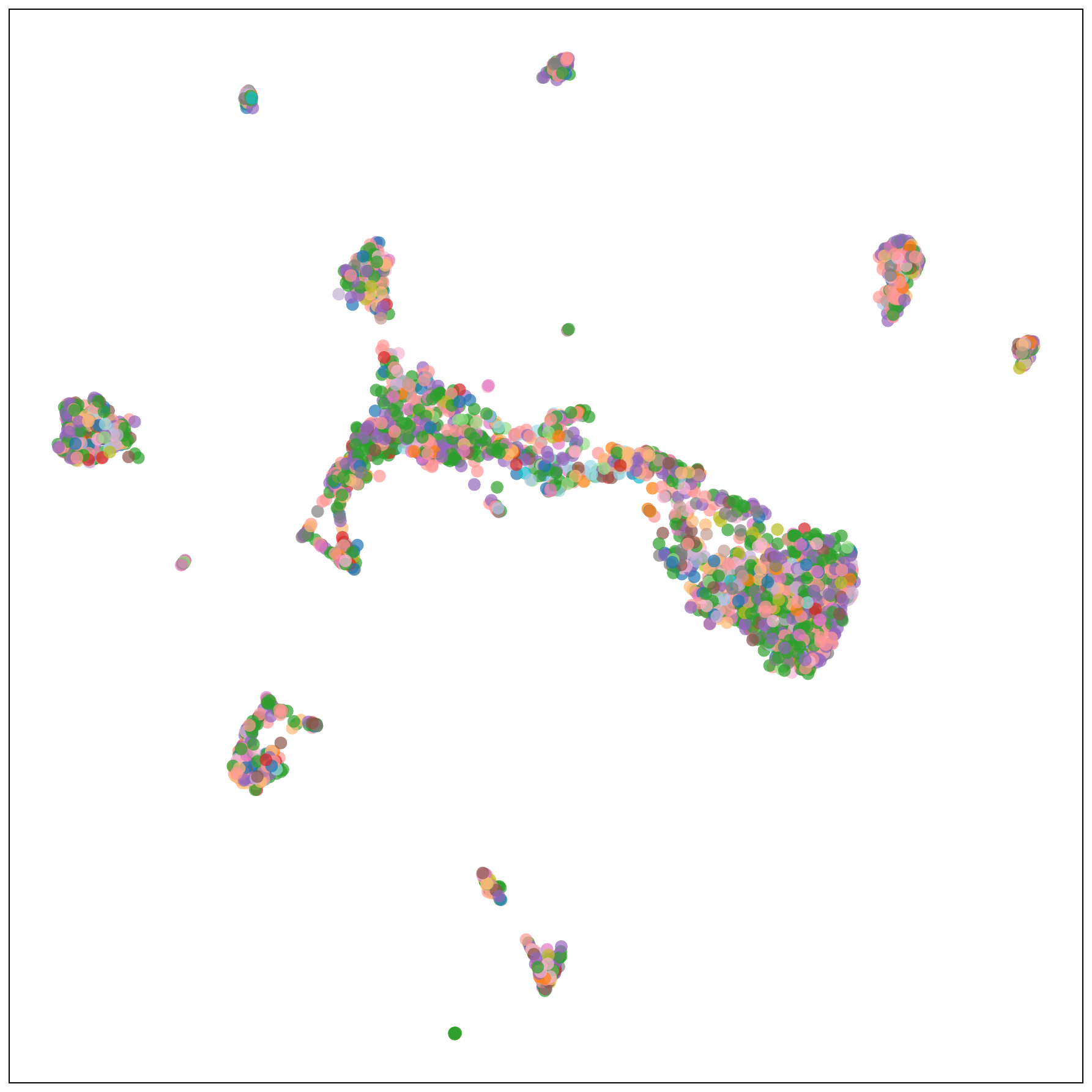}&
\includegraphics[width=0.21\linewidth]{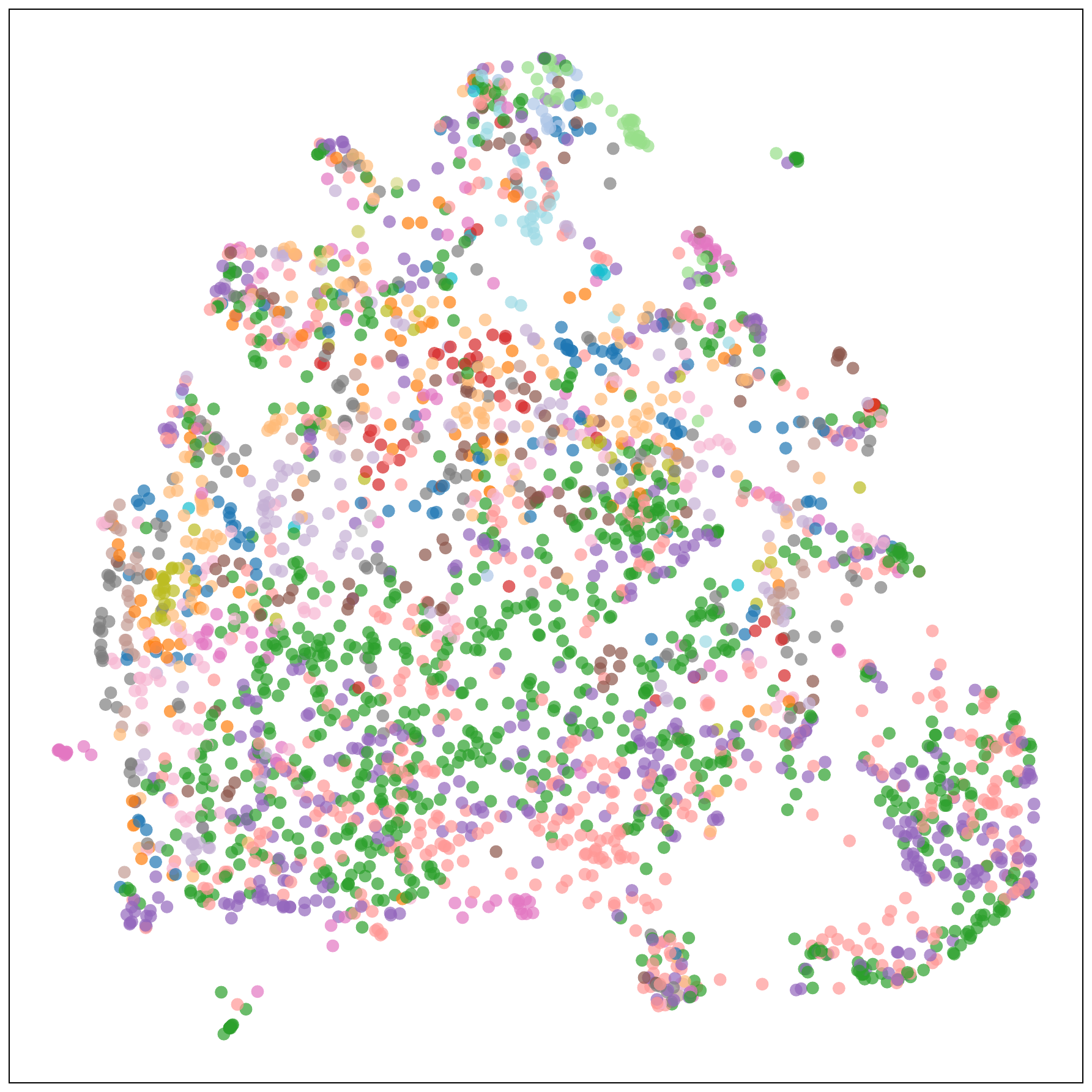}&
\includegraphics[width=0.21\linewidth]{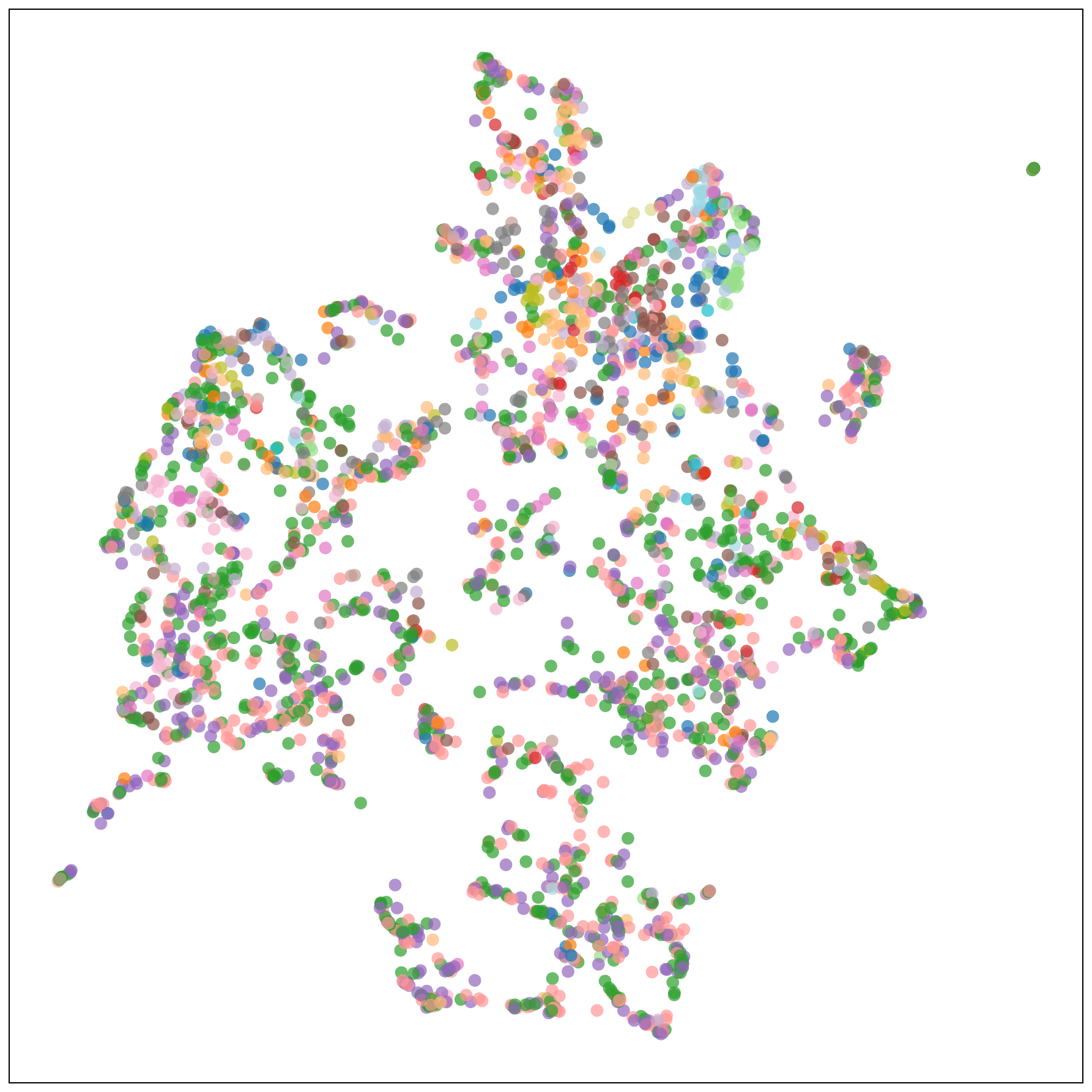}&
\multirow{2}{*}[2.4cm]{
\hspace{-2.2mm}
\begin{minipage}[t][0.03\linewidth][t]{0.085\linewidth}
\resizebox{!}{1.05\linewidth}{
\input{figures/legend_spurious}
}
\end{minipage}
}\\
&\small k-NN score: 41.2 & \small k-NN score: 25.8 & \small k-NN score: 86.8 & \small k-NN score: 74.5 & \\
\rotatebox{90}{\footnotesize \qquad\; target variable}
&
\includegraphics[width=0.21\linewidth]{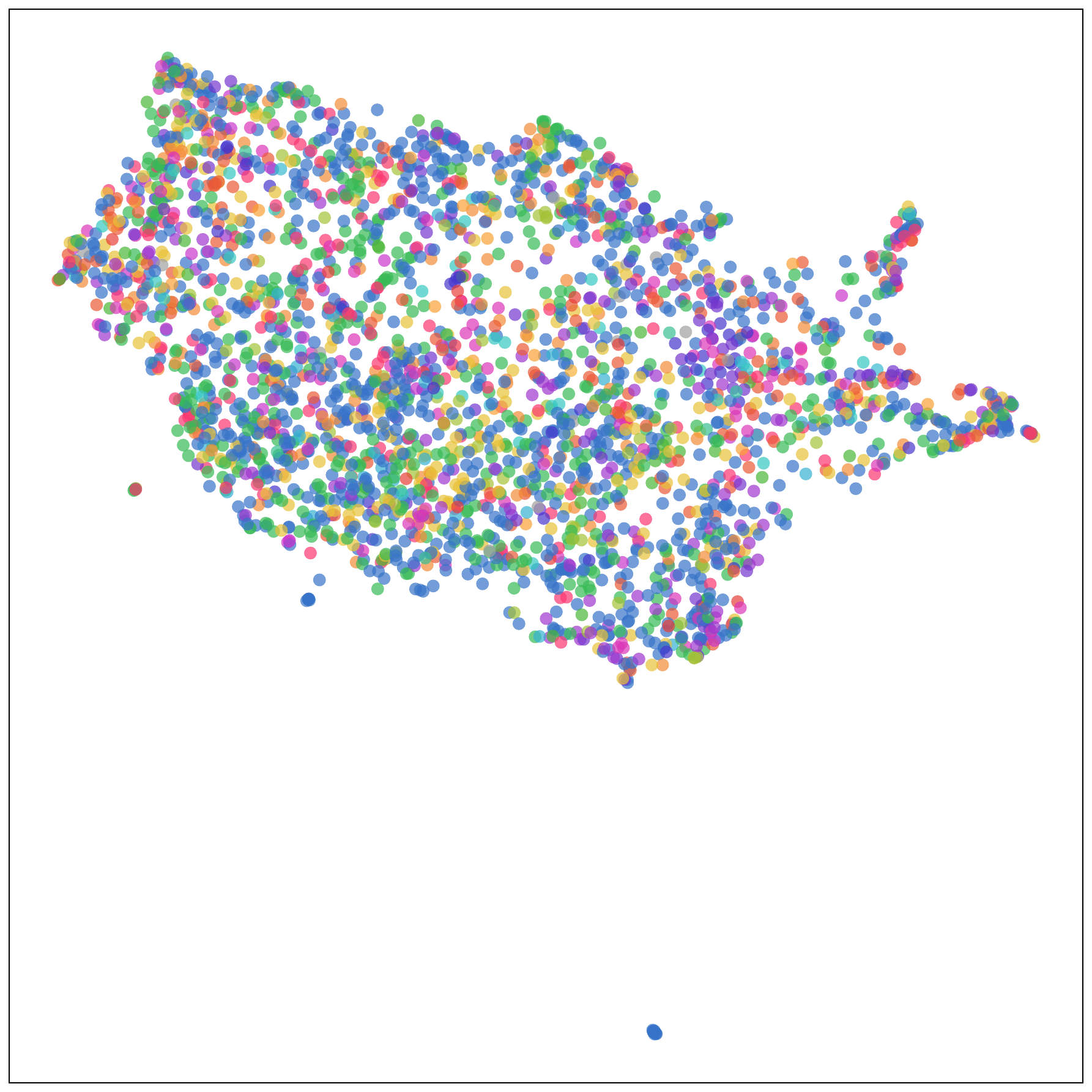}&
\includegraphics[width=0.21\linewidth]{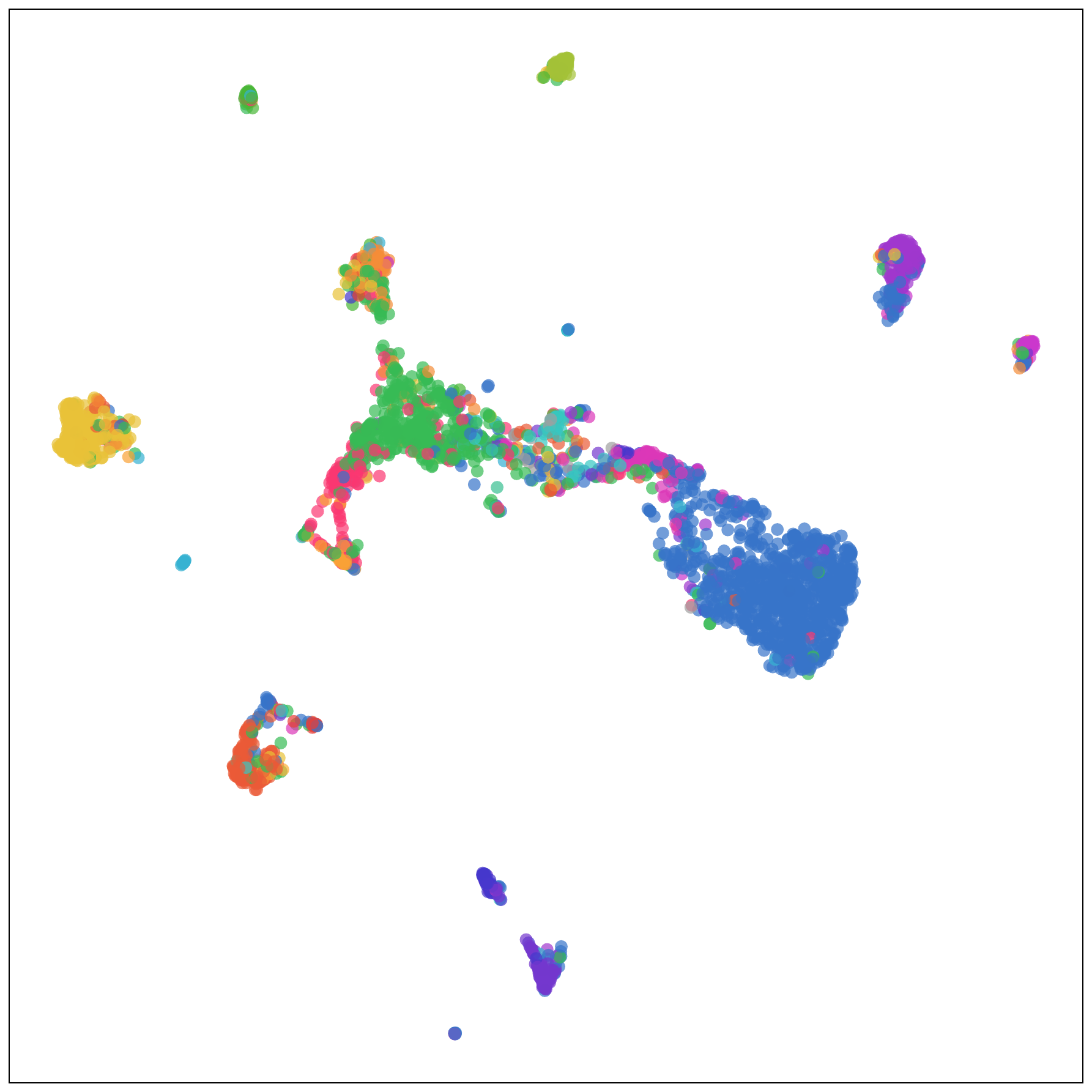}&
\includegraphics[width=0.21\linewidth]{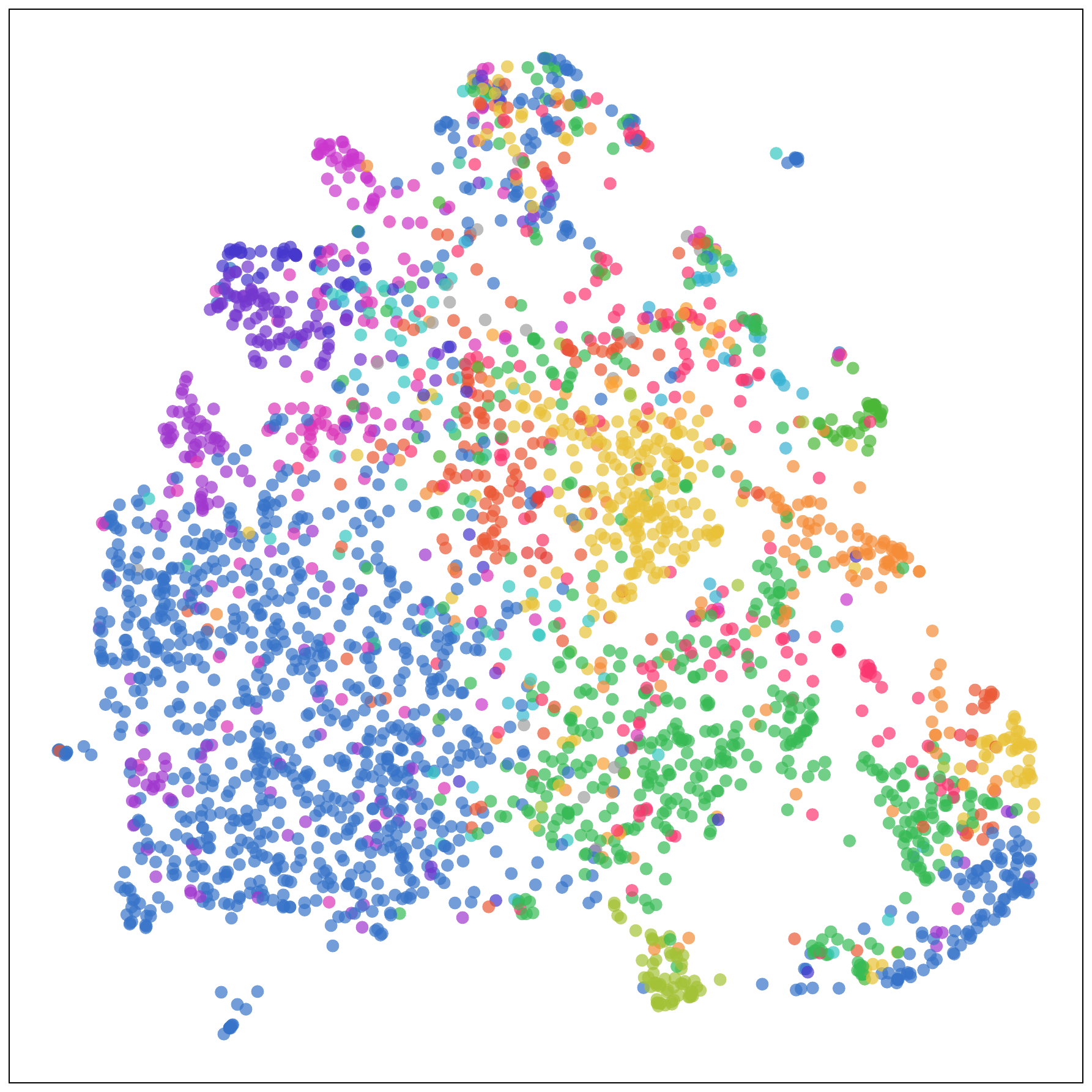}&
\includegraphics[width=0.21\linewidth]{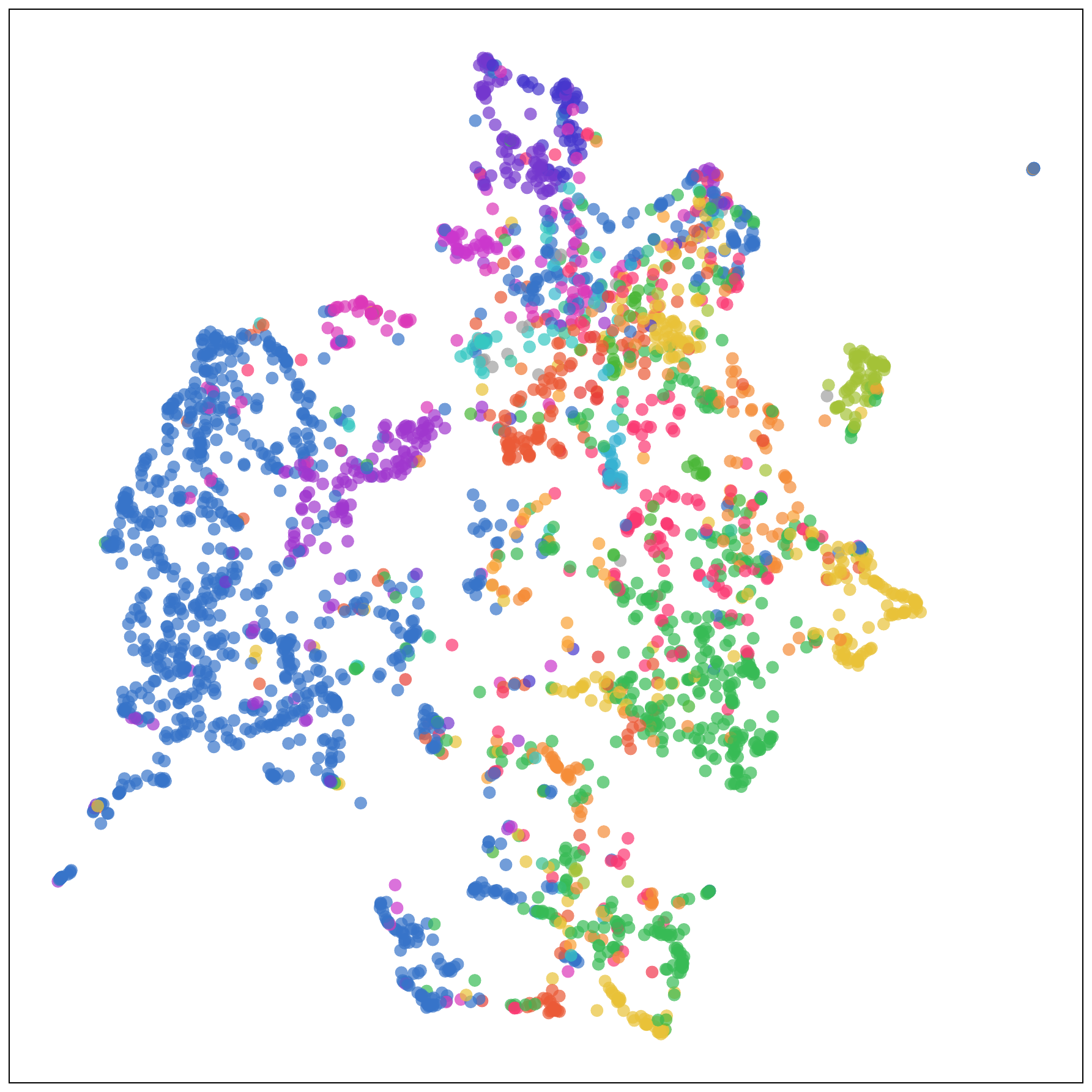}&
\multirow{2}{*}[2.68cm]{
\hspace{-3.2mm}
\begin{minipage}[t][0.04\linewidth][t]{0.1\linewidth}
\resizebox{!}{1.05\linewidth}{
\input{figures/legend_target}
}
\end{minipage}
}
\\
&\small k-NN score: 3.8 & \small k-NN score: 48.0 & \small k-NN score: 73.2 & \small k-NN score: {\bf 83.0} & \\
\end{tabular}\\

%% file: figures/legend_spurious.tex
\tikzset{
    dot/.style={circle, minimum size=3.0mm, inner sep=0pt, draw=none}
}%
\definecolor{c1}{RGB}{31,119,180}%
\definecolor{c2}{RGB}{255,127,14}%
\definecolor{c3}{RGB}{44,160,44}%
\definecolor{c4}{RGB}{214,39,40}%
\definecolor{c5}{RGB}{148,103,189}%
\definecolor{c6}{RGB}{140,86,75}%
\definecolor{c7}{RGB}{227,119,194}%
\definecolor{c8}{RGB}{127,127,127}%
\definecolor{c9}{RGB}{188,189,34}%
\definecolor{c10}{RGB}{23,190,207}%
\definecolor{c11}{RGB}{57,139,207}%
\definecolor{c12}{RGB}{255,157,54}%
\definecolor{c13}{RGB}{74,180,74}%
\definecolor{c14}{RGB}{235,79,80}%
\definecolor{c15}{RGB}{168,123,209}%
\definecolor{c16}{RGB}{170,116,105}%
\definecolor{c17}{RGB}{247,149,214}%
\definecolor{c18}{RGB}{157,157,157}%
\definecolor{c19}{RGB}{208,209,64}%
\definecolor{c20}{RGB}{53,210,227}%
\definecolor{c21}{RGB}{21,99,150}%
\definecolor{c22}{RGB}{225,97,0}%
\definecolor{c23}{RGB}{24,130,24}%
\definecolor{c24}{RGB}{180,20,20}%
\definecolor{c25}{RGB}{118,73,159}%
\definecolor{c26}{RGB}{110,66,55}%
\definecolor{c27}{RGB}{197,89,164}%
\definecolor{c28}{RGB}{97,97,97}%
\definecolor{c29}{RGB}{158,159,4}%
\definecolor{c30}{RGB}{3,160,177}%
\definecolor{c31}{RGB}{0,150,136}%
\definecolor{c32}{RGB}{255,87,34}%
\definecolor{c33}{RGB}{63,81,181}%
\definecolor{c34}{RGB}{205,220,57}%
\definecolor{c35}{RGB}{121,85,72}%
\begin{tabular}{@{}l l@{\hspace{1.2em}} r l@{}}
\tikz[baseline=-0.5ex]{\node[dot, fill=c1] {};}  & BJ
&
\tikz[baseline=-0.5ex]{\node[dot, fill=c19] {};} & AF22
\\
\tikz[baseline=-0.5ex]{\node[dot, fill=c20] {};} & HEL &
\tikz[baseline=-0.5ex]{\node[dot, fill=c3] {};}  & RH-30
\\
\tikz[baseline=-0.5ex]{\node[dot, fill=c21] {};} & Hep G2&
\tikz[baseline=-0.5ex]{\node[dot, fill=c4] {};}  & SH-SY5Y
\\
\tikz[baseline=-0.5ex]{\node[dot, fill=c5] {};}  & U-2 OS
&
\tikz[baseline=-0.5ex]{\node[dot, fill=c23] {};} & THP-1
\\
\tikz[baseline=-0.5ex]{\node[dot, fill=c24] {};} & CACO-2 &
\tikz[baseline=-0.5ex]{\node[dot, fill=c7] {};}  & HaCaT
\\
\tikz[baseline=-0.5ex]{\node[dot, fill=c25] {};} & JURKAT &
\tikz[baseline=-0.5ex]{\node[dot, fill=c8] {};}  & A-431
\\
\tikz[baseline=-0.5ex]{\node[dot, fill=c27] {};} & SuSa &
\tikz[baseline=-0.5ex]{\node[dot, fill=c10] {};} & HEK 293
\\
\tikz[baseline=-0.5ex]{\node[dot, fill=c28] {};} & REH &
\tikz[baseline=-0.5ex]{\node[dot, fill=c11] {};} & A549
\\
\tikz[baseline=-0.5ex]{\node[dot, fill=c29] {};} & HDLM-2 &
\tikz[baseline=-0.5ex]{\node[dot, fill=c12] {};} & RT4
\\
\tikz[baseline=-0.5ex]{\node[dot, fill=c30] {};} & K-562 &
\tikz[baseline=-0.5ex]{\node[dot, fill=c13] {};} & HeLa
\\
\tikz[baseline=-0.5ex]{\node[dot, fill=c31] {};} & hTCEpi &
\tikz[baseline=-0.5ex]{\node[dot, fill=c14] {};} & MCF7
\\
\tikz[baseline=-0.5ex]{\node[dot, fill=c32] {};} & NB-4 &
\tikz[baseline=-0.5ex]{\node[dot, fill=c15] {};} & PC-3
\\
\tikz[baseline=-0.5ex]{\node[dot, fill=c33] {};} & HAP1  &
\tikz[baseline=-0.5ex]{\node[dot, fill=c34] {};} & OE19
\\
\tikz[baseline=-0.5ex]{\node[dot, fill=c35] {};} & SiHa &
\tikz[baseline=-0.5ex]{\node[dot, fill=c18] {};} & EFO-21
\\
\tikz[baseline=-0.5ex]{\node[dot, fill=c2] {};}
& \multicolumn{3}{l}{LHCN-M2}
\\
\tikz[baseline=-0.5ex]{\node[dot, fill=c26] {};} & \multicolumn{3}{l}{RPTEC TERT1}
\\
\tikz[baseline=-0.5ex]{\node[dot, fill=c9] {};}  & \multicolumn{3}{l}{U-251 MG}
\\
\tikz[baseline=-0.5ex]{\node[dot, fill=c6] {};}  & \multicolumn{3}{l}{ASC TERT1}
\\
\tikz[baseline=-0.5ex]{\node[dot, fill=c22] {};} &
\multicolumn{3}{l}{HUVEC TERT2}
\\
\tikz[baseline=-0.5ex]{\node[dot, fill=c16] {};} & \multicolumn{3}{l}{hTERT-RPE1}
\\
\tikz[baseline=-0.5ex]{\node[dot, fill=c17] {};} & \multicolumn{3}{l}{SK-MEL-30}
\end{tabular}

%% file: figures/legend_target.tex
\tikzset{
    dot/.style={circle, minimum size=3.0mm, inner sep=0pt, draw=none}
}%
\definecolor{c1}{RGB}{53,171,209}%
\definecolor{c2}{RGB}{51,174,205}%
\definecolor{c3}{RGB}{54,189,188}%
\definecolor{c4}{RGB}{47,177,76}%
\definecolor{c5}{RGB}{56,188,158}%
\definecolor{c6}{RGB}{55,184,143}%
\definecolor{c7}{RGB}{52,181,72}%
\definecolor{c8}{RGB}{74,177,57}%
\definecolor{c9}{RGB}{163,189,43}%
\definecolor{c10}{RGB}{160,160,160}%
\definecolor{c11}{RGB}{224,182,45}%
\definecolor{c12}{RGB}{238,171,52}%
\definecolor{c13}{RGB}{241,156,50}%
\definecolor{c14}{RGB}{238,137,49}%
\definecolor{c15}{RGB}{236,122,50}%
\definecolor{c16}{RGB}{236,95,55}%
\definecolor{c17}{RGB}{239,72,61}%
\definecolor{c18}{RGB}{241,55,79}%
\definecolor{c19}{RGB}{243,54,96}%
\definecolor{c20}{RGB}{243,49,110}%
\definecolor{c21}{RGB}{209,53,185}%
\definecolor{c22}{RGB}{187,57,204}%
\definecolor{c23}{RGB}{148,62,211}%
\definecolor{c24}{RGB}{109,65,212}%
\definecolor{c25}{RGB}{70,67,210}%
\definecolor{c26}{RGB}{58,115,197}%
\begin{tabular}{rl}
\tikz[baseline=-0.5ex]{\node[dot, fill=c1] {};}  & actin filaments \\
\tikz[baseline=-0.5ex]{\node[dot, fill=c2] {};}  & aggresome \\
\tikz[baseline=-0.5ex]{\node[dot, fill=c3] {};}  & centrosome \\
\tikz[baseline=-0.5ex]{\node[dot, fill=c4] {};}  & focal adhesion sites \\
\tikz[baseline=-0.5ex]{\node[dot, fill=c5] {};}  & cytokinetic bridge \\
\tikz[baseline=-0.5ex]{\node[dot, fill=c6] {};}  & cytoplasmic bodies \\
\tikz[baseline=-0.5ex]{\node[dot, fill=c7] {};}  & cytosol \\
\tikz[baseline=-0.5ex]{\node[dot, fill=c8] {};}  & intermediate filaments \\
\tikz[baseline=-0.5ex]{\node[dot, fill=c9] {};}  & microtubules \\
\tikz[baseline=-0.5ex]{\node[dot, fill=c10] {};} & microtubule org. cent. \\
\tikz[baseline=-0.5ex]{\node[dot, fill=c11] {};} & mitochondria \\
\tikz[baseline=-0.5ex]{\node[dot, fill=c12] {};} & rods \& rings \\
\tikz[baseline=-0.5ex]{\node[dot, fill=c13] {};} & cell junctions \\
\tikz[baseline=-0.5ex]{\node[dot, fill=c14] {};} & endoplasmic reticulum \\
\tikz[baseline=-0.5ex]{\node[dot, fill=c15] {};} & endosomes \\
\tikz[baseline=-0.5ex]{\node[dot, fill=c16] {};} & golgi apparatus \\
\tikz[baseline=-0.5ex]{\node[dot, fill=c17] {};} & lipid droplets \\
\tikz[baseline=-0.5ex]{\node[dot, fill=c18] {};} & lysosomes \\
\tikz[baseline=-0.5ex]{\node[dot, fill=c19] {};} & peroxisomes \\
\tikz[baseline=-0.5ex]{\node[dot, fill=c20] {};} & plasma membrane \\
\tikz[baseline=-0.5ex]{\node[dot, fill=c21] {};} & nuclear bodies \\
\tikz[baseline=-0.5ex]{\node[dot, fill=c22] {};} & nuclear membrane \\
\tikz[baseline=-0.5ex]{\node[dot, fill=c23] {};} & nuclear speckles \\
\tikz[baseline=-0.5ex]{\node[dot, fill=c24] {};} & nucleoli \\
\tikz[baseline=-0.5ex]{\node[dot, fill=c25] {};} & nucleoli fibrillar cent. \\
\tikz[baseline=-0.5ex]{\node[dot, fill=c26] {};} & nucleoplasm \\
\end{tabular}

%% file: tables/transfert.tex
\begingroup
\setlength{\tabcolsep}{6pt}
\small
\begin{tabular}{@{}lccc@{}}
\toprule
& \makecell{HPA $\rightarrow$\\OpenCell~\citep{cho2022opencell}} & \makecell{FMoW $\rightarrow$\\FLAIR-Hub~\citep{garioud2026flair}} & \makecell{MIMIC-CXR $\rightarrow$\\CheXpert~\citep{irvin2019chexpert}} \\
Method & $100{\times}$ARI & mIoU & AUROC \\
\midrule
Frozen DINOv3 & 46.6 & 62.0 & 87.0 \\
Supervised adaptation & 53.0 & 61.9 & 87.3 \\
\textsc{FINO} & $\mathbf{58.8}$ & $\mathbf{62.3}$ & $\mathbf{88.0}$ \\
\bottomrule
\end{tabular}
\endgroup

%% file: figures/lowsup.tex
\definecolor{SUPCOLOR}{RGB}{27,158,119}    %
\definecolor{UDACOLOR}{RGB}{217,95,2}      %
\definecolor{METACOLOR}{RGB}{231,41,138}   %
\definecolor{DROCOLOR}{RGB}{117,112,179}   %

\begin{tikzpicture}
\begin{axis}[
    width=1\linewidth,
    height=4.6cm,%
    xmode=log,
    log basis x=10,
    x dir=reverse,
    xmin=0.9, xmax=110,
    ymin=20, ymax=65,
    ytick={20,25,30,35,40,45,50,55,60,65},
    ylabel={F1 score (\%)},
    xtick={1,2,5,10,50,100},
    xticklabels={1\%,2\%,5\%,10\%,50\%,100\%},
    log ticks with fixed point,
    xlabel={Fraction of labels (log scale)},
    axis lines=left,
    axis line style={black!70},
    tick style={black!70},
    grid=major,
    grid style={gray!20},
    legend style={
        at={(0.02,0.02)},
        anchor=south west,
        draw=none,
        fill=none,
        font=\small
    },
    tick label style={font=\small},
    label style={font=\small},
]

\addplot[name path=meta upper, draw=none, forget plot] coordinates {
    (1,51.478)     %
    (2,54.668)     %
    (5,56.657)     %
    (10,58.405)    %
    (50,59.897)    %
    (100,59.270)   %
};
\addplot[name path=meta lower, draw=none, forget plot] coordinates {
    (1,50.308)     %
    (2,52.688)     %
    (5,54.241)     %
    (10,55.583)    %
    (50,59.131)    %
    (100,59.270)   %
};
\addplot[METACOLOR, fill opacity=0.35, draw=none, forget plot] fill between[
    of=meta upper and meta lower
];
\addplot[
    very thick,
    color=METACOLOR,
    mark=diamond*,
    mark size=3pt,
] coordinates {
    (1,50.893)
    (2,53.678)
    (5,55.449)
    (10,56.994)
    (50,59.514)
    (100,59.270)
};
\addlegendentry{\textsc{FINO}}

\addplot[name path=sup upper, draw=none, forget plot] coordinates {
    (1,29.907)     %
    (2,36.837)     %
    (5,45.398)     %
    (10,51.494)    %
    (50,54.799)    %
    (100,53.310)   %
};
\addplot[name path=sup lower, draw=none, forget plot] coordinates {
    (1,28.759)     %
    (2,34.837)     %
    (5,43.100)     %
    (10,50.110)    %
    (50,52.075)    %
    (100,53.310)   %
};
\addplot[SUPCOLOR, fill opacity=0.35, draw=none, forget plot] fill between[
    of=sup upper and sup lower
];
\addplot[
    very thick,
    color=SUPCOLOR,
    mark=*,
    mark size=2.5pt,
] coordinates {
    (1,29.333)
    (2,35.837)
    (5,44.249)
    (10,50.802)
    (50,53.437)
    (100,53.310)
};
\addlegendentry{Supervised fine-tuning}

\addplot[name path=dro upper, draw=none, forget plot] coordinates {
    (1,27.090)     %
    (2,32.533)     %
    (5,42.388)     %
    (10,49.638)    %
    (50,53.107)    %
    (100,52.580)   %
};
\addplot[name path=dro lower, draw=none, forget plot] coordinates {
    (1,24.376)     %
    (2,29.109)     %
    (5,41.588)     %
    (10,47.026)    %
    (50,51.261)    %
    (100,52.580)   %
};
\addplot[DROCOLOR, fill opacity=0.35, draw=none, forget plot] fill between[
    of=dro upper and dro lower
];
\addplot[
    very thick,
    color=DROCOLOR,
    mark=square*,
    mark size=2.5pt,
] coordinates {
    (1,25.733)
    (2,30.821)
    (5,41.988)
    (10,48.332)
    (50,52.184)
    (100,52.580)
};
\addlegendentry{Group-DRO}

\end{axis}
\end{tikzpicture}

%% file: tables/ablation.tex
\begin{tabular}{cccc c@{\hspace{0.6em}}l}
\toprule
\makecell{meta-\\data}&
\makecell{regula-\\risation} &
\makecell{pre-\\trained}&
\makecell{discrete\\guide}&
\multicolumn{2}{c}{F1-score}\\
\midrule
\yes & SIGReg & \yes& proto & $\mathbf{54.7{\pm}0.7}$ & (ours) \\
\rowcolor{gray!10}\rno &  SIGReg & \wyes& N/A & 53.0 & \showchange{-1.7} \\
\wyes &  \textcolor{red!70!black}{KoLeo} & \wyes & proto & 53.8 & \showchange{-0.9} \\
\rowcolor{gray!10} \rno & \textcolor{red!70!black}{KoLeo} & \wyes & N/A & 51.9 & \showchange{-2.8} \\
\wyes & SIGReg&  \rno & proto & 51.4 & \showchange{-3.3} \\
\rowcolor{gray!10} \wyes & SIGReg & \wyes& \textcolor{red!70!black}{MLP} & 51.4 & \showchange{-3.3} \\

\bottomrule
\end{tabular}

%% file: tables/siglip.tex
\begingroup
\setlength{\tabcolsep}{5pt}
\small
\begin{tabular}{@{}lcccc@{}}
\toprule
& HPA & iWildCam & FMoW & MIMIC \\
SigLIP2 backbone & F1 & OOD F1 & WGA & AUROC \\
\midrule
Frozen & 38.0 & 11.3 & 30.0 & 71.5 \\
SSL adaptation ($\cL_\text{DINO}$) & 45.7 & 16.9 & 35.1 & 77.6 \\
\textsc{FINO} & $\mathbf{47.3}$ & $\mathbf{21.4}$ & $\mathbf{36.8}$ & $\mathbf{79.7}$ \\
\bottomrule
\end{tabular}
\endgroup

%% file: sections/supmat.tex
\renewcommand*{\theHsection}{App.\the\value{section}}
\setcounter{section}{0}
\renewcommand*{\thesection}{A.\arabic{section}}
\setcounter{figure}{0}
\renewcommand*{\theHfigure}{App.\thefigure}
\renewcommand\thefigure{A.\arabic{figure}}
\setcounter{table}{0}
\renewcommand*{\theHtable}{App.\thetable}
\renewcommand\thetable{A.\arabic{table}}

{\noindent\large\bfseries Contents}\par
\vspace{0.3em}\hrule height 0.5pt \vspace{0.7em}
{\small
\begingroup
\setlength{\parindent}{0pt}
\renewcommand{\arraystretch}{1.15}
\begin{tabular}{@{}r@{\hspace{0.9em}}p{0.78\linewidth}@{}}
A.1 & Additional Robustness Experiments \dotfill \pageref{app:additional_robustness} \\
A.2 & Backbone Ablation: SigLIP2 \dotfill \pageref{app:siglip} \\
A.3 & Extended Dataset Overview \dotfill \pageref{app:datasets} \\
A.4 & Per-Factor Metadata Analysis \dotfill \pageref{app:metadata_analysis} \\
A.5 & Implementation Details \dotfill \pageref{app:training} \\
A.6 & General Experimental Setup \dotfill \pageref{app:general} \\
A.7 & Probing Protocol Details \dotfill \pageref{app:evaluation} \\
A.8 & Baseline Implementation Details \dotfill \pageref{app:baselines} \\
A.9 & Cross-Dataset Transfer: Protocols and Results \dotfill \pageref{app:transfer} \\
    & \quad A.9.1\enspace HPA $\rightarrow$ OpenCell \dotfill \pageref{app:transfer_opencell} \\
    & \quad A.9.2\enspace FMoW $\rightarrow$ FLAIR-Hub \dotfill \pageref{app:transfer_flair} \\
    & \quad A.9.3\enspace MIMIC-CXR $\rightarrow$ CheXpert \dotfill \pageref{app:transfer_chexpert} \\
\end{tabular}
\endgroup
}
\vspace{0.5em}\hrule height 0.5pt \vspace{1em}

\input{sections/additional_robustness}

\section{Backbone Ablation: SigLIP2}
\label{app:siglip}

The main paper builds on a DINOv3~\citep{simeoni2025dinov3} initialisation, whose self-distilled features are already strong on the type of dense, fine-grained content emphasised by our benchmarks. A natural concern is therefore whether the gains attributed to \textsc{FINO} actually come from metadata-guided adaptation, or whether they are tied to specifics of the DINOv3 recipe. To probe this, we replicate the entire representation adaptation pipeline on top of a SigLIP2~\citep{tschannen2025siglip} ViT-L/16 checkpoint, using the publicly available HuggingFace weights\footnote{\url{https://huggingface.co/collections/google/siglip2}} and keeping all hyperparameters, heads, and probing protocols identical to the DINOv3 runs (cf.\ App.~\ref{app:training} and App.~\ref{app:evaluation}). Tab.~\ref{tab:siglip_ablation} in the main text reports each benchmark's primary metric for the off-the-shelf frozen backbone, the $\cL_\text{DINO}$-only adaptation, and the full \textsc{FINO} objective.

\paragraph{Discussion.}
On the SigLIP2 backbone, both stages of the adaptation pipeline yield monotonic gains across all four datasets. On iWildCam, self-supervised adaptation alone lifts OOD F1 from $11.3$ to $16.9$ ($+5.6$), and adding metadata guidance contributes a further $+4.5$, reaching $21.4$. The same ordering holds on FMoW worst-group accuracy, where $\cL_\text{DINO}$ recovers $+5.1$ points over the frozen baseline ($30.0 \to 35.1$) and \textsc{FINO} adds another $+1.7$ on top ($35.1 \to 36.8$). The magnitudes of both contributions are comparable to those reported on a DINOv3 initialisation in Tab.~\ref{tab:main_extended}, despite SigLIP2 being trained with a fundamentally different (image--text contrastive) objective. This supports the interpretation that metadata-guided adaptation acts on top of, rather than in lieu of, a strong visual prior, and that its benefit is largely backbone-agnostic.

\section{Extended Dataset Overview}
\label{app:datasets}

We provide in Fig.~\ref{fig:datasets} a detailed overview of all datasets used in this paper. We define \emph{micro-domains} as the fine-grained domains within an application domain (e.g., plates in microscopy, countries in Earth observation). In-distribution (ID) and out-of-distribution (OOD) evaluations are defined with respect to these micro-domains, which correspond to the domain shifts typically considered in UDA.

\begin{compactitem}
    \item {\bf Human Protein Atlas (HPA)~\citep{thul2017subcellular, thul2018hpa}.} A fluorescence microscopy dataset for protein localisation, consisting of four-channel cellular images with strong batch effects and acquisition variability. Micro-domains correspond to \emph{plates}, i.e., groups of images acquired under the same experimental conditions (e.g., staining, illumination, or preparation batch). The source (train) data contains 94,270 images from 1,239 distinct plates. These images correspond to the Kaggle train images and images downloaded from the human protein atlas website as recommended~\citep{ouyang2019hpa}. They match the HPA foV images listed by \cite{moutakanni2025cell}, resized to 768$\times$768 resolution. The target data corresponds to the test set from the Human Protein Atlas Kaggle competition~\citep{ouyang2019hpa} and contains 10,634 images from 218 plates. We report the macro F1-score. The metadata include plate identity, antibody type, and cell line, all discrete.

    \item {\bf Functional Map of the World (FMoW)~\citep{christie2018fmow}.} A large-scale Earth observation dataset of satellite images annotated for land use classification. We use the WILDS benchmark split~\citep{koh2021wildsbenchmarkinthewilddistribution}, which introduces substantial geographic and temporal distribution shifts. Micro-domains correspond to geographic regions (countries) and acquisition time (year). The source data comprises 76,863 images across 146 countries, while the target data contains 22,108 images spanning 171 countries. We report macro-accuracy and worst-group accuracy, defined as the accuracy on the worst-performing domain. The metadata include both discrete (region, country, year, month) and continuous (hour, coordinates, ground sampling distance, sun elevation, off-nadir angle) factors.

    \item {\bf iWildCam~\citep{beery2020iwildcam}.} A camera trap dataset for wildlife classification, characterised by long-tailed distributions and strong environmental variations across locations. Micro-domains correspond to camera trap locations. The source data contains 129,809 images from 243 locations, while the target data follows the WILDS split with 42,791 images from 48 distinct locations for evaluation. We report the macro F1-score on both the in-distribution (ID) and out-of-distribution (OOD) test sets defined in the WILDS benchmark. The metadata include trap identity (discrete) and timestamp (continuous). We used 336$\times$336 global crops similarly to \cite{choi2024autoft}.

    \item {\bf MIMIC-CXR~\citep{johnson2019mimic}.} A large-scale dataset of de-identified chest radiographs paired with free-text radiology reports, used here for multi-label pathology classification over $14$ findings. Unlike the other three primary benchmarks, this dataset has no natural micro-domain partition, so we adopt an in-distribution (IID) protocol with $368{,}960$ training images and $5{,}159$ IID test images. We report the macro AUROC averaged across the $14$ pathology labels. The dataset exposes both discrete metadata (radiograph view position, patient sex) and continuous metadata (patient age); \textsc{FINO} relies only on the discrete view position.
\end{compactitem}

\begin{figure}[h]
    \centering
    \input{figures/datasets}

    \caption{{\bf Extended overview of evaluation datasets.} Datasets span Earth observation (FMoW, FLAIR-Hub), wildlife monitoring (iWildCam), fluorescence microscopy (HPA, OpenCell), and medical imaging (MIMIC-CXR, CheXpert). Representative sample images are shown alongside the application domain, micro-domain structure (fine-grained factors of variation such as geographic regions, camera traps, or experimental plates), dataset scale, and available metadata, categorised as discrete (\discrete) or continuous (\continuous).}
    \label{fig:datasets}
\end{figure}
\footnotetext{The X-ray illustrations are not from MIMIC-CXR nor CheXpert, but a personal Xray from one of the authors.}

\FloatBarrier

\section{Per-Factor Metadata Analysis}
\label{app:metadata_analysis}

The compass plot of Fig.~\ref{fig:compass} sweeps every candidate metadata factor through the two single-branch assignments (adding it in isolation to the encouraged set $\bM_+$ or the suppressed set $\bM_-$) and records the resulting change $\Delta$ in the dataset's primary metric. Each factor's pair of $\Delta$s places it in one of four quadrants: \emph{always beneficial} (helps in either branch), \emph{always harmful} (hurts in either), \emph{informative only} (helps in $\bM_+$, hurts in $\bM_-$), and \emph{spurious only} (the converse). Most factors fall cleanly into \emph{always beneficial}, \emph{informative only}, or \emph{spurious only}. Two HPA factors do not: \emph{plate} surfaces as \emph{informative only}, even though as the canonical carrier of microscopy batch effects it ought to live in \emph{spurious only}; and \emph{cell line} lands in \emph{always harmful} despite carrying obvious task-relevant structure. Both anomalies trace back to entanglement between candidate factors. App.~\ref{app:plate} and App.~\ref{app:cell_line} diagnose each in turn, and App.~\ref{app:correlated_factors} distills the lesson into a practical guideline.

\subsection{Plate: ``informative only'' as a side-effect of two independent leakage pathways}
\label{app:plate}

The plate factor lands at $\Delta\!=\!+0.31$ when encouraged and $\Delta\!=\!-0.51$ when suppressed (Fig.~\ref{fig:compass}), placing it in the \emph{informative only} quadrant. This is surprising: a plate is a physical experimental unit sharing illumination, fixation, and imaging settings, and is the canonical unit of \emph{batch effect} in high-throughput microscopy.

\paragraph{Two independent leakage pathways.} The HPA dataset (source plus target) covers $104{,}904$ images, $1{,}457$ plates, $14{,}678$ antibodies, and $35$ cell lines. Each plate uses a single cell line ($99.25\%$ of plates carry exactly one) and hosts a panel of $\sim\!25$ antibodies (median $25$, max $64$); each antibody mostly lives on one plate. Tab.~\ref{tab:plate_correlation} measures these dependencies through the asymmetric uncertainty coefficient $U(Y\!\mid\!X) = 1 - H(Y\!\mid\!X)/H(Y) \in [0,1]$, the fraction of $Y$'s entropy removed by knowing $X$. Two entries dominate: $U(\text{cell line}\!\mid\!\text{plate})\!=\!0.999$ (one cell line per plate) and $U(\text{antibody}\!\mid\!\text{plate})\!=\!0.652$ (each plate selects a $\sim\!25$-antibody panel out of $14{,}678$). The two pathways are independent: controlling for cell line leaves $U(\text{antibody}\!\mid\!\text{plate}, \text{cell line})\!=\!0.652$ unchanged, so plate's predictiveness of antibody is not mediated by cell line. Plate therefore carries two distinct task-relevant axes (the cell substrate and the antibody panel chosen on it) rather than a single nuisance variable.

\begin{table}[h]
\centering
\caption{{\bf Asymmetric dependence between HPA discrete metadata factors.} Each cell reports $U(Y\!\mid\!X) = 1 - H(Y\!\mid\!X)/H(Y) \in [0,1]$, the fraction of the entropy of $Y$ removed by knowing $X$. Cardinalities ($1{,}457$ plates, $14{,}678$ antibodies, $35$ cell lines) explain part of the asymmetries: a low-cardinality variable on the column side is easy to ``determine'' from a high-cardinality one. The two load-bearing entries are $U(\text{cell line}\!\mid\!\text{plate})\!=\!0.999$ (one cell line per plate by construction) and $U(\text{antibody}\!\mid\!\text{plate})\!=\!0.652$ (each plate selects a $\sim\!25$-antibody panel).}
\label{tab:plate_correlation}
\small
\begin{tabular}{@{}l|ccc@{}}
\toprule
$U(Y\!\mid\!X)$, $X\,\downarrow\;Y\!\rightarrow$ & plate     & antibody  & cell line  \\
\midrule
plate                                            & $1.000$   & $0.652$   & $0.999$    \\
antibody                                         & $0.869$   & $1.000$   & $0.567$    \\
cell line                                        & $0.303$   & $0.129$   & $1.000$    \\
\bottomrule
\end{tabular}
\end{table}

\paragraph{Encouragement: plate is a coarsening of both informative axes.} Plate guidance can be read in two complementary ways. Through $U(\text{cell line}\!\mid\!\text{plate})\!=\!0.999$, pulling plate-aligned images together pulls same-cell-line images together: the $1{,}457$ plate prototypes split each of the $35$ cell-line clusters into $\sim\!40$ sub-clusters. Through $U(\text{plate}\!\mid\!\text{antibody})\!=\!0.869$, the same operation also coarsens antibody guidance, bundling the $14{,}678$ antibodies into groups of $\sim\!25$. Plate's encouragement endpoint sits between those of the two axes it spans: cell line at $\Delta\!=\!-1.30$ (the per-cell-line trap of App.~\ref{app:cell_line}), antibody at $\Delta\!=\!+1.80$, plate at $+0.31$, closer to the antibody side because the finer the partition, the further it pushes the encoder away from collapsing onto cell-line identity. The benefit is real but largely subsumed by antibody when both branches are available.

\paragraph{Suppression: two independent leaks of informative signal.} Gradient reversal on plate hits both pathways at once. Through $U(\text{cell line}\!\mid\!\text{plate})\!=\!0.999$, it amounts to gradient reversal on cell line, which erases the cell-morphology substrate against which the four fluorescence channels are read (App.~\ref{app:cell_line}). Through the residual $U(\text{antibody}\!\mid\!\text{plate}, \text{cell line})\!=\!0.652$, it also suppresses the antibody-panel structure within each cell line, removing the fine-grained protein-related signal that the antibody branch would otherwise encourage. With no antibody anchor in $\bM_+$ to point to, the encoder has no direction left: the desired batch-effect-only signal is the smallest of the three components (cell line, antibody panel, residual batch effect), and the $-0.51$ endpoint reflects the cost of erasing the other two.

\subsection{Cell line: a necessary information}
\label{app:cell_line}

Cell line measures $\Delta\!=\!-1.30$ when encouraged and $\Delta\!=\!-0.52$ when suppressed (Fig.~\ref{fig:compass}), placing it in the \emph{always harmful} quadrant. The label is correct \emph{for the binary $\bM_+/\bM_-$ knob}, but masks two opposing pressures with only a narrow benign band between them:
\begin{compactitem}
\item \emph{Cell morphology is part of the input substrate.} Protein localisation in HPA is read \emph{against} the cellular context (nucleus, membrane, cytoskeleton); fully unlearning morphology removes the reference frame the four fluorescence channels are interpreted in, so aggressive suppression collapses both the substrate and the main task.
\item \emph{Cell morphology correlates with the target.} Different cell lines express different protein repertoires and exhibit systematically different localisation patterns, so encouraging the encoder to organise its features along cell-line identity traps protein representations in a per-cell-line manifold and hurts cross-cell-line generalisation.
\end{compactitem}
The sweet spot retains enough cell morphology to read the image at all, while staying loose enough that protein-localisation features are not pinned to cell type. The binary knob lands on neither side: $\bM_+$ overshoots into the per-cell-line trap, $\bM_-$ overshoots into substrate erasure, and both endpoints register as harmful.

\paragraph{The UMAPs and the metadata cardinalities corroborate this reading.} Fig.~\ref{fig:umap} shows that cell-line k-NN accuracy is high under self-supervised adaptation but drops to $74\%$ for \textsc{FINO} (with antibody as the only informative guide and $\bM_-\!=\!\varnothing$), while protein-localisation k-NN climbs from $4\%$ to $83\%$. Two mechanisms explain why an antibody guide alone lands close to the sweet spot:
\begin{compactitem}
\item \emph{Cardinality split.} HPA contains $14{,}678$ antibodies for only $35$ cell lines, so each cell-line cluster contains on average $\sim\!420$ antibody prototypes. The antibody contrastive loss splits each cell-line cluster into roughly that many sub-clusters and pushes them apart, breaking the coarse cell-line clustering captured by frozen DINOv3. The decoupling is structural: it follows from the partition refinement, not from any specific correlation in the data.
\item \emph{Anchor pressure, not collapse.} The antibody guide is the dataset's target-aligned anchor, so it overlaps with the rest of the meaningful structure by design: that is what makes it informative, and what stops it from over-erasing the other axes. The dependence $U(\text{cell line}\!\mid\!\text{antibody})\!=\!0.57$ in Tab.~\ref{tab:plate_correlation} quantifies the effect: pulling same-antibody images together preserves a substantial share of cell-line structure as a by-product, consistent with cell-line k-NN dropping to $74\%$ rather than to chance.
\end{compactitem}

\subsection{Practical guideline}
\label{app:correlated_factors}

Both HPA cases above stem from the same issue (entanglement between candidate factors makes the binary $\bM_+/\bM_-$ assignment misleading), though they manifest differently: plate as a load-bearing aggregate of two informative axes, cell line as a target whose two failure modes the binary knob cannot interpolate between. The diagnostic is itself derived from the metadata table alone, and is computable in $\sim\!2$ minutes from the discrete metadata columns without running a single training job; the compass endpoints are the consistency check on it, not its source. Three takeaways follow:
\begin{compactitem}
\item \emph{Check the dependence structure first.} Before assigning a candidate factor to $\bM_+$ or $\bM_-$, compute $U(Y\!\mid\!X)$ as in Tab.~\ref{tab:plate_correlation} against the other available factors (or use mutual information / coefficient of determination for symmetric or mixed cases). Conditional variants such as $U(Y\!\mid\!X, Z)$ separate direct from mediated dependence, as in the plate-antibody-cell-line analysis above. The full table is a histogram count over the metadata columns; on HPA's $\sim\!10^5$-row table it runs in minutes on a CPU, so the test is available \emph{before} any branch assignment is committed to.
\item \emph{Prefer positive guidance for entangled factors.} When the dependence test of the previous bullet shows that a candidate factor is a coarsening of a target-aligned axis (here, plate determines cell line at $U\!=\!0.999$ and selects a $\sim\!25$-antibody panel at $U\!=\!0.652$), the binary knob has only one safe setting: assigning it to $\bM_+$ recovers most of that signal at no risk of collapse, at the cost of leaving residual batch effects in the representation. This inverts the textbook microscopy reflex of suppressing plate as a pure batch carrier (\eg, Harmony~\citep{korsunsky2019harmony}, Symphony~\citep{kang2021symphony}, \citet{scalbert2024domaininvariantselfsupervisedlearningbatch}), and the inversion is exactly what the dependence table predicts: the standard reflex is calibrated on the assumption that plate is conditionally independent of the target given the image, which the $0.999$/$0.652$ entries reject. Plate's $+0.31$/$-0.51$ compass endpoints are then the empirical confirmation of that prior prediction, not the basis for it.
\item \emph{Never suppress an entangled factor in isolation.} Suppression alone is what destabilises the optimisation: the gradient-reversal pressure has no direction to point in. Pairing it with a positive guide on the entangled axis is the natural fix, in the spirit of DANN~\citep{ganin2015unsuperviseddomainadaptationbackpropagation} coupling a domain-adversarial head to a label-supervised classifier; here the analogue would be to suppress plate jointly with antibody as a $\bM_+$ anchor. When the entangled factor is purely nuisance, our recipe already supplies the scaffolding: place the informative variable in $\bM_+$ and the nuisance in $\bM_-$ in the \emph{same} run, and let the per-branch gradient equalisation of App.~\ref{app:grad_equal} balance the two pressures (as we do on FMoW with sub-region encouraged and year suppressed).
\end{compactitem}

\section{Implementation Details}
\label{app:training}

\begin{figure}[h]
\centering
\fbox{
\begin{minipage}{0.95\linewidth}
\small
\input{figures/algo}
\end{minipage}
}
\caption{{\bf One training step of \textsc{FINO}.} Informative metadata is encouraged, while spurious metadata is suppressed via gradient reversal.}
\label{fig:algo}
\end{figure}

\paragraph{Input resolution.}
Wherever a benchmark has an established input resolution from prior work, we adopt it directly: we match \citet{choi2024autoft} on the WILDS benchmarks for both FMoW and iWildCam ($336^2$ on iWildCam), and use $512^2$ on MIMIC-CXR following the standard chest-radiograph protocol. HPA is the one dataset where we deliberately depart from the source resolution: we use $768^2$ in the main runs and $576^2$ in the SigLIP2 ablation of App.~\ref{app:siglip}. Both are downscales adopted on \emph{compute} grounds (a single full-resolution HPA run already dominates the per-dataset GPU-hours of Tab.~\ref{tab:gpu_hours}) and not a recipe constraint. The method itself imposes no resolution requirement and operates identically at either size; the two HPA values simply bracket what comfortably fit our compute envelope.

\paragraph{Training schedule.}
Training proceeds in two phases unless otherwise noted. The first phase is a short frozen-backbone warm-up in which the patch embedding, the DINO and iBOT heads, the metadata guidance modules, and SIGReg on the DINO bottleneck are trained while the rest of the encoder is held fixed; we did not tune its length and the method is robust to that choice. In the second phase the backbone is unfrozen with a linear learning-rate warmup over 1k iterations, after which all parameters are trained jointly. We recommend training for at least 50 epochs and selecting the best-validation checkpoint: this length acts as an upper bound on the search rather than a tuned hyperparameter, since best-validation selection is applied uniformly to every method we compare against and the same envelope is therefore granted to all of them.

iWildCam is the one departure from this two-phase recipe. As a camera-trap dataset of essentially natural images it sits unusually close to the DINOv3 pretraining distribution (closer than any typical adaptation target the method is designed for) and we observed prolonged adaptation to actively erode the pretrained features. We therefore skip the frozen warm-up and cap adaptation at $\sim$12 epochs. This reflects the proximity of iWildCam to DINOv3's pretraining distribution and is not a hyperparameter the method asks the user to tune on more typical adaptation targets.

\paragraph{Heads and remaining hyperparameters.}
Both DINO and iBOT heads use 65{,}536 prototypes, a hidden dimension of 2048, a bottleneck dimension of 256, and 3 MLP layers. Discrete metadata prototype branches use a temperature $\tau = 0.023$ and a centroid momentum $\alpha = 0.99$. We use a constant learning-rate schedule after warmup and \texttt{float16} mixed precision. The method showed robustness to $\tau$; the retained value $0.023 = 0.07/3$ is one of three values explored during method development.

\paragraph{$\ell_2$ normalisation in the discrete prototype branches.}
The inner product $\langle \phi(x), p^{t}_{m} \rangle$ in Eq.~(1) of the main paper is computed as a cosine similarity: both the student embedding $\phi(x)$ and each prototype $p^{t}_{m}$ are $\ell_2$-normalised onto the unit sphere immediately before the dot product. Consistently, the EMA update of Eq.~(2) is applied to the $\ell_2$-normalised teacher embedding $\phi_{\text{teacher}}(x) / \|\phi_{\text{teacher}}(x)\|_2$, and the resulting $p^{t}_{m}$ is re-projected onto the unit sphere before being used in any subsequent similarity computation.

Tab.~\ref{tab:loss_weights} reports the loss weights used to balance the different objectives during training. These values are held fixed across datasets and metadata types: every metadata branch in the main results uses $\lambda_{\text{meta}}^{(t)}{=}0.03$, regardless of the application domain or whether the branch is discrete or continuous. We adopt $\lambda_{\text{SIGReg}}{=}0.05$ as proposed by the original LeJEPA paper~\citep{balestriero2025lejepa} and never tune it. The only loss-weight knob we ever sweep is $\lambda_{\text{KoLeo}}{=}0.1$, used only by the $\cL_{\text{DINO}}$-only baseline runs in place of SIGReg.

\begin{table}[h]
\centering
\caption{{\bf Loss weights used during training.} Each term is scaled by a constant factor in the total objective. The DINO and iBOT loss weights are kept at their defaults ($\lambda_{\text{DINO}}{=}\lambda_{\text{iBOT}}{=}1$) and are not listed. $\cL_{\text{KoLeo}}$ is only used for the $\cL_{\text{DINO}}$-only baseline runs and is set to $0$ whenever $\cL_{\text{SIGReg}}$ is active.}
\label{tab:loss_weights}
\small
\begin{tabular}{@{}lll@{}}
\toprule
Loss & Weight & Value \\
\midrule
$\cL_{\text{KoLeo}}$      & $\lambda_{\text{KoLeo}}$      & $0.1$ \\
$\cL_{\text{SIGReg}}$     & $\lambda_{\text{SIGReg}}$     & $0.05$ \\
$\cL_{\text{meta}}^{(t)}$ & $\lambda_{\text{meta}}^{(t)}$ & $0.03$ \\
\bottomrule
\end{tabular}
\end{table}

\paragraph{Metadata gradient schedule.}
The gradient flowing from each metadata branch back into the backbone is modulated by a scalar $\gamma(s) \in [0,1]$ following the standard DANN ramp~\citep{ganin2015unsuperviseddomainadaptationbackpropagation}:
\begin{align}
    \gamma(s) = \frac{2}{1 + e^{-10\,s/S}} - 1,
\end{align}
where $s$ and $S$ are counted from the moment the backbone is unfrozen (i.e.\ from the Phase~1$\to$Phase~2 transition when a frozen Phase~1 is used). The sign is flipped for spurious factors $t \in \bM_-$ (gradient reversal). The schedule applies symmetrically to both metadata branches: continuous heads $g^{(t)}$ keep training with $+1$ on their own parameters, and discrete prototype banks $\{p_k^{t}\}_{k=1}^{K}$ keep being updated by EMA from teacher embeddings, so both representations of the metadata structure converge independently of $\gamma(s)$. Only the encoder is shielded: it sees no metadata gradient at the moment it unfreezes and ramps in smoothly, avoiding adversarial pressure against heads or prototypes that have not yet stabilised.

\paragraph{Per-branch gradient equalisation.}
\label{app:grad_equal}
The fixed weights $\lambda_{\text{meta}}^{(t)}$ of Tab.~\ref{tab:loss_weights} encode the relative importance the user assigns to each metadata branch. When two or more branches are trained jointly, that intent is corrupted by the very different raw gradient magnitudes of the underlying losses (prototypical contrastive on discrete factors, $\ell_2$ regression on continuous ones), under which a single branch can dominate the encoder's updates. We therefore apply, at every training step, a multiplicative correction $s_t$ that equalises gradient $\ell_2$ norms across branches before the fixed $\lambda_{\text{meta}}^{(t)}$ are honoured. With $|\mathcal{T}| = 1$ the target $\bar n$ collapses to $\tilde n_t$ and $s_t \equiv 1$, so the procedure is a no-op and is only meaningful in the multi-guide regime. It has no learnable parameters and no auxiliary objective.

\emph{Per-step update.} The full procedure is given in Alg.~\ref{alg:grad_equal}. At the first step, each $\tilde n_t$ is initialised to the raw branch gradient norm rather than to zero or an externally supplied value, so that the EMA starts on the correct scale. Probing at the shared CLS embedding $\phi(x)$ rather than at the encoder parameters is what keeps it cheap: a single \texttt{torch.autograd.grad} call returns the branch gradient norm without an extra full backward pass through the FSDP-sharded backbone. This relaxation, of using feature-level rather than parameter-level gradients as a surrogate for the shared encoder, was previously used in IMTL-G~\citep{liu2021imtl} and RotoGrad~\citep{javaloy2022rotograd} for the same reason.

\begin{figure}[h]
\centering
\fbox{
\begin{minipage}{0.95\linewidth}
\small
\input{figures/algo_grad_equal}
\end{minipage}
}
\caption{{\bf Per-step gradient equalisation across metadata branches.} For each active branch $t$, an EMA $\tilde n_t$ of the gradient $\ell_2$ norm is maintained at the CLS embedding $\phi(x)$. The branch loss is rescaled by $s_t = \bar n / \tilde n_t$, with $\bar n$ the geometric mean of the smoothed norms; the fixed weights $\lambda_{\text{meta}}^{(t)}$ of Tab.~\ref{tab:loss_weights} are applied on top, separating user-specified relative importance ($\lambda$) from runtime scale calibration ($s$). The detach prevents second-order gradient flow through the scale.}
\label{alg:grad_equal}
\end{figure}

\emph{Behaviour in practice.} The scales $s_t$ are recomputed every step from the live EMAs and drift slowly toward unity as the EMAs stabilise. On FMoW, for instance, $s_t$ for the \texttt{sub\_region} and \texttt{year} branches start out about $12\%$ apart and converge to $\approx 1$ within a few hundred steps, evidence that the correction is genuinely tracking online imbalances rather than acting as a precomputed constant.

\emph{Relation to multi-task gradient balancing.} The metadata branches play, with respect to the shared encoder, a role analogous to tasks in multi-task learning, and the design space of per-task gradient balancers has been extensively studied in that setting. Methods broadly fall into two families: (i) \emph{magnitude balancing}, which rescales each task loss so that gradient magnitudes are comparable on the shared backbone (GradNorm~\citep{chen2018gradnorm}, IMTL-G~\citep{liu2021imtl}, RotoGrad's magnitude block~\citep{javaloy2022rotograd}, MGDA~\citep{sener2018mgda}, FAMO~\citep{liu2023famo}); and (ii) \emph{direction balancing}, which modifies update directions to reduce inter-task conflict (PCGrad~\citep{yu2020pcgrad}, CAGrad~\citep{liu2021cagrad}, NashMTL~\citep{navon2022nashmtl}, Aligned-MTL~\citep{senushkin2023alignedmtl}, RotoGrad's rotation block~\citep{javaloy2022rotograd}). Our procedure belongs to the first family: it neither projects gradients onto cones nor solves an inner optimisation over directions, and is therefore conceptually orthogonal to the projection-based methods (PCGrad, CAGrad, NashMTL, Aligned-MTL). Within the magnitude-balancing family, three closed-form methods are directly comparable; we contrast them below.
\begin{compactitem}
    \item \emph{Vs.\ GradNorm~\citep{chen2018gradnorm}.} GradNorm treats the loss weights $w_i$ as \emph{learnable} parameters and updates them by SGD on a separate gradient loss $\cL_\text{grad} = \sum_i \big| G^{(i)}_W - \bar G \cdot r_i^\alpha \big|_1$, with $\bar G$ the average gradient norm, $r_i$ a relative inverse training rate, and $\alpha$ a tunable asymmetry hyperparameter. The learned $w_i$ are renormalised to $\sum_i w_i = T_\text{tasks}$ at each step. Our $\lambda_{\text{meta}}^{(t)}$ are fixed; nothing is learned, $\alpha$ is absent, and there is no sum constraint. We apply the closed-form ratio $s_t = \bar n / \tilde n_t$ as a multiplicative correction \emph{on top of} the fixed user weight, so that ``relative importance'' (the table value $\lambda$) and ``scale calibration'' (the live $s$) are separated rather than absorbed into a single learned coefficient.
    \item \emph{Vs.\ IMTL-G~\citep{liu2021imtl}.} Like ours, IMTL-G is closed-form, hyperparameter-free, and computed at the shared feature. The two methods differ in target: IMTL-G solves for the unique $\{\alpha_t\}$ such that the aggregated gradient has \emph{equal projections} onto each unit task gradient (the angle bisector of $\{g_t\}$), whereas we equalise the smoothed \emph{magnitudes} of $\{g_t\}$ to their geometric mean. IMTL-G uses instantaneous per-batch gradients with no temporal smoothing; we maintain an EMA $\tilde n_t$, which absorbs the high variance of branch gradient norms in early training without biasing the long-run scale. IMTL-G also reparameterises the loss weights ($\sum_t \alpha_t = 1$, with $\alpha_1$ derived from $\alpha_{2:T}$), whereas we leave the user weights $\lambda_{\text{meta}}^{(t)}$ untouched and apply $s_t$ on top.
    \item \emph{Vs.\ RotoGrad's magnitude block~\citep{javaloy2022rotograd}.} RotoGrad equalises per-task gradient norms to a common target $C = \sum_k \alpha_k \|G_k\|$ with $\alpha_k \propto \|G_k\|/\|G_k^0\|$, i.e.\ a convergence-weighted \emph{arithmetic} mean against the \emph{initial} gradient norms $\|G_k^0\|$. We instead use the \emph{geometric} mean of an \emph{EMA} of current norms: this avoids both the dependence on a single early-training reference (which can be unrepresentative under our two-phase schedule with a frozen backbone) and the need to rank tasks by relative convergence. RotoGrad additionally introduces learnable per-task rotation matrices $R_k$ to align directions; our procedure has no such learnable component and addresses magnitude only.
\end{compactitem}
We do not claim a categorical superiority over these methods (we have not run head-to-head experiments swapping them in for our scheme), but rather a different, lightweight point in the design space, motivated by the specifics of metadata-guided pretraining: a small number of branches ($|\mathcal{T}| \leq 3$ in our experiments), strong heterogeneity between contrastive (discrete) and regression (continuous) losses, and a desire to keep the user-facing loss weights $\lambda_{\text{meta}}^{(t)}$ as a single, interpretable knob for relative importance.

\paragraph{Expanded SIGReg}
\label{app:sigreg}

We apply SIGReg~\citep{balestriero2025lejepa} to the pre-normalisation bottleneck features $z_n$ produced by the DINO head MLP. The loss matches the empirical characteristic function of projected features to that of an isotropic Gaussian:
\begin{align}
    \cL_{\text{SIGReg}}
    =
    \frac{1}{|A|}\sum_{a \in A}
    \int
    \left|
        \frac{1}{N}\sum_{n=1}^{N} e^{i\omega\, a^\top z_n}
        -
        e^{-\omega^2/2}
    \right|^2
    e^{-\omega^2/2}\, d\omega~,
\end{align}
where $A$ is a set of random unit directions, $N$ is the batch size, and $e^{-\omega^2/2}$ is the characteristic function of $\mathcal{N}(0,1)$. The original SIGReg formulation uses a dedicated projection head; we instead reuse the DINO head bottleneck and apply the loss before $\ell_2$ normalisation and prototype projection. All other hyperparameters (number of slicing directions, frequency sampling, and integration scheme), as well as the loss weight $\lambda_{\text{SIGReg}}{=}0.05$, are taken directly from \citep{balestriero2025lejepa} and never tuned per dataset.

\section{General Experimental Setup}
\label{app:general}

\paragraph{Backbone initialisation.}
All experiments use a ViT-L/16 backbone (306M parameters) initialised from the open-source DINOv3~\citep{simeoni2025dinov3} checkpoint released by the authors,\footnote{\url{https://huggingface.co/facebook/dinov3-vitl16-pretrain-lvd1689m}} obtained by knowledge distillation from a ViT-7B teacher trained on the LVD-1689M dataset, providing a strong general-purpose visual initialisation for all downstream adaptation and fine-tuning. This same checkpoint is used in all reported settings, including supervised fine-tuning, domain adaptation, representation adaptation, and frozen probing (e.g., the \emph{frozen model} and \emph{pretrained} rows of Tab.~\ref{tab:main_extended}); the only differences across rows lie in the training objective and which parameters are updated.

\paragraph{Four-channel adaptation for fluorescence microscopy.}
The HPA dataset consists of four-channel fluorescence microscopy images, while the pretrained DINOv3 checkpoint operates on three-channel RGB inputs. To bridge this gap, we adapt the patch embedding layer. Let $\mW \in \sR^{d \times 3 \times p \times p}$ denote the pretrained weights ($d{=}1024$, $p{=}16$). We compute the channel-wise mean $\vm = \frac{1}{3}\sum_{c=1}^{3} \mW_{:,c,:,:} \in \sR^{d \times 1 \times p \times p}$ and form the four-channel embedding:
\begin{equation}
    \mW' = \tfrac{3}{4} \left[\mW \;\|\; \vm\right] \in \sR^{d \times 4 \times p \times p},
    \label{eq:4chan}
\end{equation}
where $\|$ denotes channel-wise concatenation. The factor $\nicefrac{3}{4}$ preserves the expected activation magnitude when summing over four input channels instead of three. All remaining backbone weights are kept unchanged.

\paragraph{Architecture of predictor $g^{(t)}$.}
For each continuous metadata type $t$, $g^{(t)}$ is a 3-hidden-layer MLP applied to the backbone embedding $\phi(x) \in \mathbb{R}^{d}$ ($d{=}1024$). Each hidden block is $\mathrm{Linear} \!\to\! \mathrm{GELU} \!\to\! \mathrm{Dropout}(0.5)$, with widths $[512, 512, 256]$, followed by a linear projection $\mathbb{R}^{256} \!\to\! \mathbb{R}^{n_t}$ and an output activation $\sigma_t \in \{\mathrm{sigmoid}, \tanh, \mathrm{id}\}$ matched to the target range. Cyclic factors (hour) are encoded as $(\sin, \cos)$ pairs ($n_t{=}2$), geographic coordinates as $(\mathrm{lat}, \mathrm{lon})$ ($n_t{=}2$), and scalar bounded factors (cloud cover, sun elevation, off-nadir angle, ground sampling distance) as $n_t{=}1$. Targets are linearly mapped per-dimension via $(m_{\min}^{(t)}, m_{\max}^{(t)})$ into the codomain of $\sigma_t$ ($[0,1]$ for sigmoid, $[-1,1]$ for tanh), and the regression loss is computed in this normalised space.

\paragraph{Stability of gradient reversal on continuous targets.}
With prediction and target both confined to the codomain of $\sigma_t \in \{\mathrm{sigmoid}, \tanh\}$, the squared error is bounded above and the reversed gradient saturates as $g^{(t)}$ approaches the worst-case constant predictor on that interval. SIGReg on the DINO bottleneck (App.~\ref{app:sigreg}) bounds the projected feature norm, and the DANN ramp $\gamma(s)$ of App.~\ref{app:training} keeps the reversed gradient at zero during the frozen-backbone warm-up. Identity activation is reserved for factors not under reversal and never paired with $t \in \bM_-$.

\paragraph{Compute resources.}
All experiments were conducted on H100 or H200 GPUs in a dedicated cluster. Tab.~\ref{tab:gpu_hours} reports the cumulative GPU-hours spent on each dataset, broken down between backbone adaptation (\emph{pretraining}) and downstream probing (\emph{eval}). These numbers reflect the \emph{total} compute consumed by the project, including ablations, preliminary experiments, and method development, rather than only the runs reported in the paper. For comparison, Tab.~\ref{tab:gpu_hours_final} isolates the cost of the single final \textsc{FINO} run used to produce the Tab.~\ref{tab:main_extended} numbers on each dataset, which together account for less than $15\%$ of the project total. Both tables are derived under the conservative assumption of constant 100\% GPU utilisation for the full duration of every job, which substantially overestimates actual energy use. Transfer datasets (OpenCell, CheXpert, FLAIR-Hub) are not listed as they involve only a handful of probing runs and contribute negligibly compared to the figures reported here.

\begin{table}[h]
\centering
\caption{{\bf Cumulative GPU-hours per dataset and stage.} Totals aggregate all jobs run during method development, including ablations and preliminary experiments, and assume constant 100\% GPU utilisation throughout each job.}
\label{tab:gpu_hours}
\begin{tabular}{lccccc}
\toprule
Stage & HPA & FMoW & iWildCam & MIMIC-CXR & Total \\
\midrule
Pretraining & 15{,}360 & 2{,}560 & 160 & 11{,}520 & 29{,}600 \\
Eval        & 97       & 256   & 4{,}000 & 720       & 5{,}073 \\
\midrule
Total       & 15{,}457 & 2{,}816 & 4{,}160 & 12{,}240 & 34{,}673 \\
\bottomrule
\end{tabular}
\end{table}

\paragraph{Efficiency.}
Self-supervised and metadata-guided training introduce a modest overhead relative to standard supervised training due to the use of a teacher model. In practice, this overhead remains small: supervised training runs at 0.68\,s/iter, pure self-supervision at 0.74\,s/iter (\(\sim\)9\% slower), and \textsc{FINO} at 0.76\,s/iter (\(\sim\)12\% slower than supervised training).

\begin{table}[h]
\centering
\caption{{\bf GPU-hours of the final \textsc{FINO} runs reported in Tab.~\ref{tab:main_extended}.}}
\label{tab:gpu_hours_final}
\begin{tabular}{lccccc}
\toprule
Stage & HPA & FMoW & iWildCam & MIMIC-CXR & Total \\
\midrule
Pretraining & 3{,}840 & 160   & 8.0  & 896 & 4{,}904 \\
Eval        & 15.5    & 0.8   & 2.7  & 24  & 43.0 \\
\midrule
Total       & 3{,}855.5 & 160.8 & 10.7 & 920 & 4{,}947 \\
\bottomrule
\end{tabular}
\end{table}
\if 11
\section{Probing Protocol Details}
\label{app:evaluation}
To evaluate the quality of learned representations, we add lightweight classification heads on top of the backbone. For representation adaptation models, we train only the head while keeping the pre-trained backbone frozen. For a given dataset, all baselines use the same head architecture and hyperparameter grid.
\paragraph{Linear head.}
The CLS tokens from the last four transformer blocks are concatenated with the average-pooled patch tokens from the final block. A single linear layer then maps this vector to class logits. The classifier is trained with AdamW and a cosine annealing schedule; we perform a grid search over learning rate and weight decay. For iWildCam, we report the mean across two runs to account for training variability.
\paragraph{Attentive head.}
Patch tokens from the last four transformer blocks are concatenated along the feature dimension, and the resulting sequence of dimension $4d{=}4096$ is projected to a lower-dimensional embedding space ($d_{\text{emb}}{=}512$) to reduce the number of trainable parameters. A single cross-attention layer with 8 heads and one learnable query token produces a 512-dimensional vector, which is passed through a linear head. The classifier is trained with AdamW and cosine annealing. For HPA, we use binary cross-entropy loss for multi-label classification. We perform a grid search over learning rate, weight decay, attention dropout, batch size, and the use of curriculum learning, where during the first epoch only single-labelled images are shown to the model. In the case of HPA where there is no official validation set, we use 5-fold cross-validation with splits defined by cell line for robust hyperparameter selection.

\section{Baseline Implementation Details}
\label{app:baselines}

All supervised and domain adaptation baselines share the same backbone (ViT-L/16) and classification head, differing only in the adaptation loss and its hyperparameters.

\paragraph{Classification head.}
All baselines use the same attentive pooling head as \textsc{FINO} (cf.\ App.~\ref{app:evaluation}), ensuring that performance differences reflect the adaptation strategy rather than the probe. Patch tokens from the last four transformer blocks are concatenated along the feature dimension ($4d{=}4096$), projected to $d_{\text{emb}}{=}512$, and pooled by a single learnable query through cross-attention (8 heads), followed by a linear map to the target label space. The backbone CLS token is not used by the head; pooling operates purely over spatial patch features.

\paragraph{Augmentation.}
For FMoW and iWildCam, we use DINO-style augmentation: random resized crop, horizontal flip, color jitter, random grayscale, and Gaussian blur, with label smoothing of 0.1. For HPA, a domain-specific pipeline \cite{moutakanni2025cell} self-normalises each fluorescence channel independently, using intensity statistics computed from that channel of the same image.

\paragraph{Optimiser and schedule.}
All baselines use AdamW (weight decay $0.04$, gradient clipping $3.0$) with a cosine learning rate schedule. When fine-tuning from the pretrained checkpoint, the peak learning rate is $5{\times}10^{-5}$ with a 5-epoch warmup and the backbone frozen for the first 5 epochs. All runs train for up to 300 epochs with evaluation every 10 epochs and early stopping (patience 3). Model selection uses the best validation OOD metric: worst-group accuracy for FMoW, macro F1 for iWildCam and HPA.

\paragraph{Domain adaptation methods.}
We evaluate three adaptation strategies, each governed by a single scalar hyperparameter controlling adaptation strength:
\begin{compactitem}
\item \textit{CORAL} aligns source and target feature covariance matrices, with loss weight $\lambda$ held constant during training. The covariance discrepancy is normalised by $\nicefrac{1}{4d^2}$.
\item \textit{DANN} trains a 2-layer MLP domain discriminator (hidden dimension 1024) via gradient reversal, with adversarial weight $\lambda$ following a sigmoid warm-up schedule.
\item \textit{Group DRO} reweights per-group training losses via exponentiated gradient ascent with step size $\eta$. Groups are defined by metadata: year (16 groups) on FMoW, camera location (243) on iWildCam, cell line (35) on HPA, and view position (5 groups) on MIMIC-CXR. The MIMIC-CXR grouping matches the metadata factor used by the best-performing \textsc{FINO} run on this dataset.
\end{compactitem}
CORAL and DANN additionally leverage unlabelled OOD samples as target data; Group DRO operates on source data only.

\paragraph{Loss weight sensitivity.}
Tab.~\ref{tab:baseline_sweep} reports performance across the full loss weight sweep for each method. CORAL is robust to $\lambda$ on FMoW and HPA but degrades sharply on iWildCam at larger values ($\lambda{=}1.0$: 24.6 OOD F1 vs.\ 38.5 at $\lambda{=}0.01$). DANN shows moderate sensitivity, with the optimal weight varying across datasets. Group DRO exhibits the most dataset-dependent behaviour, and no single step size $\eta$ dominates across benchmarks: on FMoW, performance is non-monotonic, with WGA peaking at $46.4$ for $\eta{=}0.1$ and falling off to $44.2$ and $43.7$ at $\eta{=}0.01$ and $\eta{=}1.0$ respectively; on HPA with cell line groups, performance is essentially flat across the sweep ($52.6 \to 52.1 \to 52.5$ Private F1); on MIMIC-CXR with view-position groups, AUROC drifts down moderately from $80.0$ at $\eta{=}0.01$ to $77.7$ at $\eta{=}1.0$; and on iWildCam, large steps trigger near-collapse, with OOD F1 falling from $26.1$ at $\eta{=}0.01$ to $9.3$ at $\eta{=}1.0$. The optimal $\eta$ therefore depends on both the granularity and the semantics of the chosen grouping: coarse, balanced partitions tolerate or even benefit from aggressive reweighting, whereas fine-grained, long-tailed partitions such as iWildCam camera locations (many with very few examples) destabilise the exponentiated-gradient updates and require small step sizes to avoid collapse. This per-dataset tuning requirement contrasts with \textsc{FINO}, which uses a single set of loss weights across all benchmarks (Tab.~\ref{tab:loss_weights}).

\begin{table}[h]
\centering
\caption{{\bf Loss weight sensitivity for domain adaptation baselines.} All methods use a DINOv3-pretrained ViT-L/16 backbone. We sweep the loss weight $\lambda$ (CORAL, DANN) or step size $\eta$ (Group DRO) and report performance: worst-group accuracy (WGA) on the FMoW test set, macro F1 on the iWildCam OOD test set, macro F1 on the HPA Kaggle private test set, and AUROC on MIMIC-CXR. Group DRO uses year groups on FMoW (16 groups), camera locations on iWildCam (243), cell line on HPA (35), and view position on MIMIC-CXR (5); see text for alternative groupings. CORAL and DANN require unlabelled target data, which is unavailable in our IID MIMIC-CXR setting (denoted by ---). Italicised column headers denote each benchmark's primary ranking metric.}
\label{tab:baseline_sweep}
\small
\begin{tabular}{@{}llcccc@{}}
\toprule
& & {FMoW} & {iWildCam} & {HPA} & {MIMIC-CXR} \\
\cmidrule(lr){3-3}\cmidrule(lr){4-4}\cmidrule(lr){5-5}\cmidrule(lr){6-6}
{Method} & {Weight} & \textit{WGA} & \textit{OOD F1} & \textit{Private F1} & \textit{AUROC} \\
\midrule
\rowcolor{cOOD!8}
CORAL & $\lambda=0.01$ & 43.8 & 38.5 & 40.5 & --- \\
\rowcolor{cOOD!8}
      & $\lambda=0.1$  & 45.9 & 33.0 & 41.1 & --- \\
\rowcolor{cOOD!8}
      & $\lambda=1.0$  & 42.2 & 24.6 & 37.8 & --- \\
\greyrule
\rowcolor{cAll!8}
DANN  & $\lambda=0.1$  & 45.2 & 38.5 & 50.2 & --- \\
\rowcolor{cAll!8}
      & $\lambda=0.5$  & 46.2 & 33.1 & 30.5 & --- \\
\rowcolor{cAll!8}
      & $\lambda=1.0$  & 45.8 & 36.3 & 39.2 & --- \\
\greyrule
\rowcolor{cIID!8}
Group DRO & $\eta=0.01$ & 44.2 & 26.1 & 52.6 & 80.0 \\
\rowcolor{cIID!8}
          & $\eta=0.1$  & 46.4 & 22.4 & 52.1 & 78.7 \\
\rowcolor{cIID!8}
          & $\eta=1.0$  & 43.7 &  9.3 & 52.5 & 77.7 \\
\bottomrule
\end{tabular}
\end{table}

\paragraph{Alternative DRO grouping strategies.}
On FMoW, using geographic region (6 groups) instead of year yields lower WGA, with the best region run reaching $44.6\%$ at $\eta{=}1.0$ versus $46.4\%$ at $\eta{=}0.1$ with year groups. On HPA, grouping by plate identity (1{,}457 groups) underperforms cell line (35 groups) on the Kaggle private F1: the best plate run reaches $50.8\%$ at $\eta{=}0.1$ versus $52.6\%$ for cell line at $\eta{=}0.01$, and $\eta{=}1.0$ further degrades plate to $42.3\%$ while cell line remains stable at $52.5\%$. This confirms that DRO performance depends on both group granularity and semantic relevance; overly fine-grained partitions fragment the optimisation and prevent effective robust learning.

\paragraph{iWildCam convergence sanity-check.}
The OOD numbers reported on iWildCam by the from-scratch supervised baseline (test OOD F1 around $10$\%) are noticeably lower than on FMoW or HPA, so it is worth checking that this baseline is not simply under-trained. Fig.~\ref{fig:iwildcam_convergence} indicates that this is unlikely: over 300 epochs the training loss falls steadily from 5.2 to 1.0 and training accuracy reaches 93\%, yet held-out F1 saturates much earlier: OOD splits flatten around 10\% by roughly $100$ epochs, while ID splits keep improving only marginally, reaching 29\% on test and 26\% on validation by the final epoch. This pattern is consistent with the difficulty of the iWildCam distribution shift, as previously documented in the WILDS benchmark~\citep{koh2021wildsbenchmarkinthewilddistribution}, rather than with an optimisation issue on our side.

\begin{figure}[h]
\centering
\input{figures/iwildcam_convergence}
\caption{{\bf iWildCam from-scratch supervised baseline: training behaviour vs.\ held-out F1.} {\it Left:} training loss (\textcolor[HTML]{B6522E}{brown}, left axis) and training accuracy (\textcolor[HTML]{7B44DB}{purple}, right axis) over 300 epochs of from-scratch supervised training of a ViT-L/16 backbone; loss decreases from $\sim$5.2 to $\sim$1.04 and accuracy rises to $\sim$93\%, indicating steady fit to the source distribution. {\it Right:} macro F1 on the held-out validation and test splits, broken down by in-distribution (\textcolor{cIID}{green}, ID) and out-of-distribution (\textcolor{cOOD}{orange}, OOD) micro-domains; thicker lines with filled markers denote the test split and thinner lines with hollow markers the validation split. OOD F1 saturates near $\sim$10\% after roughly $100$ epochs, whereas ID F1 keeps improving only marginally, reaching $\sim$29\% (test) and $\sim$26\% (validation) at epoch $300$. The low OOD numbers thus reflect the difficulty of the distribution shift rather than under-training. Training is reported in {\it accuracy}, while held-out evaluation uses the official WILDS macro {\it F1} metric.}
\label{fig:iwildcam_convergence}
\end{figure}

\section{Cross-Dataset Transfer: Protocols and Results}
\label{app:transfer}

\subsection{HPA $\rightarrow$ OpenCell}
\label{app:transfer_opencell}

\paragraph{Evaluation protocol.}
OpenCell~\citep{cho2022opencell} associates each image with a target protein. Following \citet{kobayashi2022cytoself}, representations are first averaged at the protein level, then clustered against ground-truth subcellular localisation labels. We report two unsupervised clustering metrics, the Adjusted Rand Index (ARI) and the V-Measure, both bounded in $[0,1]$ with higher values indicating better agreement with the ground-truth partition. To complement these clustering metrics, which are the only ones reported in the prior work, we additionally train an attentive probe on protein-level embeddings and report its macro F1 score on a held-out test split. Means and standard deviations are computed over 5 clustering seeds; \textsc{CytoSelf} and \textsc{SubCell} numbers are taken from \citet{kobayashi2022cytoself} and \citet{gupta2024subcell} respectively.

\begin{table}[h]
\centering
\caption{{\bf Full transfer results on HPA $\rightarrow$ OpenCell.} ARI and V-Measure are computed on protein-level embeddings against ground-truth subcellular localisation labels, following the \textsc{CytoSelf} protocol; F1 is the macro F1-score of an attentive probe trained on protein-level embeddings. The off-the-shelf DINOv3 backbone alone already exceeds both prior published methods on the metrics they report ($+0.204$ ARI over \textsc{CytoSelf}, $+0.127$ over \textsc{SubCell}); \textsc{FINO} adds a further $+0.122$ ARI, $+0.088$ V-Measure and $+12.6$ F1 on top of that strong starting point, despite never being trained on OpenCell data. Dashes indicate metrics not reported by the corresponding work.}
\label{tab:opencell_full}
\small
\begin{tabular}{@{}lccc@{}}
\toprule
Model & ARI $\uparrow$ & V-Measure $\uparrow$ & F1 $\uparrow$ \\
\midrule
\textsc{CytoSelf}~\citep{kobayashi2022cytoself} & $0.262$ & $0.500$ & --- \\
\textsc{SubCell}~\citep{gupta2024subcell} & $0.339$ & $0.570$ & --- \\
\greyrule
Frozen DINOv3 & $0.466_{\pm 0.019}$ & $0.652_{\pm 0.006}$ & $65.7$ \\
Self-supervised adaptation & $0.486_{\pm 0.022}$ & $0.668_{\pm 0.006}$ & $78.0$ \\
Supervised adaptation & $0.530_{\pm 0.010}$ & $0.710_{\pm 0.006}$ & $69.2$ \\
\rowcolor{cOOD!10}
\textsc{FINO} & $\mathbf{0.588}_{\pm 0.024}$ & $\mathbf{0.740}_{\pm 0.010}$ & $\mathbf{78.3}$ \\
\bottomrule
\end{tabular}
\end{table}

\paragraph{Discussion.}
A first observation is that the off-the-shelf DINOv3 backbone, with no adaptation at all, already clears both prior published methods on the metrics they report: Frozen DINOv3 reaches ARI $0.466$ vs.\ $0.339$ for \textsc{SubCell} and $0.262$ for \textsc{CytoSelf}, and V-Measure $0.652$ vs.\ $0.570$ and $0.500$ respectively. The genuine \textsc{FINO} contribution on top of that strong starting point is $+0.122$ ARI, $+0.088$ V-Measure and $+12.6$ F1 over Frozen DINOv3.

The ranking among the three adaptation regimes is not consistent across metrics: on the clustering metrics (ARI, V-Measure) supervised adaptation comes second behind \textsc{FINO}, with self-supervised adaptation third; on the attentive-probe F1, the order of the two intermediate baselines flips, with self-supervised adaptation second and supervised adaptation last (and even below the frozen backbone). The gap between clustering and probe performance for the supervised baseline (high ARI but lower F1) suggests that supervised adaptation tightens clusters around the HPA label set but transfers less well to a probe trained on OpenCell labels. Self-supervised adaptation, conversely, yields a more diffuse but better-transferring representation. \textsc{FINO} is the only regime that wins on all three metrics simultaneously, combining the most structured clusters with the most probe-friendly features. Fig.~\ref{fig:opencell_tsne_comparison} visualises this gap qualitatively against the UMAP published by \citet{gupta2024subcell}.

\begin{figure}[h]
\centering
\begin{tabular}{cc}
\begin{minipage}{0.46\linewidth}
\centering
\includegraphics[width=\linewidth]{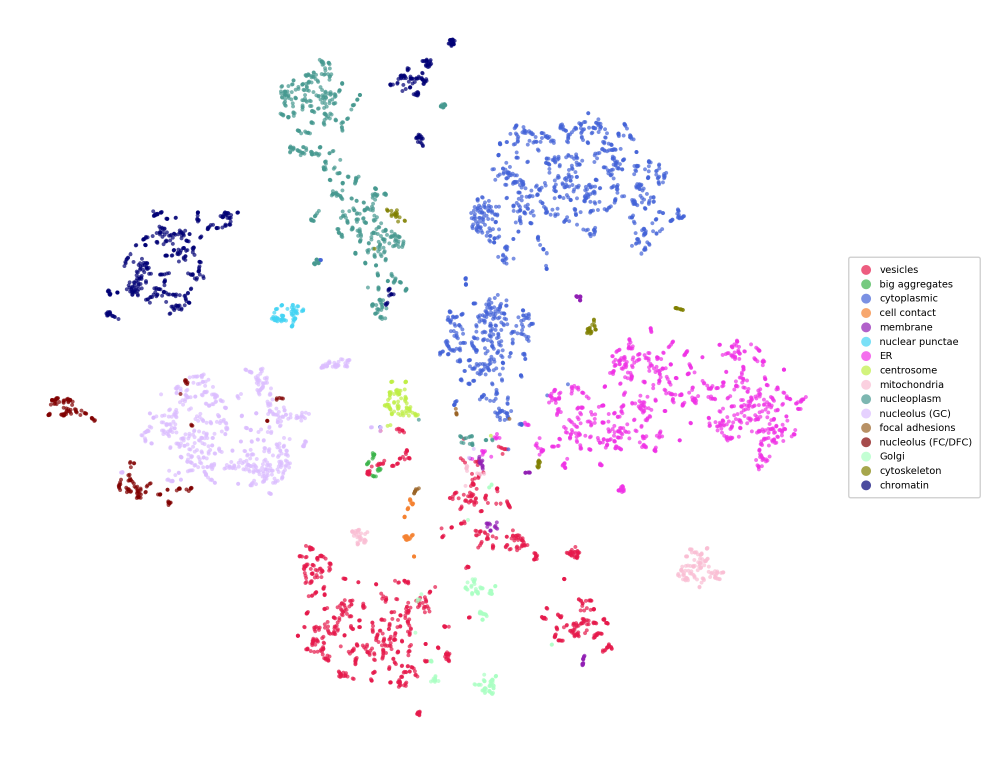}\\
\subcaption{\textsc{FINO} (ours)}
\end{minipage}&
\begin{minipage}{0.46\linewidth}
\centering
\includegraphics[width=\linewidth]{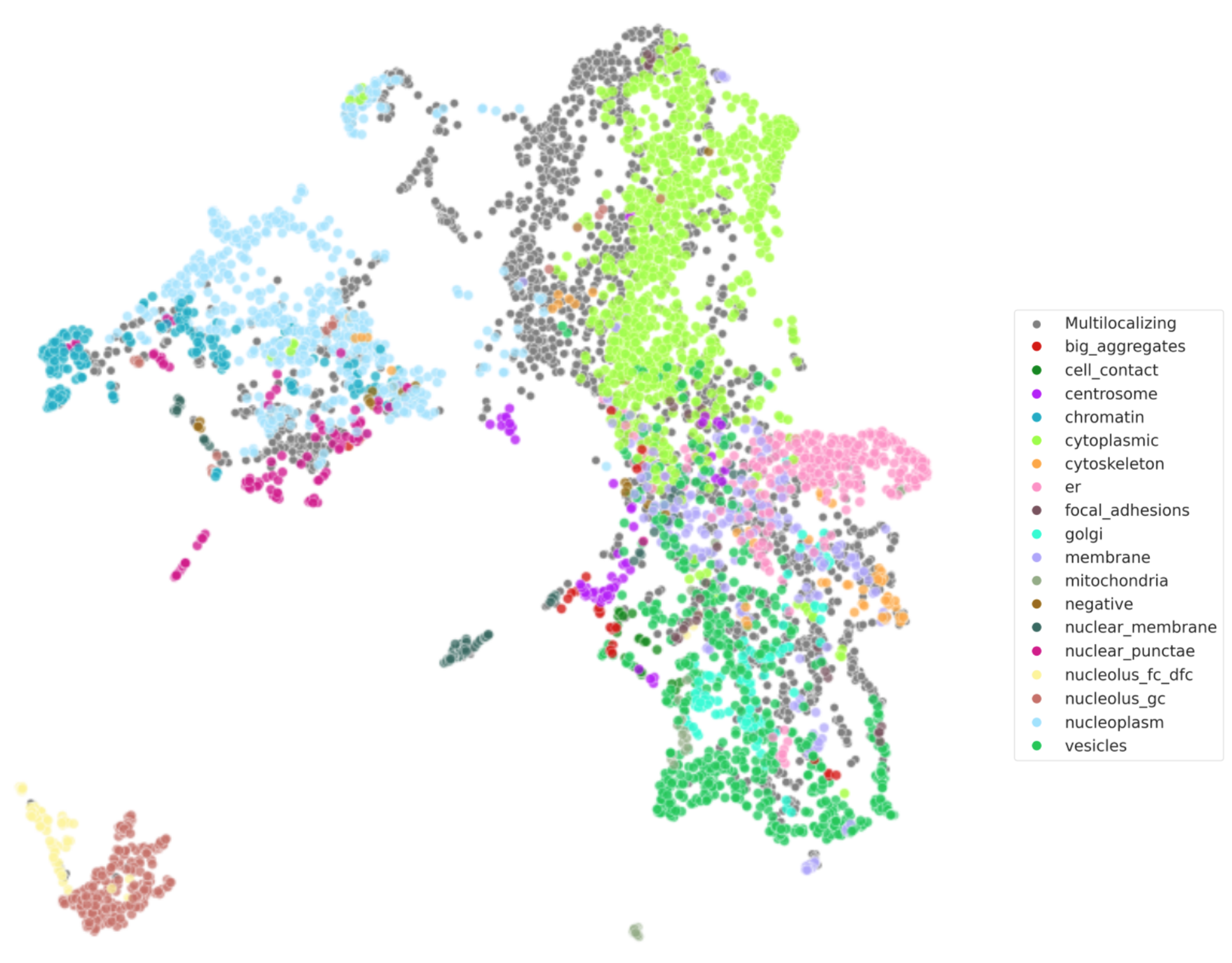}
\subcaption{SubCell~\citep{gupta2024subcell}, reproduced from original paper}
\end{minipage}
\end{tabular}
\caption{{\bf UMAP of OpenCell protein-level embeddings, coloured by subcellular localisation.} (a) Our \textsc{FINO} representation. (b) The corresponding visualisation from \citet{gupta2024subcell}, shown with their own colour scheme; colour assignments differ between the two panels and are therefore not directly comparable label-by-label. The point of comparison is cluster geometry: \textsc{FINO} produces visibly tighter, better-separated localisation clusters, consistent with the gap in clustering metrics reported in Tab.~\ref{tab:opencell_full}.}
\label{fig:opencell_tsne_comparison}
\end{figure}

\subsection{FMoW $\rightarrow$ FLAIR-Hub}
\label{app:transfer_flair}

We probe the FMoW-adapted backbone on FLAIR-Hub~\citep{garioud2026flair}, a large-scale aerial imagery benchmark for land-cover and crop mapping at $0.2$\,m/px (vs.\ $\sim$0.3--1.5\,m for FMoW satellite imagery). Only the RGB channels are kept to match the FMoW input space; the additional spectral bands are discarded.

\paragraph{DPT head.} We use the DPT architecture from \citet{ranftl2021dpt}. The last four transformer layers are extracted, and \texttt{CLS} token information is folded into the spatial features via a learned projection (concatenation followed by a linear layer). Each layer's features are projected to intermediate channel dimensions $[128, 256, 512, 1024]$ with $1{\times}1$ convolutions, then spatially aligned to four different scales using transposed convolutions ($4{\times}$ and $2{\times}$ upsampling), identity, and a strided $3{\times}3$ convolution ($2{\times}$ downsampling). All four feature maps are then projected to $256$ channels with $3{\times}3$ convolutions, and progressively fused from coarse to fine through four fusion blocks. Each fusion block consists of two pre-activation residual units followed by $2{\times}$ bilinear upsampling and a $1{\times}1$ projection. A final $3{\times}3$ convolution produces the fused representation, which is mapped to class logits by a $1{\times}1$ convolution.

\subsection{MIMIC-CXR $\rightarrow$ CheXpert}
\label{app:transfer_chexpert}

We probe the MIMIC-CXR-adapted backbone on CheXpert~\citep{irvin2019chexpert}, a chest radiograph dataset acquired at a different institution (Stanford Hospital) than MIMIC-CXR (Beth Israel Deaconess Medical Center), keeping the same multi-label pathology classification task.
\fi

%% file: sections/additional_robustness.tex
\section{Additional Robustness Experiments}
\label{app:additional_robustness}

\paragraph{Complete-adaptation reproducibility.}
We repeated the entire adaptation and evaluation pipeline rather than retraining probes around a fixed representation. Table~\ref{tab:complete_repeats} reports the resulting mean and standard deviation. Table~\ref{tab:main_extended} reports the full-high-resolution HPA result; for compute reasons, the 10-run HPA repeat uses the $224^2$ setting of Table~\ref{tab:ablation_hpa} and obtains $54.7{\pm}0.7$.

\begin{table}[h]
\centering
\caption{{\bf Reproducibility across complete adaptation runs.} HPA uses the low-resolution setting; its full-high-resolution result is $61.2$ in Table~\ref{tab:main_extended}.}
\label{tab:complete_repeats}
\small
\begin{tabular}{@{}llcc@{}}
\toprule
Dataset & Metric & Runs & Mean$\pm$std \\
\midrule
FMoW & OOD WGA & 5 & $53.3{\pm}0.8$ \\
MIMIC-CXR & AUROC & 5 & $81.7{\pm}0.1$ \\
iWildCam & OOD macro-F1 & 8 & $52.9{\pm}0.8$ \\
HPA ($224^2$) & F1 & 10 & $54.7{\pm}0.7$ \\
\bottomrule
\end{tabular}
\end{table}

\paragraph{Selection ledger.}
Table~\ref{tab:selection_ledger} in the main text makes the final metadata assignments and selection signals explicit. Candidate signs were first evaluated on held-out validation data; the adaptation checkpoint and probe hyperparameters were likewise chosen by the listed held-out metric. HPA public/private scores are evaluation metrics, not selection metrics.

\paragraph{Task-label-supervised guidance.}
We compare metadata-only \textsc{FINO} with three supervised adaptation references while retaining the matched DINO+iBOT machinery and frozen-probe evaluation: conventional SupCon; task-label guidance through our prototype objective; and SupCon combined with the selected metadata. For multi-label HPA, SupCon positives share at least one active label, while prototype guidance uses a uniform target over active classes and updates all active prototypes.

\begin{table}[h]
\centering
\caption{{\bf Supervised task-label guidance on FMoW and low-resolution HPA.} Metadata-only \textsc{FINO} is strongest on both OOD metrics.}
\label{tab:task_guidance_full}
\small
\begin{tabular}{@{}lcccc@{}}
\toprule
& \multicolumn{2}{c}{FMoW WGA} & \multicolumn{2}{c}{HPA F1} \\
\cmidrule(lr){2-3}\cmidrule(lr){4-5}
Adaptation guidance & IID & OOD & IID & OOD \\
\midrule
SupCon (task labels) & 72.3 & 48.3 & 68.6 & 54.5 \\
Task-label prototypes & $\mathbf{72.8}$ & 48.5 & 67.4 & 53.1 \\
SupCon + selected metadata & 72.7 & 51.0 & $\mathbf{69.7}$ & 54.0 \\
\textsc{FINO} (metadata only) & 72.1 & $\mathbf{53.3}$ & 68.7 & $\mathbf{54.7}$ \\
\bottomrule
\end{tabular}
\end{table}

\paragraph{FMoW performance ablations.}
LP-FT~\citep{kumar2022lpft} mitigates the degradation from direct supervised fine-tuning but remains below frozen probing and \textsc{FINO}. Although LP-FT and \textsc{FINO} both freeze and then unfreeze a backbone, LP-FT warms a task-supervised classifier; \textsc{FINO} warms unlabeled SSL/metadata heads and the patch embedding. Removing \textsc{FINO}'s frozen warm-up while keeping the overall adaptation budget fixed reduces WGA to $47.7$, confirming the value of the warm-up at a fixed adaptation budget.

\begin{table}[h]
\centering
\caption{{\bf Additional FMoW performance comparisons.} All rows report OOD worst-group accuracy under the matched evaluation protocol.}
\label{tab:fmow_additional}
\small
\begin{tabular}{@{}lc@{}}
\toprule
Method & OOD WGA \\
\midrule
Frozen backbone + probe & 45.0 \\
Supervised fine-tuning & 40.5 \\
LP-FT & 44.1 \\
SSL adaptation & 48.4 \\
\textsc{FINO} without frozen warm-up & 47.7 \\
\textsc{FINO} & $\mathbf{53.3}$ \\
\bottomrule
\end{tabular}
\end{table}

\paragraph{Selection and release protocol.}
All adaptation configurations, checkpoints, metadata assignments, and probe hyperparameters were selected exclusively with held-out validation data. Neither the HPA Kaggle public nor private score informed selection. We release the complete end-to-end code and metadata pipeline, including data acquisition and preprocessing, dataloaders, exact configurations, environments, splits, seeds, probes, metrics, commands, and table scripts. Laboratory policy prohibits redistribution of trained checkpoints.

%% file: figures/datasets.tex
\small
\renewcommand{\arraystretch}{1.15}
\newlength{\imgsize}
\setlength{\imgsize}{0.155\linewidth}
\newcommand{\squareimg}[1]{%
  \begin{tikzpicture}[baseline={(current bounding box.center)}]
    \useasboundingbox (0,0) rectangle (\imgsize, \imgsize);
    \clip (0,0) rectangle (\imgsize, \imgsize);
    \node[anchor=center, inner sep=0pt, overlay] at ({0.5\imgsize}, {0.5\imgsize}) {%
      \includegraphics[width=\imgsize]{#1}};
  \end{tikzpicture}%
}
\newcommand{\squareimgH}[1]{%
  \begin{tikzpicture}[baseline={(current bounding box.center)}]
    \useasboundingbox (0,0) rectangle (\imgsize, \imgsize);
    \clip (0,0) rectangle (\imgsize, \imgsize);
    \node[anchor=center, inner sep=0pt, overlay] at ({0.5\imgsize}, {0.5\imgsize}) {%
      \includegraphics[height=\imgsize]{#1}};
  \end{tikzpicture}%
}
\definecolor{rowshade}{RGB}{255,243,224}

\centering

\resizebox{\linewidth}{!}{%
\begin{tabular}{@{} >{\raggedright\arraybackslash}m{1.8cm} *{2}{>{\centering\arraybackslash}m{0.18\linewidth}} >{\centering\arraybackslash}m{\dimexpr0.18\linewidth+1cm\relax} >{\centering\arraybackslash}m{0.18\linewidth} @{}}
\toprule
 & HPA & iWildCam & FMoW & MIMIC-CXR \protect\footnotemark  \\
\midrule
 & \squareimg{images/HPAWhole.jpg}
 & \squareimg{images/iwildcam.jpeg}
 & \squareimgH{images/FMOW.jpg}
 & \squareimg{images/chest_xray.png} \\[2pt]
Domain
 & Fluor. microscopy & Wildlife & Earth obs. & Chest X-ray \\
\rowcolor{rowshade}
Micro-domains
 & plates & traps & countries & --- (IID) \\
Source
 & \makecell{94,270 img\\1,239 plates}
 & \makecell{129,809 img\\243 traps}
 & \makecell{76,863 img\\146 countries}
 & \makecell{368,960 img\\5 view pos.} \\
\rowcolor{rowshade}
Target
 & \makecell{23,599 img\\218 plates}
 & \makecell{42,791 img\\48 traps}
 & \makecell{22,108 img\\171 countries}
 & \makecell{5,159 img\\(IID test)} \\
 Metadata used in Tab. 1
 & \makecell{\discrete antibody ($\bM_+$)}
 & \makecell{\continuous time of day ($\bM_-$)}
 & \makecell{\discrete subregion ($\bM_+$)\\\discrete year ($\bM_-$)}
 & \makecell{\discrete view position ($\bM_+$)} \\
\rowcolor{rowshade}
 Other metadata
 & \makecell{\discrete plates\\\discrete cell line}
 & \makecell{\discrete trap\\\discrete month}
 & \makecell{\discrete region, country, month\\\continuous hour, lon, lat\\\continuous sun elev, view angle, res}
 & \makecell{\discrete patient sex\\\continuous patient age} \\
\bottomrule
\end{tabular}%
}

\vspace{3mm}

\begin{tabular}{@{} >{\raggedleft\arraybackslash}m{2.4cm} *{3}{>{\centering\arraybackslash}m{0.18\linewidth}} @{}}
\toprule
 & \textit{OpenCell} & \textit{FLAIR-Hub} & \textit{CheXpert} \protect\footnotemark[\value{footnote}] \\
\midrule
 & \squareimg{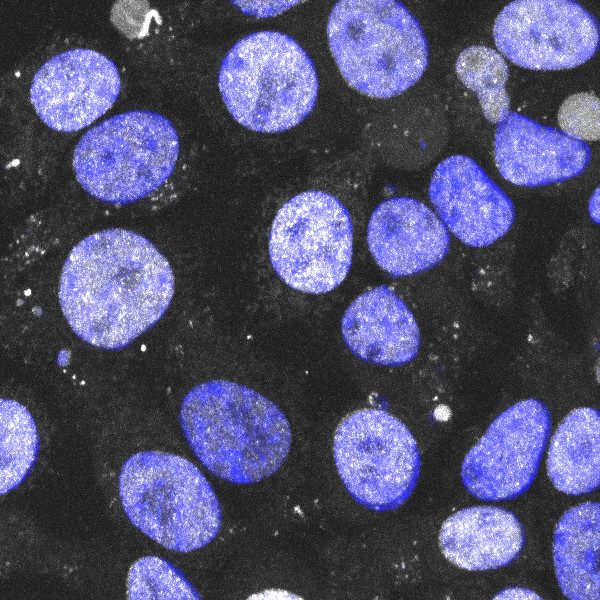}
 & \squareimg{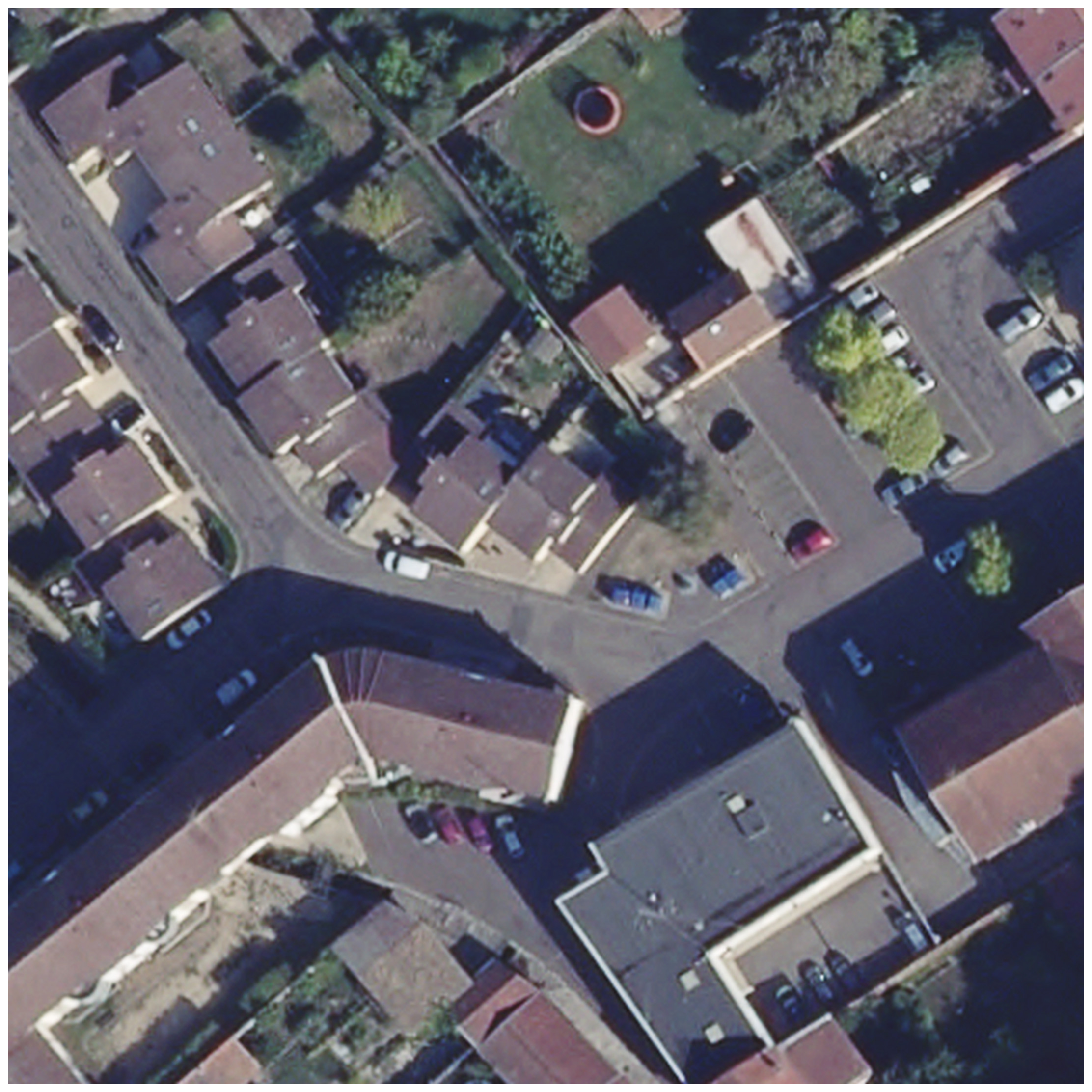}
 & \squareimgH{images/chest_xray.png} \\[2pt]
Transfer from
 & HPA & FMoW & MIMIC-CXR \\
\rowcolor{rowshade}
Channels
 & 2: nucleus, protein \textit{(microtubule, ER dropped)}
 & RGB only \textit{(other bands dropped)}
 & single channel \textit{(unchanged)} \\
Resolution
 & $100^2$ crops \textit{(vs $768^2$ in HPA)}
 & $0.2$\,m/px aerial \textit{(vs $\sim$0.3--1.5\,m satellite)}
 & high-res X-ray \textit{(unchanged)} \\
\rowcolor{rowshade}
Labels
 & subcellular localisation \textit{(clustering protocol)}
 & 15 land-cover classes, segmentation masks \textit{(vs 62 image-level classes in FMoW)}
 & 14 findings, \textit{Stanford vs Beth Israel} \\
\bottomrule
\end{tabular}

\vspace{1mm}
{\footnotesize \discrete\,discrete \quad \continuous\,continuous metadata}

%% file: figures/algo.tex
\begin{algorithmic}[1]
\State Sample minibatch $\{(x_n, \{m_n^{(t)}\}_{t \in \mathcal{T}})\}_{n=1}^{B}$
\State Compute student $\phi(x_n)$ and teacher $\hat{\phi}(x_n)$ embeddings
\State Compute $\cL_{\text{DINO}}$ and $\cL_{\text{iBOT}}$
\For{each metadata type $t \in \mathcal{T}$}
    \If{$t$ discrete}
        \State Compute $\cL_{\text{meta}}^{(t)}$ via prototypes $\{p_k^t\}$; update prototypes by EMA
    \Else
        \State Compute $\cL_{\text{meta}}^{(t)}$ via predictor $g^{(t)}$
    \EndIf
    \If{$t \in \bM_-$}
        \State Apply gradient reversal to encoder
    \EndIf
\EndFor
\State Equalise per-branch gradient norms by $s_t = \bar n / \tilde n_t$ (App.~\ref{app:grad_equal})
\State Apply $\cL_{\text{SIGReg}}$ to the pre-normalisation bottleneck
\State Update student by AdamW and teacher by EMA
\end{algorithmic}

%% file: figures/algo_grad_equal.tex
\begin{algorithmic}[1]
\Require metadata branches $\mathcal{T}$; EMA decay $\mu = 0.99$; fixed weights $\{\lambda_{\text{meta}}^{(t)}\}_{t \in \mathcal{T}}$; running norms $\{\tilde n_t\}_{t \in \mathcal{T}}$
\For{each branch $t \in \mathcal{T}$}
    \State $\tilde n_t \gets \mu\, \tilde n_t + (1-\mu)\, \big\lVert \nabla_{\phi(x)}\, \cL_{\text{meta}}^{(t)} \big\rVert_2$ \Comment{EMA of gradient norm at the CLS embedding}
\EndFor
\State $\bar n \gets \big(\prod_{t \in \mathcal{T}} \tilde n_t\big)^{1/|\mathcal{T}|}$ \Comment{geometric-mean target across branches}
\For{each $t \in \mathcal{T}$}
    \State $s_t \gets \texttt{detach}\big(\bar n / \tilde n_t\big)$ \Comment{closed-form scale; no second-order gradient}
\EndFor
\State \Return $\sum_{t \in \mathcal{T}} s_t\, \lambda_{\text{meta}}^{(t)}\, \cL_{\text{meta}}^{(t)}$ \Comment{balanced metadata loss for this step}
\end{algorithmic}

%% file: figures/iwildcam_convergence.tex
\providecommand{\cIID}{}    %
\providecommand{\cOOD}{}
\definecolor{cTrainLoss}{HTML}{B6522E}     %
\definecolor{cTrainAcc}{HTML}{7B44DB}      %

\pgfplotsset{
    iwc/.style={
        width=0.46\linewidth,
        height=5.2cm,
        xmin=0, xmax=300,
        xtick={0,50,100,150,200,250,300},
        xlabel={Epoch},
        axis line style={black!70},
        tick style={black!70},
        grid=major,
        grid style={gray!18},
        tick label style={font=\small},
        label style={font=\small},
        ylabel near ticks,
        legend style={font=\scriptsize, draw=none, fill=none, row sep=-2pt},
    },
}

\begin{tikzpicture}

\begin{axis}[
    iwc,
    name=train,
    axis y line*=left,
    axis x line*=bottom,
    ylabel={\textcolor{cTrainLoss}{Training loss}},
    ymin=1.0, ymax=5.5,
    ytick={1,2,3,4,5},
    title={\small\bf Training dynamics},
    legend style={
        at={(0.98,0.55)}, anchor=east,
        font=\scriptsize, draw=none, fill=white, fill opacity=0.85, text opacity=1,
        row sep=-2pt, legend cell align=left,
    },
]
\addplot[cTrainLoss, very thick] coordinates {
  (0.00,5.2117)(1.59,3.0711)(3.17,2.6563)(4.76,2.4050)(6.35,2.2687)(7.94,2.1877)
  (9.52,2.0691)(11.11,1.9988)(12.70,2.0655)(14.29,2.0600)(15.87,1.9894)(17.46,1.9339)
  (19.05,1.9170)(20.63,2.0135)(22.22,1.9444)(23.81,2.0315)(25.40,1.9113)(26.98,1.9471)
  (28.57,1.8677)(30.16,1.8670)(31.75,1.8746)(33.33,1.8902)(34.92,1.8864)(36.51,1.8157)
  (38.10,1.8351)(39.68,1.7373)(41.27,1.6945)(42.86,1.7644)(44.44,1.7526)(46.03,1.7460)
  (47.62,1.7044)(49.21,1.7436)(50.79,1.7614)(52.38,1.7138)(53.97,1.7936)(55.56,1.6987)
  (57.14,1.7623)(58.73,1.7512)(60.32,1.6762)(61.90,1.6952)(63.49,1.6511)(65.08,1.6366)
  (66.67,1.6218)(68.25,1.6298)(69.84,1.6289)(71.43,1.6245)(73.02,1.6494)(74.60,1.6916)
  (76.19,1.5998)(77.78,1.5579)(79.37,1.5874)(80.95,1.5936)(82.54,1.5267)(84.13,1.5100)
  (85.71,1.5394)(87.30,1.5600)(88.89,1.5550)(90.48,1.5203)(92.06,1.5139)(93.65,1.4992)
  (95.24,1.4978)(96.83,1.4789)(98.41,1.4012)(100.00,1.4621)(101.59,1.3942)(103.17,1.3921)
  (104.76,1.3888)(106.35,1.3590)(107.94,1.3748)(109.52,1.3609)(111.11,1.3496)(112.70,1.3447)
  (114.29,1.3365)(115.87,1.2999)(117.46,1.3106)(119.05,1.2915)(120.63,1.3292)(122.22,1.2893)
  (123.81,1.2321)(125.40,1.2452)(126.98,1.2579)(128.57,1.2914)(130.16,1.2181)(131.75,1.2665)
  (133.33,1.2287)(134.92,1.2339)(136.51,1.2163)(138.10,1.2137)(139.68,1.1973)(141.27,1.2253)
  (142.86,1.2084)(144.44,1.2077)(146.03,1.1862)(147.62,1.1812)(149.21,1.1485)(150.79,1.1777)
  (152.38,1.1924)(153.97,1.1670)(155.56,1.1670)(157.14,1.1874)(158.73,1.1527)(160.32,1.1363)
  (161.90,1.1854)(163.49,1.1309)(165.08,1.1351)(166.67,1.1511)(168.25,1.1426)(169.84,1.1431)
  (171.43,1.1710)(173.02,1.1241)(174.60,1.1549)(176.19,1.1550)(177.78,1.1636)(179.37,1.1521)
  (180.95,1.1309)(182.54,1.1188)(184.13,1.1535)(185.71,1.1321)(187.30,1.1488)(188.89,1.1384)
  (190.48,1.1449)(192.06,1.1138)(193.65,1.1160)(195.24,1.1393)(196.83,1.1205)(198.41,1.1447)
  (200.00,1.1419)(201.59,1.1724)(203.17,1.1836)(204.76,1.1838)(206.35,1.1538)(207.94,1.1587)
  (209.52,1.1320)(211.11,1.1391)(212.70,1.1502)(214.29,1.1446)(215.87,1.1199)(217.46,1.1307)
  (219.05,1.1001)(220.63,1.1261)(222.22,1.1099)(223.81,1.1456)(225.40,1.1203)(226.98,1.1413)
  (228.57,1.1214)(230.16,1.1158)(231.75,1.1332)(233.33,1.1012)(234.92,1.0895)(236.51,1.0815)
  (238.10,1.1009)(239.68,1.0951)(241.27,1.1078)(242.86,1.0830)(244.44,1.0993)(246.03,1.0816)
  (247.62,1.0845)(249.21,1.0994)(250.79,1.0700)(252.38,1.0705)(253.97,1.0675)(255.56,1.0618)
  (257.14,1.0825)(258.73,1.0956)(260.32,1.0756)(261.90,1.0735)(263.49,1.0525)(265.08,1.0827)
  (266.67,1.0758)(268.25,1.0534)(269.84,1.0487)(271.43,1.0749)(273.02,1.0512)(274.60,1.0562)
  (276.19,1.0604)(277.78,1.0481)(279.37,1.0436)(280.95,1.0594)(282.54,1.0817)(284.13,1.0631)
  (285.71,1.0745)(287.30,1.0716)(288.89,1.0747)(290.48,1.0794)(292.06,1.0587)(293.65,1.0384)
  (295.24,1.0623)(296.83,1.0528)(298.41,1.0437)
};
\end{axis}

\begin{axis}[
    iwc,
    at=(train.south west), anchor=south west,
    axis y line*=right,
    axis x line=none,
    ylabel={\textcolor{cTrainAcc}{Training accuracy (\%)}},
    ymin=0, ymax=100,
    ytick={0,20,40,60,80,100},
    grid=none,
]
\addplot[cTrainAcc, very thick] coordinates {
  (0.00,0.342)(1.59,40.185)(3.17,45.654)(4.76,52.344)(6.35,55.078)(7.94,55.762)
  (9.52,60.791)(11.11,62.842)(12.70,62.647)(14.29,62.256)(15.87,62.988)(17.46,64.307)
  (19.05,66.016)(20.63,62.695)(22.22,65.332)(23.81,63.281)(25.40,65.723)(26.98,63.135)
  (28.57,66.357)(30.16,66.748)(31.75,66.357)(33.33,66.553)(34.92,65.918)(36.51,68.750)
  (38.10,66.943)(39.68,71.045)(41.27,72.949)(42.86,70.312)(44.44,70.166)(46.03,70.606)
  (47.62,71.728)(49.21,69.629)(50.79,69.580)(52.38,70.654)(53.97,68.555)(55.56,71.680)
  (57.14,69.873)(58.73,70.312)(60.32,72.022)(61.90,72.217)(63.49,73.047)(65.08,72.363)
  (66.67,73.682)(68.25,73.438)(69.84,74.561)(71.43,74.414)(73.02,73.975)(74.60,72.119)
  (76.19,75.049)(77.78,76.269)(79.37,76.025)(80.95,74.707)(82.54,77.441)(84.13,77.441)
  (85.71,76.514)(87.30,76.221)(88.89,77.393)(90.48,75.977)(92.06,77.734)(93.65,78.809)
  (95.24,78.467)(96.83,78.467)(98.41,81.787)(100.00,80.371)(101.59,80.762)(103.17,80.664)
  (104.76,82.227)(106.35,82.129)(107.94,81.641)(109.52,81.689)(111.11,82.275)(112.70,83.203)
  (114.29,83.838)(115.87,84.473)(117.46,84.961)(119.05,85.107)(120.63,83.203)(122.22,84.717)
  (123.81,87.647)(125.40,86.719)(126.98,87.451)(128.57,84.766)(130.16,87.695)(131.75,85.938)
  (133.33,87.549)(134.92,87.109)(136.51,86.475)(138.10,87.451)(139.68,88.281)(141.27,86.768)
  (142.86,88.184)(144.44,87.744)(146.03,89.356)(147.62,88.965)(149.21,90.088)(150.79,88.965)
  (152.38,87.451)(153.97,88.916)(155.56,89.453)(157.14,88.623)(158.73,89.404)(160.32,89.697)
  (161.90,87.988)(163.49,90.918)(165.08,90.723)(166.67,89.600)(168.25,90.478)(169.84,90.674)
  (171.43,88.965)(173.02,91.016)(174.60,89.893)(176.19,89.307)(177.78,90.088)(179.37,89.307)
  (180.95,90.967)(182.54,91.504)(184.13,89.990)(185.71,90.430)(187.30,89.648)(188.89,90.283)
  (190.48,89.844)(192.06,91.406)(193.65,90.723)(195.24,90.234)(196.83,90.576)(198.41,89.941)
  (200.00,90.186)(201.59,89.111)(203.17,88.477)(204.76,88.232)(206.35,89.990)(207.94,89.795)
  (209.52,90.527)(211.11,90.576)(212.70,89.893)(214.29,89.990)(215.87,91.943)(217.46,90.918)
  (219.05,92.090)(220.63,89.795)(222.22,90.674)(223.81,90.234)(225.40,90.625)(226.98,90.186)
  (228.57,90.772)(230.16,91.553)(231.75,90.381)(233.33,91.406)(234.92,91.797)(236.51,92.627)
  (238.10,91.260)(239.68,92.090)(241.27,91.064)(242.86,91.992)(244.44,90.967)(246.03,92.041)
  (247.62,91.846)(249.21,91.309)(250.79,92.871)(252.38,93.213)(253.97,93.457)(255.56,92.432)
  (257.14,91.992)(258.73,91.992)(260.32,92.822)(261.90,92.432)(263.49,93.408)(265.08,92.334)
  (266.67,93.408)(268.25,93.115)(269.84,93.701)(271.43,92.236)(273.02,93.115)(274.60,92.627)
  (276.19,92.481)(277.78,93.603)(279.37,93.799)(280.95,92.627)(282.54,91.602)(284.13,93.213)
  (285.71,91.846)(287.30,92.578)(288.89,91.846)(290.48,92.236)(292.06,93.164)(293.65,93.994)
  (295.24,93.115)(296.83,93.066)(298.41,93.652)
};
\end{axis}

\node[anchor=east, font=\scriptsize, inner sep=2pt,
      fill=white, fill opacity=0.85, text opacity=1, draw=none]
    at ($(train.east)+(-0.5cm,-0.7cm)$) {%
        \begin{tabular}{@{}l@{\,}l@{}}
            \textcolor{cTrainLoss}{\rule[0.5ex]{10pt}{1.2pt}} & loss \\[-1pt]
            \textcolor{cTrainAcc}{\rule[0.5ex]{10pt}{1.2pt}} & accuracy \\
        \end{tabular}
    };

\begin{axis}[
    iwc,
    at=(train.south east), anchor=south west, xshift=2.4cm,
    name=eval,
    ylabel={Macro F1 (\%)},
    ymin=0, ymax=32,
    ytick={0,5,10,15,20,25,30},
    title={\small\bf Held-out F1 (val vs.\ test)},
    legend columns=2,
    legend style={
        at={(0.98,0.50)}, anchor=east,
        font=\scriptsize, draw=none,
        fill=white, fill opacity=0.9, text opacity=1,
        /tikz/every even column/.append style={column sep=6pt},
        row sep=-2pt, legend cell align=left,
    },
]
\addplot[cIID, very thick, mark=*, mark size=1.4pt] coordinates {
  (10.00,10.9554)(20.00,13.6008)(30.00,16.4111)(40.00,16.7680)(50.00,16.3761)
  (60.00,17.8396)(70.00,16.0545)(80.00,19.7999)(90.00,20.7092)(100.00,21.1621)
  (110.00,21.8302)(120.00,24.9129)(130.00,26.2432)(140.00,26.1603)(150.00,24.9638)
  (160.00,26.4278)(170.00,27.2079)(180.00,27.4633)(190.00,26.7317)(200.00,26.3776)
  (210.00,25.9781)(220.00,25.7926)(230.00,28.9252)(240.00,27.5096)(250.00,28.7490)
  (260.00,28.8165)(270.00,28.7894)(280.00,28.9006)(290.00,29.5948)(300.00,29.1491)
};
\addlegendentry{test ID}
\addplot[cIID, semithick, mark=o, mark size=1.4pt, mark options={solid, fill=white, line width=0.6pt}] coordinates {
  (10.00,13.9386)(20.00,15.0837)(30.00,17.1293)(40.00,16.3003)(50.00,18.5756)
  (60.00,22.3012)(70.00,22.0244)(80.00,20.3153)(90.00,22.9216)(100.00,24.9464)
  (110.00,23.3894)(120.00,24.8344)(130.00,24.9520)(140.00,26.9924)(150.00,25.8088)
  (160.00,27.9722)(170.00,25.9897)(180.00,26.5779)(190.00,26.3026)(200.00,26.8054)
  (210.00,25.9039)(220.00,27.3137)(230.00,27.7753)(240.00,26.0345)(250.00,28.0556)
  (260.00,26.2041)(270.00,26.2436)(280.00,26.3776)(290.00,26.0739)(300.00,25.9421)
};
\addlegendentry{val ID}
\addplot[cOOD, very thick, mark=square*, mark size=1.4pt] coordinates {
  (10.00,4.9609)(20.00,6.1326)(30.00,6.5595)(40.00,7.4321)(50.00,7.8810)
  (60.00,7.3211)(70.00,7.1730)(80.00,7.9594)(90.00,7.0971)(100.00,8.0861)
  (110.00,9.2945)(120.00,9.1930)(130.00,9.6853)(140.00,9.4477)(150.00,9.5373)
  (160.00,9.2635)(170.00,9.3762)(180.00,9.5238)(190.00,9.8409)(200.00,9.8542)
  (210.00,9.8713)(220.00,9.8129)(230.00,10.6119)(240.00,10.4036)(250.00,10.3846)
  (260.00,10.0955)(270.00,9.9999)(280.00,9.6979)(290.00,9.5756)(300.00,9.6255)
};
\addlegendentry{test OOD}
\addplot[cOOD, semithick, mark=square, mark size=1.4pt, mark options={solid, fill=white, line width=0.6pt}] coordinates {
  (10.00,6.5072)(20.00,7.2280)(30.00,7.5657)(40.00,7.5762)(50.00,7.6209)
  (60.00,8.4277)(70.00,8.3737)(80.00,7.7685)(90.00,8.5077)(100.00,9.1112)
  (110.00,9.4377)(120.00,9.5538)(130.00,9.2990)(140.00,9.6080)(150.00,10.2458)
  (160.00,10.0961)(170.00,10.1647)(180.00,10.2671)(190.00,10.1050)(200.00,10.1599)
  (210.00,9.9846)(220.00,10.1233)(230.00,10.5441)(240.00,10.4518)(250.00,11.2138)
  (260.00,11.3380)(270.00,11.1371)(280.00,11.1571)(290.00,11.0118)(300.00,10.9932)
};
\addlegendentry{val OOD}
\end{axis}

\end{tikzpicture}

%% file: checklist.tex
\section*{NeurIPS Paper Checklist}

\begin{enumerate}

\item {\bf Claims}
    \item[] Question: Do the main claims made in the abstract and introduction accurately reflect the paper's contributions and scope?
    \item[] Answer: \answerYes{} %
    \item[] Justification: We make two central claims. First, that metadata can be used to adapt a generic foundation model to a new application domain. Second, that this form of label-free adaptation yields representations that are both more robust and more general than standard label-driven adaptation. The first claim is supported by the overall performance gains over standard adaptation approaches, and further analyzed in~\cref{fig:compass}, which measures the impact of individual metadata factors. The second claim is supported by the results in Tab.~\ref{tab:main_extended}, where our method outperforms matched adaptation baselines and obtains higher reported scores than selected specialized systems on three domains while matching the fourth, under differing cross-paper protocols, using only a lightweight probe. We further demonstrate improved generality in Tab.~\ref{tab:transfert}, where representations adapted without labels transfer across datasets within the same domain better than supervised adaptation. Finally, Fig.~\ref{fig:lowsup} shows that our representations remain expressive across both low- and high-supervision regimes.

\item {\bf Limitations}
    \item[] Question: Does the paper discuss the limitations of the work performed by the authors?
    \item[] Answer: \answerYes{} %
    \item[] Justification: Limitations are discussed in the Conclusion and its Broader impact paragraph, including the need to distinguish informative, spurious, redundant, and entangled metadata factors, and potential bias from sensitive or imbalanced metadata.

\item {\bf Theory assumptions and proofs}
    \item[] Question: For each theoretical result, does the paper provide the full set of assumptions and a complete (and correct) proof?
    \item[] Answer: \answerNA{}
    \item[] Justification: The paper does not include theoretical results or formal proofs; all contributions are empirical.

    \item {\bf Experimental result reproducibility}
    \item[] Question: Does the paper fully disclose all the information needed to reproduce the main experimental results of the paper to the extent that it affects the main claims and/or conclusions of the paper (regardless of whether the code and data are provided or not)?
    \item[] Answer: \answerYes{}
    \item[] Justification: We provide detailed training procedures, hyperparameters, and a step-by-step algorithmic description in the supplementary material. In addition, we release the complete end-to-end code (data and metadata acquisition, preprocessing, training, probing, and evaluation); laboratory policy prevents redistribution of trained checkpoints.

\item {\bf Open access to data and code}
    \item[] Question: Does the paper provide open access to the data and code, with sufficient instructions to faithfully reproduce the main experimental results, as described in supplemental material?
    \item[] Answer: \answerYes{}
    \item[] Justification: All datasets are existing benchmarks. Our public repository provides the end-to-end code for data and metadata download, preprocessing, training, probing, and evaluation, with configurations and commands; the supplementary material documents the protocols. Laboratory policy prevents redistribution of trained checkpoints. %

\item {\bf Experimental setting/details}
    \item[] Question: Does the paper specify all the training and test details (e.g., data splits, hyperparameters, how they were chosen, type of optimizer) necessary to understand the results?
    \item[] Answer: \answerYes{}
    \item[] Justification: The paper specifies all key experimental details, including datasets, evaluation protocols, and model components in the main text. Full training details, hyperparameters, and implementation specifics are provided in the supplementary material.

\item {\bf Experiment statistical significance}
    \item[] Question: Does the paper report error bars suitably and correctly defined or other appropriate information about the statistical significance of the experiments?
    \item[] Answer: \answerYes{}
    \item[] Justification: For experiments involving stochastic label subsampling (low-supervision regime), we report standard deviations across different label samplings (Fig.~\ref{fig:lowsup}). For the main results, \textsc{FINO} reports mean$\pm$standard deviation over complete adaptation repeats (8, 5, and 5 runs on iWildCam, FMoW, and MIMIC-CXR; 10 low-resolution runs on HPA), see Tab.~\ref{tab:main_extended}, Tab.~\ref{tab:ablation_hpa}, and the supplementary material.

\item {\bf Experiments compute resources}
    \item[] Question: For each experiment, does the paper provide sufficient information on the computer resources (type of compute workers, memory, time of execution) needed to reproduce the experiments?
    \item[] Answer: \answerYes{}
\item[] Justification: We provide a detailed breakdown of compute resources in the supplementary material, including development and final trainings.

\item {\bf Code of ethics}
    \item[] Question: Does the research conducted in the paper conform, in every respect, with the NeurIPS Code of Ethics \url{https://neurips.cc/public/EthicsGuidelines}?
    \item[] Answer: \answerYes{}
    \item[] Justification: The research complies with the NeurIPS Code of Ethics. All datasets used are publicly available and commonly used in the literature, and no sensitive or personally identifiable information is processed. We also discuss potential biases and ethical considerations related to metadata usage in the paper.

\item {\bf Broader impacts}
    \item[] Question: Does the paper discuss both potential positive societal impacts and negative societal impacts of the work performed?
    \item[] Answer: \answerYes{}
    \item[] Justification: The Broader impact paragraph discusses the main risk, namely that metadata guidance may reproduce or amplify biases encoded by sensitive or imbalanced attributes, and recommends careful metadata selection and auditing beyond aggregate performance. The positive impact, improved label-free adaptation of foundation models to scientific domains, is the subject of the paper.

\item {\bf Safeguards}
    \item[] Question: Does the paper describe safeguards that have been put in place for responsible release of data or models that have a high risk for misuse (e.g., pre-trained language models, image generators, or scraped datasets)?
    \item[] Answer: \answerNA{}
    \item[] Justification: The paper does not introduce or release models or datasets that pose a high risk of misuse. It relies on standard vision models and publicly available datasets, and therefore does not require specific safeguards for controlled release.

\item {\bf Licenses for existing assets}
    \item[] Question: Are the creators or original owners of assets (e.g., code, data, models), used in the paper, properly credited and are the license and terms of use explicitly mentioned and properly respected?
    \item[] Answer: \answerYes{}
    \item[] Justification: All datasets and models used in this work are publicly available and properly cited. We follow their respective terms of use and licenses, as described in the original publications.

\item {\bf New assets}
    \item[] Question: Are new assets introduced in the paper well documented and is the documentation provided alongside the assets?
    \item[] Answer: \answerYes{}
    \item[] Justification: The released code is documented in the public repository, with instructions, configurations, and commands for every stage; trained checkpoints are not redistributed.

\item {\bf Crowdsourcing and research with human subjects}
    \item[] Question: For crowdsourcing experiments and research with human subjects, does the paper include the full text of instructions given to participants and screenshots, if applicable, as well as details about compensation (if any)?
    \item[] Answer: \answerNA{}
    \item[] Justification: The paper does not involve crowdsourcing or experiments with human subjects. All datasets used are publicly available and were collected by their original authors.

\item {\bf Institutional review board (IRB) approvals or equivalent for research with human subjects}
    \item[] Question: Does the paper describe potential risks incurred by study participants, whether such risks were disclosed to the subjects, and whether Institutional Review Board (IRB) approvals (or an equivalent approval/review based on the requirements of your country or institution) were obtained?
    \item[] Answer: \answerNA{}
\item[] Justification: The paper does not involve crowdsourcing or research with human subjects. All datasets used are publicly available and no new data collection was performed.

\item {\bf Declaration of LLM usage}
    \item[] Question: Does the paper describe the usage of LLMs if it is an important, original, or non-standard component of the core methods in this research? Note that if the LLM is used only for writing, editing, or formatting purposes and does \emph{not} impact the core methodology, scientific rigor, or originality of the research, declaration is not required.
    \item[] Answer: \answerNA{}
    \item[] Justification: LLMs were used only for editing purposes and do not contribute to the core methodology or results of the paper.

\end{enumerate}